\ifdefined\pdfminorversion\pdfminorversion=7\relax\fi
\documentclass[10pt,twocolumn,letterpaper]{article}
\usepackage[pagenumbers]{cvpr}
\usepackage{amsmath,tabularx}
\usepackage{placeins}

\usepackage{booktabs,array,multirow,amssymb}
\usepackage{tikz}
\usepackage{colortbl}
\usepackage{pifont}
\usepackage{fontawesome5}
\usepackage{rotating}

\definecolor{cA1}{HTML}{3F7CAC}\definecolor{cA2}{HTML}{C96F4F}
\definecolor{cA3}{HTML}{4E9A72}\definecolor{cA4}{HTML}{9B6FA8}
\definecolor{bluemain}{HTML}{2F5F8F}
\definecolor{neutralL}{HTML}{CFCECE}\definecolor{neutralM}{HTML}{767676}
\definecolor{neutralD}{HTML}{4D4D4D}
\colorlet{oursbg}{cA1!12}

\definecolor{cTxt}{HTML}{2F5F8F}\definecolor{cImg}{HTML}{3C9B9B}
\definecolor{cRoom}{HTML}{8E62A6}\definecolor{cDoor}{HTML}{C08A4A}
\definecolor{cScene}{HTML}{4E9A72}
\definecolor{cRel}{HTML}{5B7691}\definecolor{cProg}{HTML}{5B7691}
\definecolor{cOut}{HTML}{2F5F8F}

\newcommand{\ym}[1]{{\fontsize{6.5}{6.5}\selectfont\textcolor{#1}{\faCheck}}}

\newcommand{\xmark}{{\fontsize{6}{6}\selectfont\textcolor{neutralL!85!neutralM}{\faTimes}}}
\newcommand{\hm}[1]{{\fontsize{6}{6}\selectfont\textcolor{#1}{\faCircle[regular]}}}

\newcommand{\ghub}{\,{\fontsize{5.5}{5.5}\selectfont\textcolor{neutralM}{\faGithub}}}

\newcommand{\seg}[1]{\textcolor{#1}{\rule[0.2pt]{3pt}{4.2pt}}}
\newcommand{\meter}[1]{\hspace{2pt}\ifcase#1 \seg{neutralL}\seg{neutralL}\seg{neutralL}\seg{neutralL}\or\seg{cA4}\seg{neutralL}\seg{neutralL}\seg{neutralL}\or\seg{cA4}\seg{cA4}\seg{neutralL}\seg{neutralL}\or\seg{cA4}\seg{cA4}\seg{cA4}\seg{neutralL}\else\seg{cA4}\seg{cA4}\seg{cA4}\seg{cA4}\fi}
\newcommand{\tHard}{\textbf{Hard}\meter{4}}
\newcommand{\tCheck}{\textbf{Check}\meter{3}}
\newcommand{\tSolver}{Solver\meter{2}}
\newcommand{\tBoK}{Best-of-$K$\meter{1}}
\newcommand{\tRL}{RL\meter{1}}
\newcommand{\tLoss}{Loss\meter{1}}

\newcommand{\tNone}{None\meter{0}}

\newcommand{\chip}[2]{\tikz[baseline=-0.55ex]{\node[fill=#1!22,text=#1!95!black,rounded corners=1.2pt,inner xsep=2pt,inner ysep=1.2pt,font=\fontsize{5.4}{5.4}\selectfont\sffamily\bfseries]{#2};}\hspace{1pt}}
\newcommand{\iTxt}{\chip{cTxt}{T}}\newcommand{\iImg}{\chip{cImg}{I}}
\newcommand{\iRoom}{\chip{cRoom}{R}}\newcommand{\iDoor}{\chip{cDoor}{D}}
\newcommand{\iScene}{\chip{cScene}{S}}\newcommand{\iNoise}{\chip{cRel}{N}}
\newcommand{\iRules}{\chip{cRel}{rules}}
\newcommand{\oRel}{\chip{cRel}{rel}}\newcommand{\oProg}{\chip{cProg}{prog}}
\newcommand{\oOps}{\chip{cProg}{ops}}
\newcommand{\oTwo}{\chip{cOut}{2D}}\newcommand{\oThree}{\chip{cOut}{3D}}
\newcommand{\oMesh}{\chip{cOut}{mesh}}
\newcommand{\lnk}{\hspace{0.3pt}\raisebox{2.4pt}{\textcolor{neutralL!70!neutralM}{\rule{5pt}{0.55pt}}}\hspace{1.5pt}}
\newcolumntype{L}[1]{>{\raggedright\arraybackslash}p{#1}}
\newcommand{\rtab}[3]{\multicolumn{#1}{c}{\textcolor{#2}{\fontsize{7}{7}\selectfont\bfseries #3}}}
\newcolumntype{S}{>{\columncolor{cA1!14}}>{\centering\arraybackslash}p{1.55cm}}

\usepackage{titletoc}
\titlecontents{appsection}[0pt]{\addvspace{7pt}\large\bfseries}
  {\thecontentslabel\hspace{1em}}{}{\hfill\textcolor{cvprblue}{\contentspage}}
\titlecontents{appsubsection}[1.6em]{\addvspace{2pt}\normalsize}
  {\thecontentslabel\hspace{1em}}{}{\hfill\textcolor{cvprblue}{\contentspage}}
\definecolor{cvprblue}{rgb}{0.21,0.49,0.74}
\usepackage[pagebackref,breaklinks,colorlinks,allcolors=cvprblue]{hyperref}

\def\confName{CVPR}
\def\confYear{2027}
\newcommand{\method}{\textsc{WorldEdge}}
\title{%
    \texorpdfstring{%
        \raisebox{-0.2em}{%
            \includegraphics[height=1.15em]{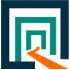}%
        }\hspace{0.3em}%
    }{}%
    \method: Persistent Task-Driven World Expansion
}
\author{%
    Liantian Shan\textsuperscript{1,*}\quad Runyi Yang\textsuperscript{2,*}\quad
    Yang Zhang\textsuperscript{3}\quad Chuling Li\textsuperscript{3}\quad Yuqi Lin\textsuperscript{3}\\[2pt]
    Jialin Yu\textsuperscript{4}\quad Zhenyu Zhang\textsuperscript{5}\quad
    Danda Pani Paudel\textsuperscript{2}\quad Mengyue Yang\textsuperscript{1,\textdagger}\\[5pt]
    {\small\textsuperscript{1}University of Bristol\qquad
           \textsuperscript{2}INSAIT, Sofia University ``St. Kliment Ohridski''}\\
    {\small\textsuperscript{3}University of Nottingham\qquad
           \textsuperscript{4}University of Bath\qquad
           \textsuperscript{5}Stanford University}%
}
\hypersetup{
    pdftitle={WorldEdge: Persistent Task-Driven World Expansion},
    pdfauthor={Liantian Shan, Runyi Yang, Yang Zhang, Chuling Li, Yuqi Lin, Jialin Yu, Zhenyu Zhang, Danda Pani Paudel, Mengyue Yang}
}
\begin{document}
\twocolumn[{%
    \centering
    {\Large\bfseries\thetitle\par}\vspace{9pt}
    {\large\begin{tabular}{@{}c@{}}\csname @author\endcsname\end{tabular}\par}
    \vspace{7pt}
    \includegraphics[width=.87\textwidth]{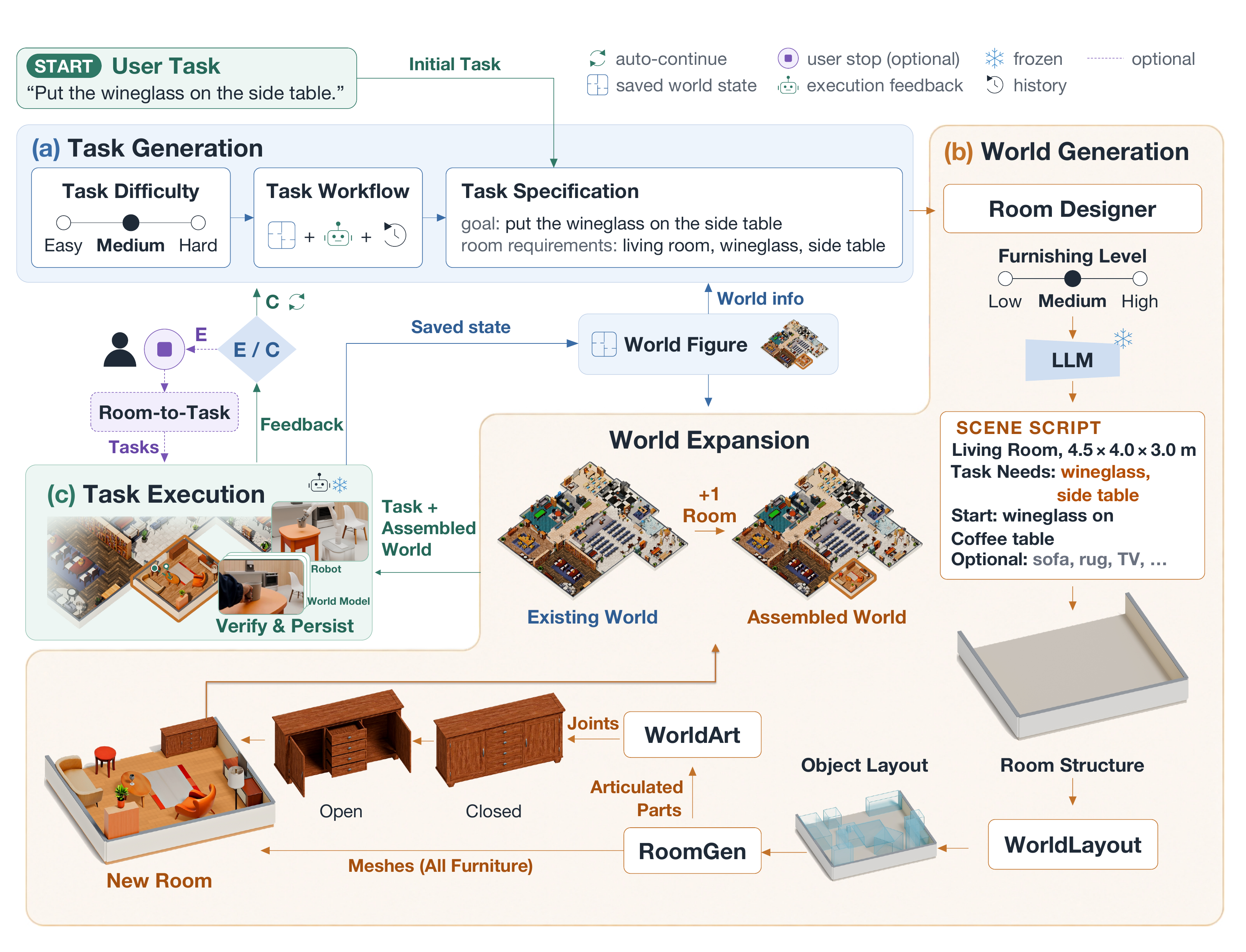}

    \captionof{figure}{
        \textbf{Overview of \method{}.}
        (a) Task generation specifies new room requirements.
        (b) World generation incrementally expands the persistent world.
        (c) Downstream models can provide optional feedback for subsequent tasks and world expansion.
    }
    \label{fig:pipeline}
    \vspace{8pt}
}]
\begingroup
\renewcommand{\thefootnote}{}
\footnotetext{\begin{tabular}[t]{@{}l@{}}
    \textsuperscript{*}Equal contribution.\quad
    \textsuperscript{\textdagger}Corresponding author.\\
    Email: \href{mailto:liantian.shan@bristol.ac.uk}{\texttt{liantian.shan@bristol.ac.uk}}
\end{tabular}}
\endgroup

\begin{abstract}
Recent advances in 3D generation have enabled diverse and interactive environments, yet most approaches focus on constructing isolated scenes rather than persistent worlds that evolve across tasks. Supporting such worlds requires accommodating new task demands while preserving existing structures and the consequences of past interactions. We introduce WorldEdge, a framework for persistent task-driven world expansion that couples task generation, world construction, and optional execution feedback. Given the current world and its interaction history, WorldEdge generates feasible tasks that guide the addition of connected rooms with task-relevant objects, validated layouts, and articulated structures. Each expansion preserves existing geometry, object identities, and interaction states, while execution feedback can inform subsequent tasks. We construct a task dataset containing 480 scenes and 3,581 tasks. Experiments demonstrate improved task-conditioned environment generation, history-dependent task generation, and layout validity over existing methods. WorldEdge achieves a 98.9\% yield in symbolic next-task generation, with only 0.53\% object collisions in generated layouts. WorldEdge further provides persistent interactive environments for evaluating robotic policies and video world models, advancing the construction of 3D worlds that grow with successive tasks and interactions. Our code and data will be available soon.
\end{abstract}


\section{Introduction}
\label{sec:intro}

\looseness=-1

Interactive environments are not static collections of objects: they change as agents act, shaping what agents can do next. A robot that sets a dining table should find the dishes where it left them, while a new task may require objects or rooms that do not exist. Yet, most 3D generation approaches focus on constructing individual scenes rather than environments that persist and expand across tasks. This raises a fundamental challenge: how can we generate interactive worlds that accommodate new task requirements without losing the structure and history of what exists?

Procedural and language-guided methods can generate
diverse interactive environments at scale
~\cite{deitke2022,holodeck}.
RoboGen~\cite{wang2024robogen} connects task proposal,
scene construction, and robotic skill learning, while
recent systems support editable scenes, articulation,
and task-conditioned generation
~\cite{wang2026scenecode,wang2026embodiedgen}.
However, generating or editing scenes does not
necessarily support continuity across successive
tasks, where new requirements must be grounded in
the accumulated world and its interaction history.

Such continuity presents challenges at both task and environment levels. New tasks must account for existing objects and states while identifying additional requirements. World expansion must preserve geometry and object identities, ensuring that new rooms and articulated objects remain physically valid, accessible, and usable. Execution-induced changes must remain available to subsequent tasks.

We introduce \method{}, a framework for persistent task-driven world expansion. Its history-aware task generator proposes and verifies goals grounded in the current world, while its world generator constructs connected rooms with validated layouts, meshes, and articulated objects. A shared representation preserves geometry, object identities, and interaction states across expansions. Task execution updates the persistent world and provides feedback for subsequent tasks. Our contributions are threefold:
\begin{itemize}
    \item We introduce \method{}, a framework for persistent
    task-driven world expansion that couples successive tasks
    with incremental spatial growth while preserving world
    structure and interaction history.
    \item We develop history-aware task generation that grounds
    new tasks in the evolving world, verifies their symbolic
    feasibility, and selects tasks according to target difficulty.
    \item We develop a task-conditioned world generation pipeline
    that integrates connected rooms, validated layouts, and
    articulated objects into existing environments while
    maintaining physical validity and task accessibility.
\end{itemize}

\section{Related Work}
\label{sec:related}

\subsection{Task Generation}

Automatic task generation specifies agent goals and environment requirements, as in RoboGen~\cite{wang2024robogen} and OMNI-EPIC~\cite{faldor2025omni}. Task libraries are bootstrapped in GenSim~\cite{wang2024gensim} and verified in PARTNR~\cite{ICLR2025_a3cf318f}. State-conditioned continuation appears in Voyager~\cite{wang2023voyager} for Minecraft exploration, AgentSynth~\cite{xie2026agentsynth} for state-chained computer-use tasks, and SEAgent~\cite{sun2025seagent} for software curricula. WorldEdge instead couples task continuation to spatial expansion of a persistent indoor world: each task specifies a connected room and goals referencing existing objects and prior interaction states, while supporting user-specified tasks and fixed-world generation once expansion stops.

\subsection{3D World Generation and Editing}

Indoor generation spans procedural houses in ProcTHOR~\cite{deitke2022} and
language-guided scenes in Holodeck~\cite{holodeck}. Layout planners express
spatial intent via structured text in LayoutGPT~\cite{layoutgpt},
relational backtracking in I-Design~\cite{idesign}, global-to-local tree
search~\cite{gltreesearch}, semantic-geometric refinement in
LayoutVLM~\cite{layoutvlm}, agent clearance in RoboLayout~\cite{robolayout},
reachability guidance in PhyScene~\cite{yang2024physcene},
preference learning in DirectLayout~\cite{directlayout}, human-corrected
layouts in OptiScene~\cite{optiscene}, and hierarchical support motifs in
HSM~\cite{pun2026hsm}. For continued spatial growth,
WorldGrow~\cite{li2026worldgrow} uses block inpainting under a fixed
generic text condition, whereas WorldEdge drives room additions through
task requirements grounded in retained object states.

Simulation-ready pipelines connect layouts to assets and kinematics through hierarchical inpainting in ARCHITECT~\cite{wang2024architect}, reflective tool use in SceneWeaver~\cite{sceneweaver}, and task-conditioned critic refinement in SAGE~\cite{sage} and SceneSmith~\cite{pfaff2026scenesmith} (averaging 3.43 hours per room). Articulated objects consist of rigid parts connected by joints that permit relative motion. Shape2Motion~\cite{wang2019shape2motion} predicts movable parts and motion attributes from static 3D shapes, while the category-level ANCSH approach~\cite{li2020category}
estimates part poses and joint parameters from a single depth image. PARTICULATE~\cite{li2026particulate} predicts articulation from 3D meshes, while Articraft~\cite{zhou2026articraft} generates articulated assets programmatically and retrains PARTICULATE with them. Instruct-Particulate~\cite{li2026instruct} adds kinematic control. Persistent editing is supported by iterative object updates in ReSpace~\cite{respace}, goal-regressive edit planning in Edit-As-Act~\cite{Noh_2026_CVPR}, program-mesh-joint registries in SceneCode~\cite{wang2026scenecode}, stateful multi-room generation with state-preserving edits in EmbodiedGen V2~\cite{wang2026embodiedgen}, multi-floor physical write-back with persisted task-precondition edits in MANSION~\cite{Che_2026_CVPR}, and validated incremental editing in EnvDreamer~\cite{swain2026envdreamerlargescalemultimodaltoenvironmentgeneration}. These systems edit a given scene for a given task. WorldEdge couples each room addition to a history-conditioned task, preserving existing object identities and states while initializing new objects.

\subsection{Task Execution and Feedback}

Pretrained robot policies such as $\pi_{0.5}$\cite{black2025pi05} and GR00T\cite{bjorck2025gr00t} support language-conditioned manipulation. Interactive video world models such as Matrix-Game2.0\cite{he2025matrixgame2} and LingBot-World~\cite{robbyant2026lingbot} generate action-conditioned rollouts. SIMPLER~\cite{li2025simpler} and WorldScore~\cite{duan2025worldscore} benchmark policies and world generation models. WorldEdge provides environments for evaluating these models, with an LLM planner and skills as a reference. Execution feedback can refine environments: Code2Worlds-2~\cite{zhang2026code2worlds} revises physical dynamics from simulator feedback, while EmbodiedSmith~\cite{qin2026embodiedsmith} repairs scenes and action plans during data generation. SimWorld Studio~\cite{kang2026simworld} co-evolves generated environments with the agent. Optional task selection in WorldEdge draws on Sampling for Learnability (SFL)~\cite{sfl}.
\section{Method}
\label{sec:method}
As shown in Fig.~\ref{fig:pipeline}, \method{} expands a persistent indoor world across tasks. In each cycle, task generation defines the target interaction and spatial prerequisites, world generation synthesizes the required connected room, and execution updates the persistent state with feedback. Persistent object identities let later tasks build on prior state changes. The pipeline supports user-defined tasks and in-place interaction without spatial expansion. Module details are given in the supplementary material.
\subsection{Task and World Representation}
\label{sec:task_specification}

We represent the persistent world at step $t$ as
$W_t=(\mathcal R_t,\mathcal O_t,\mathcal E_t,\xi_t)$,
comprising rooms, objects with their states, spatial relations,
and the agent state. After object-role binding, each normalized
task is defined as
\begin{equation}
\tau_t=(\Delta_t,\mathcal G_t,u_t),
\label{eq:method:task}
\end{equation}
where $\Delta_t=(\mathcal R_{\Delta,t},\mathcal O_{\Delta,t},\mathcal E_{\Delta,t})$
specifies the new room, required new objects, and their initial
conditions, $\mathcal G_t$ defines the task goals, and $u_t$ is
the corresponding natural-language instruction. When spatial expansion stops,
$\Delta_t=\varnothing$, supporting continued task execution within the
accumulated persistent world without adding any additional rooms.

Runs start either from an empty world $W_0=(\varnothing,\varnothing,\varnothing,\xi_0)$
with an empty interaction history $H_0=\varnothing$, or from a
supplied 3D scene. The first task establishes the initial room,
and each expansion attaches one connected room while preserving
existing object IDs and states:
\begin{equation}
\begin{gathered}
\mathcal R(\widehat W_t)=\mathcal R_t\cup\mathcal R_{\Delta,t},\quad
\mathcal O(\widehat W_t)=\mathcal O_t\sqcup\mathcal O_{\Delta,t}, \\
\operatorname{id}(\mathcal O_t)\cap\operatorname{id}(\mathcal O_{\Delta,t})=\varnothing.
\end{gathered}
\label{eq:method:identity}
\end{equation}
New objects in $\mathcal O_{\Delta,t}$ receive fresh unique IDs.
Existing object states are preserved during expansion and updated
after execution according to Eq.~\eqref{eq:method:persistence}.
Because $\mathcal G_t$ can span
both $\mathcal O_t$ and $\mathcal O_{\Delta,t}$, object
identities remain consistent across generation, execution, and
future tasks.

\subsection{Task Generation}
\label{sec:task_generation}

\begin{figure}[!tbp]
\centering
\includegraphics[width=0.92\columnwidth]{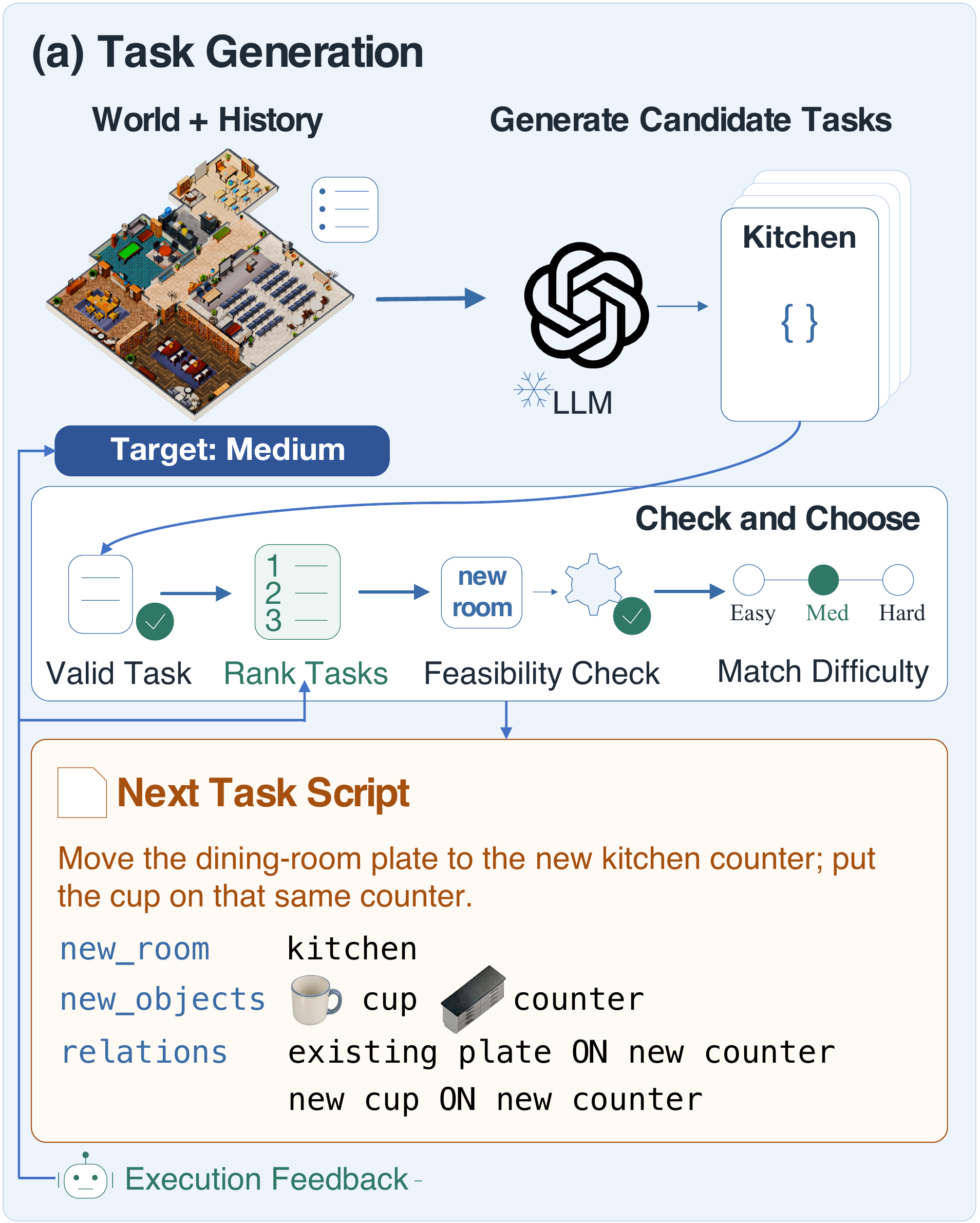}
\caption{\textbf{History-conditioned task generation.}
A structured prompt combines the current world state, persistent
interaction history, and target difficulty to guide candidate task
generation. Candidates are screened, ranked, checked for
symbolic task feasibility, and matched to the specified target difficulty.
Execution feedback always returns to the task generator; adaptive selection is optional.}
\label{fig:task_generation}
\end{figure}

\paragraph{History-conditioned proposals.}
The first task is user-specified or sampled from Room-to-Task
(Sec.~\ref{sec:experimental_setting}). For online continuation,
GPT-6.1-sol~\cite{openai2026gpt61sol} conditions on the current
world $W_t$, persistent history $H_t$, execution feedback, and
target difficulty $d^\star\in\{\text{easy},\text{medium},\text{hard}\}$
(Fig.~\ref{fig:task_generation}). Each proposal specifies a room
type, object requirements, and goals that can reuse prior
interaction effects.

\paragraph{Task verification and selection.}
Candidates are screened for valid references, consistent goals,
and required room use. Before geometric synthesis, a symbolic
solver checks each task under the action model in a temporary
world $\widehat W_{t,i}$ containing its proposed additions:
\begin{equation}
\begin{aligned}
\mathcal C_t^{\mathrm{feas}} &= \big\{\tau_{t,i}\ \big|\
\pi_{t,i}=\operatorname{Solve}(\widehat W_{t,i},\mathcal G_{t,i})\neq\varnothing\big\}, \\
d(\tau_{t,i}) &= |\pi_{t,i}|,
\end{aligned}
\label{eq:method:solver}
\end{equation}
For online proposals, $d$ counts atomic actions, including
navigation; tiers also depend on room use.
After diversity ranking, selection favors a feasible target-tier
match, then the nearest available tier. Symbolic feasibility does
not guarantee geometric access or successful robot execution.
Optional adaptive selection uses accumulated execution feedback
to adjust subsequent task selection without modifying the underlying
LLM.

\subsection{Task-Conditioned World Generation}
\label{sec:world_generation}

\begin{figure*}[!tbp]
\centering
\includegraphics[width=\textwidth]{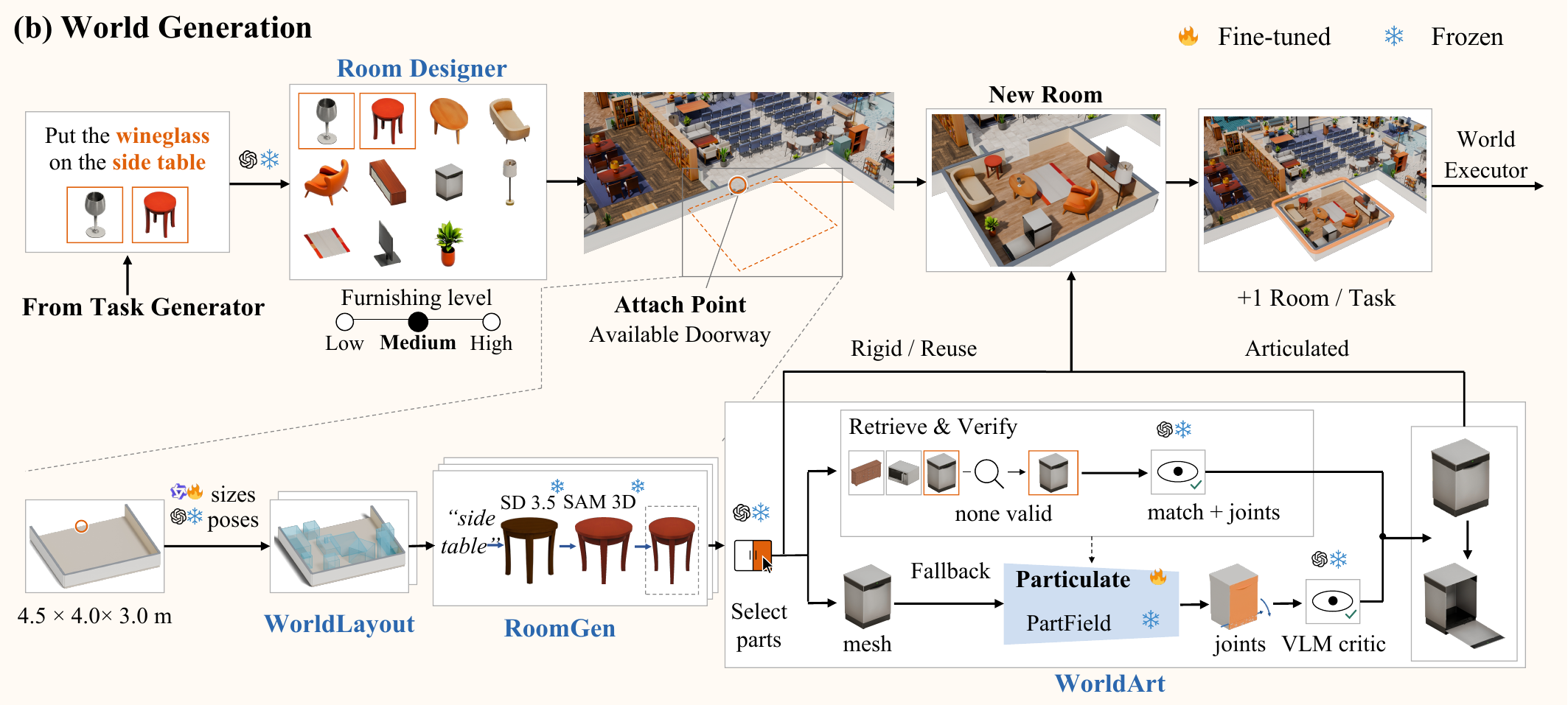}
\caption{\textbf{Task-conditioned world generation.}
Room design and verified layout establish the required object inventory
and spatial arrangement. Mesh generation and WorldArt supply detailed geometry
and task-relevant articulated motion. The generated room establishes or extends
the persistent world. Flames and snowflakes denote
fine-tuned and frozen model components, respectively.}
\label{fig:world_generator}
\end{figure*}

While a task specifies target interactions, world generation must supply
the corresponding geometry and articulated motion. As shown in
Fig.~\ref{fig:world_generator}, we decouple room design, box-level layout,
and asset realization. Persistent object IDs link boxes, meshes, and
articulated forms across these stages. WorldLayout fixes bounding-box dimensions and searches
for usable placements before mesh generation, so placement repair avoids
rebuilding assets. RoomGen and WorldArt then supply detailed meshes and
articulated motion. Final integration rechecks the realized geometry
because bounding boxes only approximate shape.

\subsubsection{Room Design}
\label{sec:room_design}

The Room Designer first identifies the objects and supports needed for
the task’s specified initial conditions, then adds contextual furnishing.
Following hierarchical furnishing in
EmbodiedGen V2~\cite{wang2026embodiedgen} and iterative critique in
SceneCode~\cite{wang2026scenecode}, it progressively adds large furniture, surface objects,
and small items at increasing density while keeping task bindings and
initial object states fixed.
Although symbolic difficulty is fixed before furnishing, clutter can
still affect geometric access and execution.

The design specifies room dimensions, an entrance, object descriptions,
supports, and placement constraints. Added rooms fix their entrance at
an available world connection; initial rooms use their designed entrance.
Initial object placement is limited to the floor or surfaces of floor-standing
furniture; physically unsupported configurations are rejected before final metric placement.

\subsubsection{Layout with Physical and Access Constraints}
\label{sec:worldlayout}

WorldLayout separates spatial intent from geometric acceptance:
a size model supplies dimensions, an agent proposes layouts, and a
program checks physical validity and robot access.
Following OptiScene~\cite{optiscene}, we fine-tune the size model;
retrieved assets use measured dimensions. Sizes remain fixed during
placement and repair.

The agent predicts support relations and object poses under room, door,
and task constraints, following relation-based
planning~\cite{holodeck,idesign,wang2026embodiedgen}.
A whole-room verifier checks each layout $L$ for collisions, support,
boundaries, access routes, and opening
clearance~\cite{sceneeval,wang2026embodiedgen}.
Geometric repair ranks layouts lexicographically by
\begin{equation}
\mathbf k(L)=(|\mathcal V_\ell|,\,M_\ell,\,|\mathcal V_a|,\,M_a,\,E_{\mathrm{soft}}),
\label{eq:method:layout}
\end{equation}
where $\mathcal V_\ell$ and $\mathcal V_a$ are validity and access
violations, $M_\ell$ and $M_a$ their respective total magnitudes, and
$E_{\mathrm{soft}}$ the soft-relation cost. Lower values are preferred.
Pose search accepts only strictly better-ranked layouts and retains all required objects.
A semantic review proposes functional corrections, accepted only after
geometric validity checks and independent comparison of both layouts.
Search and acceptance details are given in the supplementary
material.

\subsubsection{Mesh Generation}
\label{sec:roomgen}

RoomGen generates meshes for objects not supplied by retrieval,
fitting them to WorldLayout's fixed boxes and poses while preserving
supports and object IDs. Generation and alignment details are given
in the supplementary material.

\subsubsection{Task-Required Object Articulation}
\label{sec:art:main}

WorldArt reuses or retrieves previously validated articulated assets,
otherwise predicting parts and joints with adapted
PARTICULATE~\cite{li2026particulate}, trained with mixed-source
supervision including \method{}-Art.
Candidates must pass a frozen vision-language critic’s checks of
task-relevant movable parts and motion endpoints before export.
Retrieval replaces geometry;
prediction preserves it. Both preserve original object IDs.
Training, candidate selection, and export details are given in the
supplementary material.

\subsubsection{World Integration}
\label{sec:world_integration}

The initial room establishes the world; subsequent rooms attach at the
selected doorway. After asset realization, integration rechecks geometric
collisions, doorway clearance, and task preconditions. Attachment
preserves existing geometry outside permitted doorway edits and maintains
object IDs across pose changes and part partitioning. Generation or
attachment failures before commit restore $W_t$; quality violations in
committed rooms are logged for evaluation.

\subsection{Task Execution and State Persistence}
\label{sec:execution_persistence}

Execution in $\widehat W_t$ by a robot policy or an LLM planner using
skills returns trajectory $\zeta_t$, terminal state $W_t^{\mathrm{exec}}$, and outcome
$o_t\in\{0,1\}$, checked against $\mathcal G_t$ from simulator state.
For protocol
$\rho\in\{\mathrm{retain},\mathrm{rollback}\}$:
\begin{equation}
W_{t+1}=
\begin{cases}
W_t^{\mathrm{exec}}, & o_t=1,\\
W_t^{\mathrm{exec}}, & o_t=0,\ \rho=\mathrm{retain},\\
\operatorname{Restore}(W_t^{\mathrm{exec}},\widehat W_t),
& o_t=0,\ \rho=\mathrm{rollback}.
\end{cases}
\label{eq:method:persistence}
\end{equation}
Default \emph{retain} preserves all generated rooms and terminal object states after success or failure. On failure, \emph{rollback} restores pre-execution object states while keeping the newly generated room. Both modes retain failed trajectories and return execution feedback; subsequent task proposals use $W_{t+1}$ and the updated ledger $H_{t+1}$. Execution and feedback details are given in the supplementary material.
\section{Experiments}
\label{sec:experiments}

\begin{table*}[!tbp]
\centering
\caption{\textbf{Task-conditioned room generation.} (A) Single rooms. (B) Sequential additions with scripted task outcomes. Percentages are over all scheduled tasks (150 in A, 460 in B); failed or unscored rooms count as 0. Obj./room and Time average over generated rooms.}
\label{tab:task_readiness_expansion}
\small
\setlength{\tabcolsep}{3.5pt}
\renewcommand{\arraystretch}{1.12}
\begin{tabular*}{\textwidth}{@{\extracolsep{\fill}}lcccccc@{}}
\toprule
Method &
\shortstack{Task Ground.\\(\%) $\uparrow$} &
\shortstack{Task Nav.\\(\%) $\uparrow$} &
\shortstack{Physics Pass\\(\%) $\uparrow$} &
\shortstack{Task-Ready\\(\%) $\uparrow$} &
\shortstack{Obj.\\/ room} &
\shortstack{Time\\(s) $\downarrow$} \\
\midrule
\multicolumn{7}{@{}l}{\textit{A. Single-room generation}}\\
Holodeck~\cite{holodeck}
    & 3.3 & 6.7 & 0.0 & 0.0 & 21.3 & 258 \\
HSM~\cite{pun2026hsm}
    & 66.0 & 93.3 & 0.0 & 0.0 & 26.4 & 821 \\
EmbodiedGen V2~\cite{wang2026embodiedgen}
    & 46.6 & \textbf{100.0} & 0.0 & 0.0 & 12.4 & 667 \\
\textbf{WorldEdge (Ours)}
    & \textbf{86.7} & 
    99.3 & \textbf{30.7} & \textbf{17.3} & 15.4 & 2326 \\
\midrule
\multicolumn{7}{@{}l}{\textit{B. Sequential room addition ($K$ tasks, $K$ rooms)}}\\
Holodeck~\cite{holodeck}, per room + attach
    & \multicolumn{6}{c}{N/A failed Holodeck's floor-plan parser} \\
HSM~\cite{pun2026hsm}, per room + attach
    & 98.5 & 94.6 & 0.0 & 0.0 & 24.8 & 1889 \\
EmbodiedGen V2~\cite{wang2026embodiedgen}, per room + attach
    & 63.4 & 81.7 & 0.0 & 0.0 & 12.3 & 610 \\
\textbf{WorldEdge (Ours)}
    & \textbf{98.6} & \textbf{100.0} & \textbf{43.3} & \textbf{42.2} & 9.7 & 1886 \\
\bottomrule
\end{tabular*}
\end{table*}

\subsection{Tasks and evaluation setting}
\label{sec:experimental_setting}
\label{sec:evaluation_system}

\looseness=-1
We evaluate environmental executability, task feasibility, and downstream models.
Experiments cover independent rooms and sequences of connected room additions.


\paragraph{Task preparation.}
We represent each task as $\tau=(s_0,u,\mathbf a)$, where $s_0$ denotes the initial state, $u$ is a natural-language instruction specifying the required operations and intended outcome, and $\mathbf a=(a_1,\ldots,a_d)$ is an annotated action sequence for completing the task from $s_0$.
Each action $a_i=(p_i,\theta_i)$ specifies an operation $p_i$ and its arguments $\theta_i$, with $p_i$ drawn from a predefined set comprising navigate, pick, place, open, close, press, rotate, pour, and wipe.
Using the number of actions $d$, we group tasks into three difficulty levels: easy (1--4), medium (5--10), and hard (11--20).

To instantiate these tasks, we develop \textit{Room-to-Task}, an annotation workflow implemented via agent skills~\cite{anthropic_2025,openai_2025}, and apply it to single- and multi-room scenes from 16 existing 3D datasets~\cite{Fu_2021_ICCV,Wald_2019_ICCV,avetisyan2024scenescript,Yadav_2023_CVPR,Khanna_2024_CVPR,zhou2025il3d,miao2026towards,zhong2026internscenes,Zhang_2026_CVPR,Che_2026_CVPR,mao2022multiscan,sage,Yeshwanth_2023_ICCV,pfaff2026scenesmith,mao2026spatiallm,zheng2020structured3d}.
In total, we construct 3,581 tasks across 480 scenes (24 functional categories) and 1,203 object categories.



\subsection{Task-conditioned room generation}
\label{sec:TaskGenRoom}
\noindent\textbf{Evaluation protocol.}
\looseness=-1
Panel A schedules 150 tasks per method, 50 per difficulty.
Panel B uses 460 easy tasks in 30 sequences with scripted task
outcomes, retaining generated rooms after quality checks.
Methods share instructions, room types, and required objects.
Table~\ref{tab:task_readiness_expansion} reports percentages over all
scheduled tasks, with failed or unscored rooms counted as zero.
\emph{Task Ground.} measures required-object and
initial-relation coverage; \emph{Task Nav.} tests connectivity;
\emph{Physics Pass} reports geometric and access checks under
SceneEval~\cite{sceneeval}. \emph{Task-Ready} requires full grounding
and both passes. 

\noindent\textbf{Single-room results.}
\looseness=-1
WorldEdge reaches 86.7\% task grounding and 17.3\% Task-Ready;
all evaluated baseline methods have zero Task-Ready. Physics Pass
(30.7\%) remains a major bottleneck despite 99.3\% navigation.

\noindent\textbf{Sequential expansion results.}
\looseness=-1
WorldEdge reaches 98.6\% task grounding and 42.2\% Task-Ready.
HSM~\cite{pun2026hsm} achieves comparable grounding (98.5\%), yet both
HSM and EmbodiedGen V2~\cite{wang2026embodiedgen} have zero Task-Ready.
Object coverage alone is insufficient for full Task-Ready.
Holodeck~\cite{holodeck} fails floor-plan parsing; SAGE~\cite{sage}
remains unevaluated due to runtime and budget limits.

\subsection{Next-task proposal in expanding worlds}
\label{sec:next_task_proposal}

\begin{table}[!htbp]
\centering
\caption{\textbf{Next-task generation in expanding worlds.}
Symbolic results (mean $\pm$ 95\% CI).
$^\dagger$Adapted proposers.}
\label{tab:task_generation_expansion}
\footnotesize
\setlength{\tabcolsep}{0.5pt}
\renewcommand{\arraystretch}{1.15}
\newcommand{\ci}[1]{$_{\pm #1}$}
\begin{tabular*}{\columnwidth}{@{\extracolsep{\fill}}lccccc@{}}
\toprule
Proposer & Yield$\uparrow$ &
\shortstack{Prior\\effect$\uparrow$} &
\shortstack{Prior\\room$\uparrow$} &
\shortstack{Top\\object$\downarrow$} &
\shortstack{Distinct\\@16$\uparrow$} \\
\midrule
Template & \textbf{99.5}\ci{0.4} & 13.3\ci{3.6} & 23.8\ci{4.3} & 23.2\ci{11.6} & 5.2\ci{0.4} \\
\midrule
Direct prompt & 96.5\ci{1.1} & 57.8\ci{3.2} & 69.9\ci{3.2} & 14.7\ci{2.0} & 15.5\ci{0.1} \\
\;+ Aim & 94.9\ci{1.6} & 79.3\ci{2.7} & 85.7\ci{2.8} & 18.6\ci{1.2} & \textbf{15.7}\ci{0.1} \\
\;+ Task list~\cite{xie2026agentsynth} & 94.7\ci{1.4} & 93.3\ci{1.7} & 93.5\ci{1.7} & 27.8\ci{2.5} & 15.0\ci{0.3} \\
\midrule
Voyager$^\dagger$~\cite{wang2023voyager} & 88.1\ci{4.0} & 93.7\ci{2.0} & 95.6\ci{1.9} & 17.3\ci{1.6} & 13.8\ci{0.4} \\
OMNI-EPIC$^\dagger$~\cite{faldor2025omni} & 92.7\ci{1.9} & 96.0\ci{1.7} & 97.9\ci{1.6} & 21.9\ci{2.6} & 15.3\ci{0.3} \\
SEAgent$^\dagger$~\cite{sun2025seagent} & 86.7\ci{1.7} & 84.1\ci{3.9} & 89.3\ci{2.7} & 25.7\ci{0.9} & 14.7\ci{0.3} \\
\midrule
WorldEdge (Ours) & 98.9\ci{0.8} & \textbf{97.5}\ci{0.6} & \textbf{97.9}\ci{1.0} & \textbf{10.5}\ci{1.0} & 14.1\ci{0.2} \\
\bottomrule
\end{tabular*}
\end{table}

\begin{table}[!tbp]
\centering
\caption{\textbf{Layout evaluation on 150 scenes.} All methods are given the same object list; layouts are evaluated as bounding boxes with SceneEval~\cite{sceneeval}. Placed is the share of the given objects that a method positions; an unplaced object is put at the room centre and scored. I-Design returns layouts for 135 scenes: its scores are means over these, and Placed counts the objects of the other 15 as not placed. Mean scores (\%) and time (s).}
\label{tab:panel_a_sceneeval}
\fontsize{7}{8.5}\selectfont
\setlength{\tabcolsep}{1.35pt}
\renewcommand{\arraystretch}{1.1}
\resizebox{\columnwidth}{!}{%
\begin{tabular}{@{}lrrrrrrrr@{}}
\toprule
Method & Placed$\uparrow$ & COL$_{ob}\downarrow$ & COL$_{sc}\downarrow$ & SUP$\uparrow$ & NAV$\uparrow$ & ACC$\uparrow$ & OOB$\downarrow$ & Time \\
\midrule
LayoutGPT~\cite{layoutgpt}
& \textbf{100.00} & 22.15 & 68.00 & 93.16 & 96.89 & 78.82 & 0.09 & 32.69
\\
Holodeck~\cite{holodeck}
& 92.77 & 8.55 & 27.33 & 97.32 & 96.58 & 89.36 & 0.70 & 31.98
\\
I-Design~\cite{idesign}
& 71.65 & 29.14 & 77.04 & \textbf{99.66} & 96.65 & 82.20 & \textbf{0.00} & 348.58
\\
LayoutVLM~\cite{layoutvlm}
& \textbf{100.00} & 8.00 & 30.67 & 99.28 & 95.56 & 88.06 & 22.77 & 50.05
\\
DirectLayout~\cite{directlayout}
& \textbf{100.00} & 3.82 & 18.67 & 95.96 & 98.01 & 84.61 & \textbf{0.00} & 116.68
\\
OptiScene~\cite{optiscene}
& 99.87 & 69.93 & 98.00 & 92.02 & 96.70 & 72.02 & 25.16 & 45.16
\\
SAGE~\cite{sage}
& 89.79 & 15.47 & 56.00 & 97.68 & 94.88 & 84.66 & 3.06 & 53.89
\\
RoboLayout~\cite{robolayout}
& \textbf{100.00} & 12.37 & 51.33 & 92.31 & 96.44 & 89.58 & 25.01 & 58.74
\\
MANSION~\cite{Che_2026_CVPR}
& 87.51 & 15.75 & 56.00 & 97.29 & 94.71 & 85.80 & 0.47 & 19.88
\\
\midrule
WorldLayout
& 99.52 & \textbf{0.53} & \textbf{2.00} & 99.49 & \textbf{98.67} & \textbf{94.43} & \textbf{0.00} & 479.49
\\
\bottomrule
\end{tabular}%
}
\end{table}

\begin{figure}[!tbp]
\centering
\setlength{\unitlength}{0.30488\linewidth}
\begin{picture}(3.28,5.9839)
\put(0,0){\includegraphics[width=\linewidth]{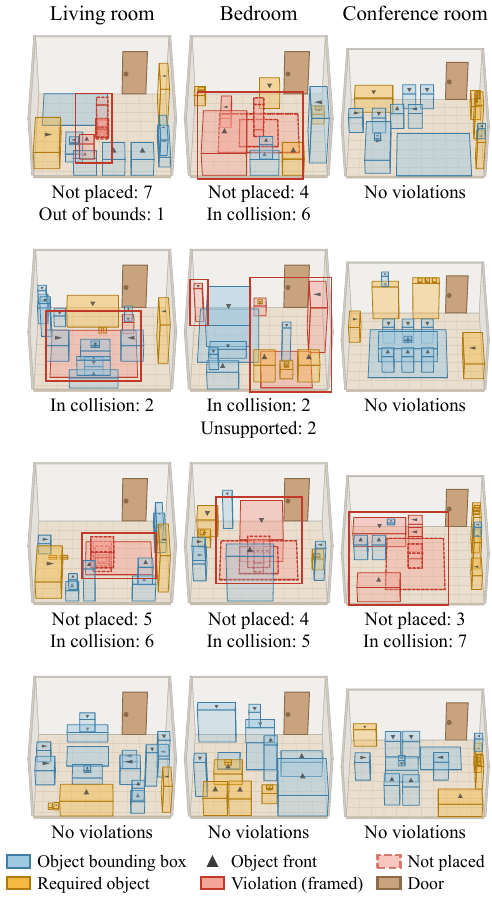}}
\put(0.0765,5.2722){\rotatebox{90}{\makebox(0,0){\small Holodeck~\cite{holodeck}}}}
\put(0.0765,3.8487){\rotatebox{90}{\makebox(0,0){\small DirectLayout~\cite{directlayout}}}}
\put(0.0765,2.4252){\rotatebox{90}{\makebox(0,0){\small MANSION~\cite{Che_2026_CVPR}}}}
\put(0.0765,1.0017){\rotatebox{90}{\makebox(0,0){\small \textbf{WorldLayout (Ours)}}}}
\end{picture}
\caption{\textbf{Layout comparison on three scenes.}
Yellow indicates required objects, red highlights violations, and triangles indicate object fronts. Objects that a method fails to place are shown with dashed outlines at the room centre, where they are scored.}
\label{fig:compare}
\end{figure}

\looseness=-1
Table~\ref{tab:task_generation_expansion} uses ten houses, three seeds, and
50 symbolic rounds with fixed tier schedules and competence estimates.
A reference solver commits verified completions; failed attempts discard
staged rooms and object changes. No meshes or robot execution are involved.
CIs use house-level means. Voyager~\cite{wang2023voyager},
OMNI-EPIC~\cite{faldor2025omni}, and SEAgent~\cite{sun2025seagent}
are adapted to these shared symbolic checks.
Direct prompt receives the current state and a diversity instruction.
Aim adds goals for history use and object balance;
Task list~\cite{xie2026agentsynth} further adds prior instructions and history queries.

\looseness=-1
Yield measures delivery per scheduled round. Other metrics use delivered
medium/hard-target tasks with applicable history and sample counts.
Prior effect measures dependence on earlier object
changes, Prior room measures earlier-room reuse, and Top object measures maximum
single-object concentration~\cite{berger1970diversity}. Distinct@16 counts
expected distinct structures in 16 samples without
replacement~\cite{hurlbert1971nonconcept}.
WorldEdge has high Yield and history dependence with the lowest
Top object (10.5\%). Template delivers slightly more tasks with less history
dependence and diversity. WorldEdge's Distinct@16 (14.1) trails
direct-prompt-plus-Aim (15.7); history dependence does not guarantee maximal diversity.



\subsection{Layout quality and result}
\label{sec:evaluate}

\looseness=-1
Table~\ref{tab:panel_a_sceneeval} reports SceneEval~\cite{sceneeval} scores.
Placed is the share of the given objects that a method positions.
COL$_{ob}$/COL$_{sc}$ measure object-/scene-level collision rates;
SUP is the supported-object rate, NAV the largest connected fraction of
free space, ACC functional-side accessibility, and OOB the out-of-bounds
object rate. Time is seconds per scene.

\textsc{WorldLayout} successfully places 99.52\% of the required objects with the lowest collision rates (0.53\%/2.00\%), and ranks first or second on every other physical score, but is slowest (479.49\,s). Four evaluated baselines place every object; the best baseline reaches 3.82\%/18.67\% collision (\textsc{DirectLayout}) and 89.58\% accessibility (\textsc{RoboLayout}). \textsc{WorldLayout} is the only evaluated method that places over 99\% of the objects with an object collision rate below 1\%. Baseline settings are given in the supplementary material.

Figure~\ref{fig:compare} shows three representative indoor scenes. Holodeck and MANSION leave
several objects unplaced, MANSION in all three scenes; DirectLayout places every
object, but some collide or remain physically unsupported. WorldLayout successfully places all required objects
in every scene without any detected violations.

\subsection{Articulation estimation}
\label{sec:art:main_experiments}

\begin{table}[!t]
\centering
\footnotesize
\setlength{\tabcolsep}{1.8pt}
\renewcommand{\arraystretch}{1.05}
\begin{tabular*}{\columnwidth}{@{\extracolsep{\fill}}lrrrrrr@{}}
\toprule
 & \multicolumn{3}{c}{(a) Rest state} & \multicolumn{3}{c}{(b) Articulated} \\
\cmidrule(lr){2-4}\cmidrule(l){5-7}
Method & gIoU$\uparrow$ & PC$\downarrow$ & mIoU$\uparrow$ & gIoU$\uparrow$ & PC$\downarrow$ & OC$\downarrow$ \\
\midrule
\multicolumn{7}{@{}l}{Lightwheel (241)} \\
PARTICULATE & 0.337 & 0.163 & 0.573 & 0.310 & 0.201 & \textbf{0.008} \\
\textbf{WorldArt} & \textbf{0.380} & \textbf{0.151} & \textbf{0.606} & \textbf{0.349} & \textbf{0.185} & 0.008 \\
\midrule
\multicolumn{7}{@{}l}{PartNet-Mobility (74)} \\
PARTICULATE & \textbf{0.884} & \textbf{0.001} & 0.886 & \textbf{0.845} & \textbf{0.005} & 0.001 \\
\textbf{WorldArt} & 0.882 & 0.003 & \textbf{0.888} & 0.840 & 0.007 & \textbf{0.001} \\
\midrule
\multicolumn{7}{@{}l}{WorldEdge-Art (141)} \\
PARTICULATE & 0.317 & 0.113 & 0.505 & 0.308 & 0.242 & 0.023 \\
\textbf{WorldArt} & \textbf{0.437} & \textbf{0.086} & \textbf{0.580} & \textbf{0.423} & \textbf{0.178} & \textbf{0.016} \\
\bottomrule
\end{tabular*}
\caption{\textbf{Articulation estimation across three evaluation datasets.}
PARTICULATE~\cite{li2026particulate}: officially released pretrained weights.
Bold: best unrounded means. Corrected motion results are reported in the supplement.}
\label{tab:art:main_results}
\end{table}

\looseness=-1
We evaluate articulation prediction before retrieval or critic
filtering on matched processed inputs across all three datasets
(Table~\ref{tab:art:main_results}). WorldArt raises rest-state
gIoU from 0.337 to 0.380 on Lightwheel~\cite{lightwheel2025simready} and from 0.317 to 0.437
on WorldEdge-Art. On WorldEdge-Art, articulated gIoU increases
substantially from 0.308 to 0.423, and articulated PC decreases from 0.242 to 0.178.

PartNet-Mobility~\cite{xiang2020sapien} shows no consistent gains. Articulated gIoU
uses the inherited box-motion approximation; corrected motion
diagnostics and evaluation details are given in the supplementary material.
Figure~\ref{fig:art:comparison} shows an ottoman example in which WorldArt
matches the lid geometry and opening motion more closely than PARTICULATE.

\begin{figure}[!tbp]
\centering
\includegraphics[width=\linewidth]{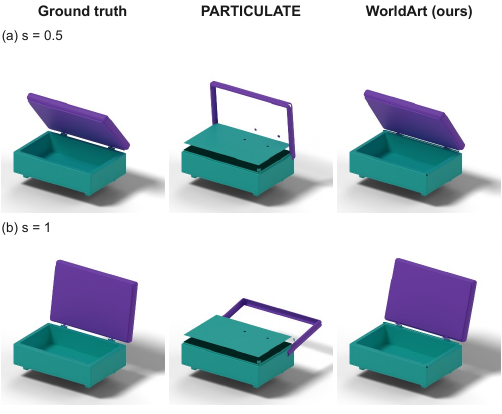}
\caption{\textbf{Articulation comparison on a storage ottoman.}
Visual comparison of ground truth, PARTICULATE, and WorldArt predictions at $s=0.5$ (top) and $s=1$ (bottom).}
\label{fig:art:comparison}
\end{figure}

\subsection{Downstream evaluation}
\label{sec:downstream}

We evaluate pretrained robotic policies and game world models without training, assessing sustained task completion and
action-conditioned scene consistency. An agentic pipeline provides a
reference for robotic execution.

\paragraph{Settings.}
\emph{Single room} uses 30 generated rooms, ten per difficulty tier, with
one task each. \emph{In the loop} executes the first ten tasks of each of
the 30 sequences in Sec.~\ref{sec:TaskGenRoom} (Panel~B), in order and without resets.
Rooms and object states persist after success or failure
(Sec.~\ref{sec:execution_persistence}).
Excluding initially satisfied or unverifiable goals leaves 149 in-the-loop
and 29 single-room tasks, evaluated by an independent simulator-state
checker.

\paragraph{Robot policies.}
We evaluate $\pi_{0.5}$~\cite{black2025pi05} on Lift2 and GR00T
N1.6/N1.7~\cite{bjorck2025gr00t}, OpenHLM~\cite{hu2026openhlm}, and
$\omega$-0~\cite{li2026omega0} on G1 with SONIC~\cite{luo2026sonic}.
All learned policies fail strict goal completion.
Even when success only requires moving the target object by at least
3\,cm, no learned policy exceeds 10.0\% SR@5 (Table~\ref{tab:robot}).
SR@5 requires success on all checkable tasks among the first five scheduled
tasks. Apparent object lifts mostly reflect tipping, exposing difficulties
with basic manipulation. In contrast, an LLM planner combined with robotic
skills achieves 80.0\% SR@5 under the same interaction criterion.
Evaluation details are in the supplement.

\paragraph{Agentic execution.}
An LLM planner calls navigation, pick, and place skills and replans from
skill outcomes. It achieves 60.4\% in-the-loop and 31.0\% single-room
success under strict goal completion, and 80.0\% SR@5 under the interaction
criterion. Most planner LLMs achieve similar success rates
(Table~\ref{tab:agentic_llms}). Navigation and placement remain the main
failure sources.


\begin{table}[t]
\centering
\caption{\textbf{Robot policy evaluation.}
Success requires target displacement of at least 3\,cm.
Rates are percentages.
Tasks/10 is a mean count.
$\dagger$: initialized at the task location.}
\label{tab:robot}
\footnotesize
\setlength{\tabcolsep}{2.5pt}
\begin{tabular}{@{}lcccc@{}}
\toprule
Policy & Task SR & SR@5 & Tasks/10 & Room SR \\
\midrule
$\pi_{0.5}$~\cite{black2025pi05}             & 0.7  & 0.0  & 0.0 & 0.0  \\
GR00T N1.6~\cite{bjorck2025gr00t}              & 22.1 & 6.7  & 1.1 & --   \\
GR00T N1.7$^{\dagger}$~\cite{bjorck2025gr00t}  & 34.2 & 10.0 & 1.7 & 22.2 \\
OpenHLM~\cite{hu2026openhlm}                 & 23.6 & 3.3  & 1.2 & 33.3 \\
$\omega$-0~\cite{li2026omega0}              & 13.8 & 6.7  & 0.6 & 22.2 \\
\midrule
Agentic (LLM + skills)  & 79.9 & 80.0 & 4.0 & 82.8 \\
\bottomrule
\end{tabular}
\end{table}

\begin{table}[t]
\centering
\caption{\textbf{LLM planner comparison.}
We vary the LLM with fixed worlds, prompts, skills, and closed-loop planning.
Evaluation details are given in the supplementary material.}
\label{tab:agentic_llms}
\footnotesize
\setlength{\tabcolsep}{3pt}
\resizebox{\columnwidth}{!}{%
\begin{tabular}{@{}lccccc@{}}
\toprule
LLM & Task SR & SR@5 & Tasks/10 & Interact & Calls \\
\midrule
GPT-6.1-sol~\cite{openai2026gpt61sol} (Ours) & 60.4 & 40.0 & 3.0 & 80.3 & 2.3 \\
\midrule
GPT-6.1-terra~\cite{openai2026gpt61terra} & 61.7 & 43.3 & 3.1 & 82.0 & 2.1 \\
GPT-5.6-sol~\cite{openai2026gpt56sol} & 60.4 & 33.3 & 3.0 & 79.2 & 2.2 \\
GLM-5.3~\cite{zeng2026glm} & 54.4 & 33.3 & 2.7 & 75.3 & 2.5 \\
DeepSeek-V4.1-Flash~\cite{xu2026deepseek} & 53.7 & 30.0 & 2.7 & 73.0 & 2.5 \\
\bottomrule
\end{tabular}}
\end{table}

\begin{figure}[t]
    \centering
    \includegraphics[width=\linewidth]{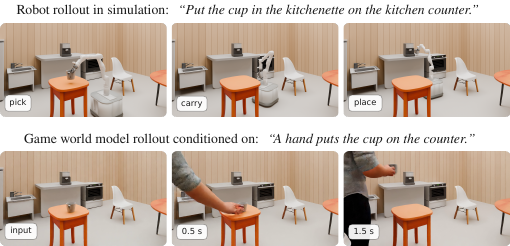}
    \caption{\textbf{Qualitative interaction rollouts.}
    Video world models generate text-conditioned object interactions in WorldEdge scenes.}
    \label{fig:exec}
\end{figure}

\paragraph{Video world models.}
We evaluate keyboard-controlled models on 60 rendered 30\,s single-room
walkthroughs, conditioned on the first frame and identical movement
commands (Table~\ref{tab:wm_game_v2}). Thresholds are calibrated on rendered
videos and fixed before evaluation.
LingBot-World~\cite{robbyant2026lingbot} achieves the highest action agreement (62.3\%), while Yume-1.5~\cite{mao2026yume1}
has the lowest 5\,s LPIPS~\cite{zhang2018unreasonable} and highest object recall (44.0\%, versus 83.9\%
on rendered references). All models introduce structures or people absent
from the scene.
Matrix-Game~2.0~\cite{he2025matrixgame2} obtains the lowest RotErr and MEt3R~\cite{asim2025met3r} despite the worst LPIPS
and object recall, consistent with limited camera motion.
Figure~\ref{fig:exec} shows qualitative text-conditioned interactions.

\begin{table}[t]
\centering
\caption{\textbf{Game world model evaluation.}
Results on 60 single-room walkthroughs with identical movement commands.
Action and recall are percentages.
RotErr is in degrees.}
\label{tab:wm_game_v2}
\setlength{\tabcolsep}{3pt}
\resizebox{\columnwidth}{!}{%
\begin{tabular}{@{}lccccccc@{}}
\toprule
 & \multicolumn{2}{c}{LPIPS~\cite{zhang2018unreasonable}} & Action & Rot & Trans & MEt3R~\cite{asim2025met3r} & Obj. \\
Model & 5\,s $\downarrow$ & 30\,s $\downarrow$ & \% $\uparrow$ & Err $\downarrow$ & Err $\downarrow$ & $\downarrow$ & recall $\uparrow$ \\
\midrule
Matrix-Game 2.0~\cite{he2025matrixgame2}             & 0.755 & 0.817 & 37.3 & 22.4 & 0.74 & 0.142 & 24.6 \\
HY-WorldPlay~\cite{sun2026worldplay}                & 0.711 & 0.685 & 52.5 & 51.1 & 0.69 & 0.186 & 39.4 \\
LingBot-World~\cite{robbyant2026lingbot}               & 0.696 & 0.733 & 62.3 & 28.5 & 0.57 & 0.200 & 31.0 \\
Yume-1.5~\cite{mao2026yume1}                    & 0.622 & 0.685 & 28.2 & 47.5 & 0.80 & 0.169 & 44.0 \\
\bottomrule
\end{tabular}}
\end{table}

\section{Conclusion}
\label{sec:conclusion}

We present \method{}, a framework for persistent task-driven expansion of
indoor worlds. History-conditioned tasks guide connected room synthesis,
geometric verification, and task-required articulation. Expansion preserves
object identities and inherited states, while execution feedback supports
subsequent tasks that reference existing objects and prior interaction
effects. Experiments evaluate room task readiness, symbolic task
continuation, and downstream robot and video models.

\paragraph{Limitations and future work.}
We evaluate scripted room expansion, symbolic sequences, and simulated
robot rollouts. Feedback-driven task selection and room growth need
evaluation. Task readiness and pretrained manipulation remain limited,
and articulation gains vary by source. Future work includes longer
closed-loop runs, geometry and motion checks, broader skills, and
physical robot tests of state persistence and task continuation.

{
    \small
    \bibliographystyle{ieeenat_fullname}
    \bibliography{main}

\begin{thebibliography}{122}
\providecommand{\natexlab}[1]{#1}
\providecommand{\url}[1]{\texttt{#1}}
\expandafter\ifx\csname urlstyle\endcsname\relax
  \providecommand{\doi}[1]{doi: #1}\else
  \providecommand{\doi}{doi: \begingroup \urlstyle{rm}\Url}\fi

\bibitem[Asim et~al.(2025)Asim, Wewer, Wimmer, Schiele, and
  Lenssen]{asim2025met3r}
Mohammad Asim, Christopher Wewer, Thomas Wimmer, Bernt Schiele, and Jan~Eric
  Lenssen.
\newblock Met3r: Measuring multi-view consistency in generated images.
\newblock In \emph{2025 IEEE/CVF Conference on Computer Vision and Pattern
  Recognition (CVPR)}, pages 6034--6044. IEEE, 2025.

\bibitem[Avetisyan et~al.(2024)Avetisyan, Xie, Howard-Jenkins, Yang, Aroudj,
  Patra, Zhang, Frost, Holland, Orme, Engel, Miller, Newcombe, and
  Balntas]{avetisyan2024scenescript}
Armen Avetisyan, Christopher Xie, Henry Howard-Jenkins, Tsun-Yi Yang, Samir
  Aroudj, Suvam Patra, Fuyang Zhang, Duncan Frost, Luke Holland, Campbell Orme,
  Jakob Engel, Edward Miller, Richard Newcombe, and Vasileios Balntas.
\newblock Scenescript: Reconstructing scenes with an autoregressive structured
  language model.
\newblock In \emph{European Conference on Computer Vision (ECCV)}, 2024.

\bibitem[Bai et~al.(2025)Bai, Chen, Liu, Wang, Ge, Song, Dang, Wang, Wang,
  Tang, Zhong, Zhu, Yang, Li, Wan, Wang, Ding, Fu, Xu, Ye, Zhang, Xie, Cheng,
  Zhang, Yang, Xu, and Lin]{Qwen2.5-VL}
Shuai Bai, Keqin Chen, Xuejing Liu, Jialin Wang, Wenbin Ge, Sibo Song, Kai
  Dang, Peng Wang, Shijie Wang, Jun Tang, Humen Zhong, Yuanzhi Zhu, Mingkun
  Yang, Zhaohai Li, Jianqiang Wan, Pengfei Wang, Wei Ding, Zheren Fu, Yiheng
  Xu, Jiabo Ye, Xi Zhang, Tianbao Xie, Zesen Cheng, Hang Zhang, Zhibo Yang,
  Haiyang Xu, and Junyang Lin.
\newblock Qwen2.5-vl technical report.
\newblock \emph{arXiv preprint arXiv:2502.13923}, 2025.

\bibitem[Berger and Parker(1970)]{berger1970diversity}
Wolfgang~H Berger and Frances~L Parker.
\newblock Diversity of planktonic foraminifera in deep-sea sediments.
\newblock \emph{Science}, 168\penalty0 (3937):\penalty0 1345--1347, 1970.

\bibitem[Bjorck et~al.(2025)Bjorck, Casta{\~n}eda, Cherniadev, Da, Ding, Fan,
  Fang, Fox, Hu, Huang, et~al.]{bjorck2025gr00t}
Johan Bjorck, Fernando Casta{\~n}eda, Nikita Cherniadev, Xingye Da, Runyu Ding,
  Linxi Fan, Yu Fang, Dieter Fox, Fengyuan Hu, Spencer Huang, et~al.
\newblock Gr00t n1: An open foundation model for generalist humanoid robots.
\newblock \emph{arXiv preprint arXiv:2503.14734}, 2025.

\bibitem[Black et~al.(2025)Black, Brown, Darpinian, Dhabalia, Driess, Esmail,
  Equi, Finn, Fusai, Galliker, Ghosh, Groom, Hausman, et~al.]{black2025pi05}
Kevin Black, Noah Brown, James Darpinian, Karan Dhabalia, Danny Driess, Adnan
  Esmail, Michael~Robert Equi, Chelsea Finn, Niccolo Fusai, Manuel~Y. Galliker,
  Dibya Ghosh, Lachy Groom, Karol Hausman, et~al.
\newblock $\pi_{0.5}$: A vision-language-action model with open-world
  generalization.
\newblock In \emph{Proceedings of the Conference on Robot Learning (CoRL)},
  pages 17--40. PMLR, 2025.

\bibitem[Bucher and Armeni(2025)]{respace}
Martin~JJ Bucher and Iro Armeni.
\newblock Respace: Text-driven autoregressive 3d indoor scene synthesis and
  editing.
\newblock \emph{arXiv preprint arXiv:2506.02459}, 2025.

\bibitem[Chang et~al.(2025)Chang, Chhablani, Clegg, Dallaire~Cote, Desai,
  Hlavac, Karashchuk, Krantz, Mottaghi, Parashar, Patki, Prasad, Puig, Rai,
  Ramrakhya, Tran, Truong, Turner, Undersander, and Yang]{ICLR2025_a3cf318f}
Matthew Chang, Gunjan Chhablani, Alexander Clegg, Mikael Dallaire~Cote, Ruta
  Desai, Michal Hlavac, Vladimir Karashchuk, Jacob Krantz, Roozbeh Mottaghi,
  Priyam Parashar, Siddharth Patki, Ishita Prasad, Xavier Puig, Akshara Rai,
  Ram Ramrakhya, Daniel Tran, Joanne Truong, John Turner, Eric Undersander, and
  Tsung-Yen Yang.
\newblock Partnr: A benchmark for planning and reasoning in embodied
  multi-agent tasks.
\newblock In \emph{International Conference on Learning Representations}, pages
  65205--65268, 2025.

\bibitem[Che et~al.(2026)Che, Wen, Huang, Wang, Yang, Dudek, Wang, and
  Su]{Che_2026_CVPR}
Lirong Che, Shuo Wen, Shan Huang, Chuang Wang, Yuzhe Yang, Gregory Dudek,
  Xueqian Wang, and Jian Su.
\newblock Mansion: Multi-floor language-to-3d scene generation for long-horizon
  tasks.
\newblock In \emph{Proceedings of the IEEE/CVF Conference on Computer Vision
  and Pattern Recognition (CVPR)}, pages 37175--37185, 2026.

\bibitem[Chen et~al.(2026)Chen, Chu, Gleize, Liang, Sax, Tang, Wang, Guo,
  Hardin, Li, et~al.]{sam3d}
Xingyu Chen, Fu-Jen Chu, Pierre Gleize, Kevin~J Liang, Alexander Sax, Hao Tang,
  Weiyao Wang, Michelle Guo, Thibaut Hardin, Xiang Li, et~al.
\newblock Sam 3d: 3dfy anything in images.
\newblock In \emph{Proceedings of the IEEE/CVF Conference on Computer Vision
  and Pattern Recognition}, pages 7220--7232, 2026.

\bibitem[Dai et~al.(2025)Dai, Wang, Wan, and Su]{dai2025manitaskgen}
Liu Dai, Haina Wang, Weikang Wan, and Hao Su.
\newblock Manitaskgen: A comprehensive task generator for benchmarking and
  improving vision-language agents on embodied decision-making.
\newblock \emph{arXiv preprint arXiv:2505.20726}, 2025.

\bibitem[Deitke et~al.(2022)Deitke, VanderBilt, Herrasti, Weihs, Ehsani,
  Salvador, Han, Kolve, Kembhavi, and Mottaghi]{deitke2022}
Matt Deitke, Eli VanderBilt, Alvaro Herrasti, Luca Weihs, Kiana Ehsani, Jordi
  Salvador, Winson Han, Eric Kolve, Aniruddha Kembhavi, and Roozbeh Mottaghi.
\newblock Procthor: Large-scale embodied ai using procedural generation.
\newblock In \emph{Advances in Neural Information Processing Systems}, pages
  5982--5994. Curran Associates, Inc., 2022.

\bibitem[Deitke et~al.(2023)Deitke, Schwenk, Salvador, Weihs, Michel,
  VanderBilt, Schmidt, Ehsani, Kembhavi, and Farhadi]{deitke2023objaverse}
Matt Deitke, Dustin Schwenk, Jordi Salvador, Luca Weihs, Oscar Michel, Eli
  VanderBilt, Ludwig Schmidt, Kiana Ehsani, Aniruddha Kembhavi, and Ali
  Farhadi.
\newblock Objaverse: A universe of annotated 3d objects.
\newblock In \emph{Proceedings of the IEEE/CVF conference on computer vision
  and pattern recognition}, pages 13142--13153, 2023.

\bibitem[Deng et~al.(2025)Deng, Qi, and Ma]{gltreesearch}
Wei Deng, Mengshi Qi, and Huadong Ma.
\newblock Global-local tree search in vlms for 3d indoor scene generation.
\newblock In \emph{2025 IEEE/CVF Conference on Computer Vision and Pattern
  Recognition (CVPR)}, pages 8975--8984. IEEE, 2025.

\bibitem[Duan et~al.(2025)Duan, Yu, Chen, Fei-Fei, and Wu]{duan2025worldscore}
Haoyi Duan, Hong-Xing Yu, Sirui Chen, Li Fei-Fei, and Jiajun Wu.
\newblock Worldscore: A unified evaluation benchmark for world generation.
\newblock In \emph{Proceedings of the IEEE/CVF international conference on
  computer vision}, pages 27713--27724, 2025.

\bibitem[Faldor et~al.(2025)Faldor, Zhang, Cully, and Clune]{faldor2025omni}
Maxence Faldor, Jenny Zhang, Antoine Cully, and Jeff Clune.
\newblock Omni-epic: Open-endedness via models of human notions of
  interestingness with environments programmed in code.
\newblock In \emph{International Conference on Learning Representations}, pages
  85260--85385, 2025.

\bibitem[Fang et~al.(2024)Fang, Madappally~Jose, Jain, Schmidt, Toshev, and
  Shankar]{fang2024data}
Alex Fang, Albin Madappally~Jose, Amit Jain, Ludwig Schmidt, Alexander Toshev,
  and Vaishaal Shankar.
\newblock Data filtering networks.
\newblock In \emph{International Conference on Learning Representations}, pages
  36221--36237, 2024.

\bibitem[Feng et~al.(2023)Feng, Zhu, Fu, Jampani, Akula, He, Basu, Wang, and
  Wang]{layoutgpt}
Weixi Feng, Wanrong Zhu, Tsu-jui Fu, Varun Jampani, Arjun Akula, Xuehai He,
  Sugato Basu, Xin~Eric Wang, and William~Yang Wang.
\newblock Layoutgpt: Compositional visual planning and generation with large
  language models.
\newblock \emph{Advances in neural information processing systems},
  36:\penalty0 18225--18250, 2023.

\bibitem[Fu et~al.(2021)Fu, Cai, Gao, Zhang, Wang, Li, Zeng, Sun, Jia, Zhao,
  and Zhang]{Fu_2021_ICCV}
Huan Fu, Bowen Cai, Lin Gao, Ling-Xiao Zhang, Jiaming Wang, Cao Li, Qixun Zeng,
  Chengyue Sun, Rongfei Jia, Binqiang Zhao, and Hao Zhang.
\newblock 3d-front: 3d furnished rooms with layouts and semantics.
\newblock In \emph{Proceedings of the IEEE/CVF International Conference on
  Computer Vision (ICCV)}, pages 10933--10942, 2021.

\bibitem[Gao et~al.(2026)Gao, Zheng, Gao, Ma, Wang, Wang, Chen, Chen, Zhang,
  Jia, et~al.]{gao2026ebench}
Ning Gao, Jinliang Zheng, Xing Gao, Haoxiang Ma, Hanqing Wang, Yukai Wang,
  Jiantong Chen, Zanxin Chen, Shujie Zhang, Mingda Jia, et~al.
\newblock Ebench: Elemental diagnosis of generalist mobile manipulation
  policies.
\newblock \emph{arXiv preprint arXiv:2606.18239}, 2026.

\bibitem[Guo et~al.(2026)Guo, Shi, Chen, and Finn]{guo2026ctrlworld}
Yanjiang Guo, Lucy Shi, Jianyu Chen, and Chelsea Finn.
\newblock Ctrl-world: A controllable generative world model for robot
  manipulation.
\newblock In \emph{International Conference on Learning Representations}, pages
  6121--6138, 2026.

\bibitem[He et~al.(2025)He, Peng, Liu, Wang, Zhang, Cui, Kang, Jiang, An, Ren,
  et~al.]{he2025matrixgame2}
Xianglong He, Chunli Peng, Zexiang Liu, Boyang Wang, Yifan Zhang, Qi Cui, Fei
  Kang, Biao Jiang, Mengyin An, Yangyang Ren, et~al.
\newblock Matrix-game 2.0: An open-source, real-time, and streaming interactive
  world model.
\newblock \emph{arXiv preprint arXiv:2508.13009}, 2025.

\bibitem[Hu et~al.(2026{\natexlab{a}})Hu, Arroyo, Debats, Manhardt, Carlone,
  and Tombari]{hu2026mixed}
Siyi Hu, Diego~Martin Arroyo, Stephanie Debats, Fabian Manhardt, Luca Carlone,
  and Federico Tombari.
\newblock Mixed diffusion for 3d indoor scene synthesis.
\newblock In \emph{2026 IEEE/CVF Winter Conference on Applications of Computer
  Vision (WACV)}, pages 1262--1272. IEEE, 2026{\natexlab{a}}.

\bibitem[Hu et~al.(2026{\natexlab{b}})Hu, Zhu, Zheng, Hu, Zhang, Chen, Zhao,
  Nai, and Gao]{hu2026openhlm}
Yingdong Hu, Haodong Zhu, Boyuan Zheng, Yihang Hu, Tong Zhang, Zunhao Chen,
  Junming Zhao, Ruiqian Nai, and Yang Gao.
\newblock {OpenHLM}: An empirical recipe for whole-body humanoid
  loco-manipulation.
\newblock In \emph{Proceedings of the Conference on Robot Learning (CoRL)},
  2026{\natexlab{b}}.

\bibitem[Hunyuan3D et~al.(2025)Hunyuan3D, Yang, Yang, Feng, Huang, Zhang, He,
  Luo, Liu, Zhao, et~al.]{hunyuan3d2025hunyuan3d}
Team Hunyuan3D, Shuhui Yang, Mingxin Yang, Yifei Feng, Xin Huang, Sheng Zhang,
  Zebin He, Di Luo, Haolin Liu, Yunfei Zhao, et~al.
\newblock Hunyuan3d 2.1: From images to high-fidelity 3d assets with
  production-ready pbr material.
\newblock \emph{arXiv preprint arXiv:2506.15442}, 2025.

\bibitem[Hurlbert(1971)]{hurlbert1971nonconcept}
Stuart~H Hurlbert.
\newblock The nonconcept of species diversity: a critique and alternative
  parameters.
\newblock \emph{Ecology}, 52\penalty0 (4):\penalty0 577--586, 1971.

\bibitem[Ilharco et~al.(2021)Ilharco, Wortsman, Carlini, Taori, Dave, Shankar,
  Namkoong, Miller, Hajishirzi, Farhadi, et~al.]{ilharco2021openclip}
Gabriel Ilharco, Mitchell Wortsman, Nicholas Carlini, Rohan Taori, Achal Dave,
  Vaishaal Shankar, Hongseok Namkoong, John Miller, Hannaneh Hajishirzi, Ali
  Farhadi, et~al.
\newblock Openclip.
\newblock \emph{Zenodo}, 2021.

\bibitem[Jiang et~al.(2026)Jiang, Atashgahi, Delgado, Brokkelkamp, Zanutto,
  S{\"o}kmen, and Shahkarami]{polylayout}
Yutong Jiang, Zahra Atashgahi, Carlos Soto~Garcia Delgado, Ruben Brokkelkamp,
  Davide Zanutto, Ef{\c{s}}an S{\"o}kmen, and Shahin Shahkarami.
\newblock Polylayout: Hierarchical vlm-guided layout generation beyond
  rectangular rooms.
\newblock \emph{arXiv preprint arXiv:2608.10838}, 2026.

\bibitem[Jin et~al.(2026)Jin, Che, Li, Zhao, Wu, Zhang, Zhao, Liu, Zhang, Ju,
  et~al.]{jin2026artvip}
Zhao Jin, Zhengping Che, Tao Li, Zhen Zhao, Kun Wu, Yuheng Zhang, Yinuo Zhao,
  Zehui Liu, Qiang Zhang, Xiaozhu Ju, et~al.
\newblock Artvip: Articulated digital assets of visual realism, modular
  interaction, and physical fidelity for robot learning.
\newblock In \emph{International Conference on Learning Representations}, pages
  75272--75301, 2026.

\bibitem[Jun and Nichol(2023)]{jun2023shap}
Heewoo Jun and Alex Nichol.
\newblock Shap-e: Generating conditional 3d implicit functions.
\newblock \emph{arXiv preprint arXiv:2305.02463}, 2023.

\bibitem[Kang et~al.(2026)Kang, Ye, Liu, Mantri, Mao, Fleming, Regmi, and
  Qin]{kang2026simworld}
Haoqiang Kang, Xiaokang Ye, Yuhan Liu, Siddhant~Hitesh Mantri, Lingjun Mao,
  James Fleming, Drishti Regmi, and Lianhui Qin.
\newblock Simworld studio: Automatic environment generation with evolving
  coding agent for embodied agent learning.
\newblock In \emph{Advances in Neural Information Processing Systems}, 2026.

\bibitem[Kerbl et~al.(2023)Kerbl, Kopanas, Leimk{\"u}hler, Drettakis,
  et~al.]{kerbl20233d}
Bernhard Kerbl, Georgios Kopanas, Thomas Leimk{\"u}hler, George Drettakis,
  et~al.
\newblock 3d gaussian splatting for real-time radiance field rendering.
\newblock \emph{ACM Trans. Graph.}, 42\penalty0 (4):\penalty0 139--1, 2023.

\bibitem[Khanna et~al.(2024)Khanna, Mao, Jiang, Haresh, Shacklett, Batra,
  Clegg, Undersander, Chang, and Savva]{Khanna_2024_CVPR}
Mukul Khanna, Yongsen Mao, Hanxiao Jiang, Sanjay Haresh, Brennan Shacklett,
  Dhruv Batra, Alexander Clegg, Eric Undersander, Angel~X. Chang, and Manolis
  Savva.
\newblock Habitat synthetic scenes dataset (hssd-200): An analysis of 3d scene
  scale and realism tradeoffs for objectgoal navigation.
\newblock In \emph{Proceedings of the IEEE/CVF Conference on Computer Vision
  and Pattern Recognition (CVPR)}, pages 16384--16393, 2024.

\bibitem[Kolve et~al.(2017)Kolve, Mottaghi, Han, VanderBilt, Weihs, Herrasti,
  Deitke, Ehsani, Gordon, Zhu, et~al.]{kolve2017ai2}
Eric Kolve, Roozbeh Mottaghi, Winson Han, Eli VanderBilt, Luca Weihs, Alvaro
  Herrasti, Matt Deitke, Kiana Ehsani, Daniel Gordon, Yuke Zhu, et~al.
\newblock Ai2-thor: An interactive 3d environment for visual ai.
\newblock \emph{arXiv preprint arXiv:1712.05474}, 2017.

\bibitem[Kuhn(1955)]{kuhn1955hungarian}
Harold~W Kuhn.
\newblock The hungarian method for the assignment problem.
\newblock \emph{Naval research logistics quarterly}, 2\penalty0 (1-2):\penalty0
  83--97, 1955.

\bibitem[Lara et~al.(2026)Lara, Milios, Luo, Sharma, Luo, Beckham, Golemo, and
  Pal]{fprlvr}
Luis Lara, Aristides Milios, Zhihao Luo, Aditya Sharma, Ge~Ya Luo, Christopher
  Beckham, Florian Golemo, and Christopher Pal.
\newblock Generative floor plan design with llms via reinforcement learning
  with verifiable rewards.
\newblock In \emph{Findings of the Association for Computational Linguistics:
  ACL 2026}, pages 26612--26627, 2026.

\bibitem[Li et~al.(2026{\natexlab{a}})Li, Yao, Zheng, Rupprecht, Lasenby, Wu,
  and Vedaldi]{li2026particulate}
Ruining Li, Yuxin Yao, Chuanxia Zheng, Christian Rupprecht, Joan Lasenby,
  Shangzhe Wu, and Andrea Vedaldi.
\newblock Particulate: Feed-forward 3d object articulation.
\newblock In \emph{Proceedings of the IEEE/CVF Conference on Computer Vision
  and Pattern Recognition}, pages 27708--27718, 2026{\natexlab{a}}.

\bibitem[Li et~al.(2026{\natexlab{b}})Li, Yao, Zhou, Zheng, Rupprecht, Lasenby,
  Wu, and Vedaldi]{li2026instruct}
Ruining Li, Yuxin Yao, Matt Zhou, Chuanxia Zheng, Christian Rupprecht, Joan
  Lasenby, Shangzhe Wu, and Andrea Vedaldi.
\newblock Instruct-particulate: Scaling feed-forward 3d object articulation
  with kinematic control.
\newblock \emph{arXiv preprint arXiv:2606.14699}, 2026{\natexlab{b}}.

\bibitem[Li et~al.(2026{\natexlab{c}})Li, Yang, Fang, Yi, Lu, Cen, Xie, Shen,
  and Tian]{li2026worldgrow}
Sikuang Li, Chen Yang, Jiemin Fang, Taoran Yi, Jia Lu, Jiazhong Cen, Lingxi
  Xie, Wei Shen, and Qi Tian.
\newblock {WorldGrow}: Generating infinite 3d world.
\newblock \emph{Proceedings of the AAAI Conference on Artificial Intelligence},
  40\penalty0 (8):\penalty0 6433--6441, 2026{\natexlab{c}}.

\bibitem[Li et~al.(2020)Li, Wang, Yi, Guibas, Abbott, and Song]{li2020category}
Xiaolong Li, He Wang, Li Yi, Leonidas~J Guibas, A~Lynn Abbott, and Shuran Song.
\newblock Category-level articulated object pose estimation.
\newblock In \emph{Proceedings of the IEEE/CVF conference on computer vision
  and pattern recognition}, pages 3706--3715, 2020.

\bibitem[Li et~al.(2025)Li, Hsu, Gu, Mees, Pertsch, Walke,
  et~al.]{li2025simpler}
Xuanlin Li, Kyle Hsu, Jiayuan Gu, Oier Mees, Karl Pertsch, Homer~Rich Walke,
  et~al.
\newblock Evaluating real-world robot manipulation policies in simulation.
\newblock In \emph{Proceedings of the Conference on Robot Learning (CoRL)},
  pages 3705--3728, 2025.

\bibitem[Li et~al.(2026{\natexlab{d}})Li, Zhang, Wei, Zhang, Yuan, Zhi, Li,
  Guo, Gao, Yang, and Zhang]{li2026omega0}
Zhe Li, Zhenzhe Zhang, Yangyang Wei, Wenjie Zhang, Xichen Yuan, Peiyuan Zhi,
  Gen Li, Xinying Guo, Fengjie Gao, Jianfei Yang, and Shanghang Zhang.
\newblock {$\omega$-0: A Latent Predictive World Action Model for Concurrent
  Humanoid Loco-Manipulation}.
\newblock \emph{arXiv preprint arXiv:2608.06375}, 2026{\natexlab{d}}.

\bibitem[Liang et~al.(2024)Liang, Wang, Wang, Bastani, Jayaraman, and
  Ma]{liang2024eurekaverse}
William Liang, Sam Wang, Hung-Ju Wang, Osbert Bastani, Dinesh Jayaraman, and
  Yecheng~Jason Ma.
\newblock Environment curriculum generation via large language models.
\newblock In \emph{Conference on Robot Learning (CoRL)}, 2024.

\bibitem[{Lightwheel}(2025)]{lightwheel2025simready}
{Lightwheel}.
\newblock {SimReady}: Simulation-ready {3D} assets, 2025.
\newblock Accessed: 2026.

\bibitem[Lin and Mu(2024)]{lin2024instructscene}
Chenguo Lin and Yadong Mu.
\newblock Instructscene: Instruction-driven 3d indoor scene synthesis with
  semantic graph prior.
\newblock In \emph{International Conference on Learning Representations}, pages
  25687--25718, 2024.

\bibitem[Littlefair et~al.(2025)Littlefair, Dutt, and Mitra]{flairgpt}
Gabrielle Littlefair, Niladri~Shekhar Dutt, and Niloy~J Mitra.
\newblock Flairgpt: Repurposing llms for interior designs.
\newblock In \emph{Computer graphics forum}, page e70036. Wiley Online Library,
  2025.

\bibitem[Liu et~al.(2025)Liu, Uy, Xiang, Su, Fidler, Sharp, and
  Gao]{liu2025partfield}
Minghua Liu, Mikaela~Angelina Uy, Donglai Xiang, Hao Su, Sanja Fidler, Nicholas
  Sharp, and Jun Gao.
\newblock Partfield: Learning 3d feature fields for part segmentation and
  beyond.
\newblock In \emph{2025 IEEE/CVF International Conference on Computer Vision
  (ICCV)}, pages 9704--9715. IEEE, 2025.

\bibitem[Loshchilov and Hutter(2019)]{loshchilov2019decoupled}
Ilya Loshchilov and Frank Hutter.
\newblock Decoupled weight decay regularization.
\newblock In \emph{International Conference on Learning Representations}, 2019.

\bibitem[Luo et~al.(2026)Luo, Yuan, Wang, Li, Casta{\~n}eda, Chen, Cao, Li,
  Minor, Ben, et~al.]{luo2026sonic}
Zhengyi Luo, Ye Yuan, Tingwu Wang, Chenran Li, Fernando Casta{\~n}eda, Sirui
  Chen, Zi-Ang Cao, Jiefeng Li, David Minor, Qingwei Ben, et~al.
\newblock Sonic: Supersizing motion tracking for natural humanoid whole-body
  control.
\newblock \emph{Science Robotics}, 11\penalty0 (117):\penalty0 eaed4592, 2026.

\bibitem[Mao et~al.(2026)Mao, Li, Li, Xu, Ying, and Zhang]{mao2026yume1}
Xiaofeng Mao, Zhen Li, Chuanhao Li, Xiaojie Xu, Kaining Ying, and Kaipeng
  Zhang.
\newblock Yume1. 5: A text-controlled interactive world generation model.
\newblock In \emph{Proceedings of the IEEE/CVF Conference on Computer Vision
  and Pattern Recognition}, pages 7752--7761, 2026.

\bibitem[Mao et~al.(2022)Mao, Zhang, Jiang, Chang, and Savva]{mao2022multiscan}
Yongsen Mao, Yiming Zhang, Hanxiao Jiang, Angel Chang, and Manolis Savva.
\newblock Multiscan: Scalable rgbd scanning for 3d environments with
  articulated objects.
\newblock In \emph{Advances in Neural Information Processing Systems}, pages
  9058--9071. Curran Associates, Inc., 2022.

\bibitem[Mao et~al.(2025)Mao, Zhong, Fang, Zheng, Tang, Zhu, Tan, and
  Zhou]{mao2026spatiallm}
Yongsen Mao, Junhao Zhong, Chuan Fang, Jia Zheng, Rui Tang, Hao Zhu, Ping Tan,
  and Zihan Zhou.
\newblock Spatiallm: Training large language models for structured indoor
  modeling.
\newblock In \emph{Advances in Neural Information Processing Systems}, pages
  45165--45195. Curran Associates, Inc., 2025.

\bibitem[Miao et~al.(2026)Miao, Wei, Ge, sun, Gao, Zhu, Wang, Tang, Xiao, Tang,
  and Li]{miao2026towards}
Bingchen Miao, Rong Wei, Zhiqi Ge, Xiaoquan sun, Shiqi Gao, Jingzhe Zhu, Renhan
  Wang, Siliang Tang, Jun Xiao, Rui Tang, and Juncheng Li.
\newblock Towards physically executable 3d gaussian for embodied navigation.
\newblock In \emph{International Conference on Learning Representations}, pages
  22945--22965, 2026.

\bibitem[Minderer et~al.(2023)Minderer, Gritsenko, and
  Houlsby]{minderer2023scaling}
Matthias Minderer, Alexey Gritsenko, and Neil Houlsby.
\newblock Scaling open-vocabulary object detection.
\newblock \emph{Advances in Neural Information Processing Systems},
  36:\penalty0 72983--73007, 2023.

\bibitem[Noh et~al.(2026)Noh, Seo, Park, and Kang]{Noh_2026_CVPR}
Seongrae Noh, SeungWon Seo, Gyeong-Moon Park, and HyeongYeop Kang.
\newblock Edit-as-act: Goal-regressive planning for open-vocabulary 3d indoor
  scene editing.
\newblock In \emph{Proceedings of the IEEE/CVF Conference on Computer Vision
  and Pattern Recognition (CVPR)}, pages 19864--19873, 2026.

\bibitem[{\"O}cal et~al.(2024){\"O}cal, Tatarchenko, Karao{\u{g}}lu, and
  Gevers]{sceneteller}
Ba{\c{s}}ak~Melis {\"O}cal, Maxim Tatarchenko, Sezer Karao{\u{g}}lu, and Theo
  Gevers.
\newblock Sceneteller: Language-to-3d scene generation.
\newblock In \emph{European Conference on Computer Vision}, pages 362--378.
  Springer, 2024.

\bibitem[{OpenAI}(2025)]{openai_2025}
{OpenAI}.
\newblock Introducing codex.
\newblock OpenAI Blog, 2025.

\bibitem[{OpenAI}(2026{\natexlab{a}})]{openai2026gpt56sol}
{OpenAI}.
\newblock {GPT-5.6 Sol}.
\newblock OpenAI API Model Documentation, 2026{\natexlab{a}}.

\bibitem[{OpenAI}(2026{\natexlab{b}})]{openai2026gpt61sol}
{OpenAI}.
\newblock Introducing {GPT-6.1 Sol}.
\newblock OpenAI Blog, 2026{\natexlab{b}}.

\bibitem[{OpenAI}(2026{\natexlab{c}})]{openai2026gpt61terra}
{OpenAI}.
\newblock {GPT-6.1 Terra}.
\newblock OpenAI API Model Documentation, 2026{\natexlab{c}}.

\bibitem[Pfaff et~al.(2026)Pfaff, Cohn, Zakharov, Cory, and
  Tedrake]{pfaff2026scenesmith}
Nicholas Pfaff, Thomas Cohn, Sergey Zakharov, Rick Cory, and Russ Tedrake.
\newblock Scenesmith: Agentic generation of simulation-ready indoor scenes.
\newblock In \emph{Forty-third International Conference on Machine Learning},
  2026.

\bibitem[Pun et~al.(2026)Pun, Tam, Wang, Huo, Chang, and Savva]{pun2026hsm}
Hou In~Derek Pun, Hou In~Ivan Tam, Austin~T Wang, Xiaoliang Huo, Angel~X Chang,
  and Manolis Savva.
\newblock Hsm: Hierarchical scene motifs for multi-scale indoor scene
  generation.
\newblock In \emph{2026 International Conference on 3D Vision (3DV)}, pages
  1356--1367. IEEE, 2026.

\bibitem[Qin et~al.(2020)Qin, Zhang, Huang, Dehghan, Zaiane, and
  Jagersand]{qin2020u2}
Xuebin Qin, Zichen Zhang, Chenyang Huang, Masood Dehghan, Osmar~R Zaiane, and
  Martin Jagersand.
\newblock U2-net: Going deeper with nested u-structure for salient object
  detection.
\newblock \emph{Pattern recognition}, 106:\penalty0 107404, 2020.

\bibitem[Qin et~al.(2026)Qin, Deng, Liang, Song, Lin, Jiang, Fu, Sun, Lei,
  Gong, Chen, Zhao, Bai, Ding, Wang, and Li]{qin2026embodiedsmith}
Yikai Qin, Yifei Deng, Mingjian Liang, Wenxuan Song, Zepeng Lin, Zhiyi Jiang,
  Jiajun Fu, Qiao Sun, Huashuo Lei, Xicheng Gong, Jiayi Chen, Han Zhao,
  Shuanghao Bai, Pengxiang Ding, Pengwei Wang, and Haoang Li.
\newblock Embodiedsmith: Scaling embodied data through recursive
  self-improvement flywheel in simulation, 2026.

\bibitem[Raistrick et~al.(2024)Raistrick, Mei, Kayan, Yan, Zuo, Han, Wen,
  Parakh, Alexandropoulos, Lipson, et~al.]{infinigenindoors}
Alexander Raistrick, Lingjie Mei, Karhan Kayan, David Yan, Yiming Zuo, Beining
  Han, Hongyu Wen, Meenal Parakh, Stamatis Alexandropoulos, Lahav Lipson,
  et~al.
\newblock Infinigen indoors: Photorealistic indoor scenes using procedural
  generation.
\newblock In \emph{2024 IEEE/CVF Conference on Computer Vision and Pattern
  Recognition (CVPR)}, pages 21783--21794. IEEE, 2024.

\bibitem[Raistrick et~al.(2026)Raistrick, Kayan, Nugent, Yan, Mei, Parakh, Wen,
  Li, Zuo, Liang, et~al.]{raistrick2026procfunc}
Alexander Raistrick, Karhan Kayan, Jack Nugent, David Yan, Lingjie Mei, Meenal
  Parakh, Hongyu Wen, Dylan Li, Yiming Zuo, Erich Liang, et~al.
\newblock Procfunc: Function-oriented abstractions for procedural 3d generation
  in python.
\newblock \emph{arXiv preprint arXiv:2604.26943}, 2026.

\bibitem[Ramakrishnan et~al.(2021)Ramakrishnan, Gokaslan, Wijmans, Maksymets,
  Clegg, Turner, Undersander, Galuba, Westbury, Chang,
  et~al.]{ramakrishnan2021habitat}
Santhosh~K Ramakrishnan, Aaron Gokaslan, Erik Wijmans, Oleksandr Maksymets,
  Alex Clegg, John Turner, Eric Undersander, Wojciech Galuba, Andrew Westbury,
  Angel~X Chang, et~al.
\newblock Habitat-matterport 3d dataset (hm3d): 1000 large-scale 3d
  environments for embodied ai.
\newblock \emph{arXiv preprint arXiv:2109.08238}, 2021.

\bibitem[Ran et~al.(2026)Ran, Li, Xu, Yu, and Dai]{directlayout}
Xingjian Ran, Yixuan Li, Linning Xu, Mulin Yu, and Bo Dai.
\newblock Direct numerical layout generation for 3d indoor scene synthesis via
  spatial reasoning.
\newblock \emph{Advances in Neural Information Processing Systems},
  38:\penalty0 125055--125081, 2026.

\bibitem[Rezatofighi et~al.(2019)Rezatofighi, Tsoi, Gwak, Sadeghian, Reid, and
  Savarese]{Rezatofighi_2019_CVPR}
Hamid Rezatofighi, Nathan Tsoi, JunYoung Gwak, Amir Sadeghian, Ian Reid, and
  Silvio Savarese.
\newblock Generalized intersection over union: A metric and a loss for bounding
  box regression.
\newblock In \emph{Proceedings of the IEEE/CVF Conference on Computer Vision
  and Pattern Recognition (CVPR)}, 2019.

\bibitem[Rutherford et~al.(2024)Rutherford, Beukman, Willi, Lacerda, Hawes, and
  Foerster]{sfl}
Alex Rutherford, Michael Beukman, Timon Willi, Bruno Lacerda, Nick Hawes, and
  Jakob Foerster.
\newblock No regrets: Investigating and improving regret approximations for
  curriculum discovery.
\newblock \emph{Advances in Neural Information Processing Systems},
  37:\penalty0 16071--16101, 2024.

\bibitem[Shamsaddinlou(2026)]{robolayout}
Ali Shamsaddinlou.
\newblock Robolayout: Differentiable 3d scene generation for embodied agents,
  2026.

\bibitem[{Stability AI}(2024)]{sd35}
{Stability AI}.
\newblock Introducing stable diffusion 3.5.
\newblock Stability AI Blog, 2024.

\bibitem[Sun et~al.(2025{\natexlab{a}})Sun, Liu, Gu, Lim, Bhat, Tombari, Li,
  Haber, and Wu]{layoutvlm}
Fan-Yun Sun, Weiyu Liu, Siyi Gu, Dylan Lim, Goutam Bhat, Federico Tombari,
  Manling Li, Nick Haber, and Jiajun Wu.
\newblock Layoutvlm: Differentiable optimization of 3d layout via
  vision-language models.
\newblock In \emph{2025 IEEE/CVF Conference on Computer Vision and Pattern
  Recognition (CVPR)}, pages 29469--29478. IEEE, 2025{\natexlab{a}}.

\bibitem[Sun et~al.(2026)Sun, Zhang, Wang, Wu, Wang, Wang, Wang, Zhang, Wang,
  and Guo]{sun2026worldplay}
Wenqiang Sun, Haiyu Zhang, Haoyuan Wang, Junta Wu, Zehan Wang, Zhenwei Wang,
  Yunhong Wang, Jun Zhang, Tengfei Wang, and Chunchao Guo.
\newblock Worldplay: Towards long-term geometric consistency for real-time
  interactive world modeling.
\newblock In \emph{International Conference on Machine Learning (ICML)}, pages
  117501--117523, 2026.

\bibitem[Sun et~al.(2025{\natexlab{b}})Sun, Liu, Zang, Cao, Dong, Wu, Lin, and
  Wang]{sun2025seagent}
Zeyi Sun, Ziyu Liu, Yuhang Zang, Yuhang Cao, Xiaoyi Dong, Tong Wu, Dahua Lin,
  and Jiaqi Wang.
\newblock Seagent: Self-evolving computer use agent with autonomous learning
  from experience, 2025{\natexlab{b}}.

\bibitem[Swain et~al.(2026)Swain, Han, and
  Torralba]{swain2026envdreamerlargescalemultimodaltoenvironmentgeneration}
Kabir Swain, Sijie Han, and Antonio Torralba.
\newblock Envdreamer: Large-scale multimodal-to-environment generation for
  embodied ai, 2026.

\bibitem[Tam et~al.(2026)Tam, Pun, Wang, Chang, and Savva]{sceneeval}
Hou In~Ivan Tam, Hou In~Derek Pun, Austin~T. Wang, Angel~X. Chang, and Manolis
  Savva.
\newblock Sceneeval: Evaluating semantic coherence in text-conditioned 3d
  indoor scene synthesis.
\newblock In \emph{Proceedings of the IEEE/CVF Winter Conference on
  Applications of Computer Vision (WACV)}, pages 7355--7365, 2026.

\bibitem[Tang et~al.(2024)Tang, Nie, Markhasin, Dai, Thies, and
  Nie{\ss}ner]{tang2024diffuscene}
Jiapeng Tang, Yinyu Nie, Lev Markhasin, Angela Dai, Justus Thies, and Matthias
  Nie{\ss}ner.
\newblock Diffuscene: Denoising diffusion models for generative indoor scene
  synthesis.
\newblock In \emph{2024 IEEE/CVF Conference on Computer Vision and Pattern
  Recognition (CVPR)}, pages 20507--20518. IEEE, 2024.

\bibitem[Team et~al.(2026)Team, Gao, Wang, Zeng, Zhu, Cheng, Li, Wang, Xu, Ma,
  et~al.]{robbyant2026lingbot}
Robbyant Team, Zelin Gao, Qiuyu Wang, Yanhong Zeng, Jiapeng Zhu, Ka~Leong
  Cheng, Yixuan Li, Hanlin Wang, Yinghao Xu, Shuailei Ma, et~al.
\newblock Advancing open-source world models.
\newblock \emph{arXiv preprint arXiv:2601.20540}, 2026.

\bibitem[Todorov et~al.(2012)Todorov, Erez, and Tassa]{todorov2012mujoco}
Emanuel Todorov, Tom Erez, and Yuval Tassa.
\newblock Mujoco: A physics engine for model-based control.
\newblock In \emph{2012 IEEE/RSJ international conference on intelligent robots
  and systems}, pages 5026--5033. IEEE, 2012.

\bibitem[Tschannen et~al.(2025)Tschannen, Gritsenko, Wang, Naeem,
  Alabdulmohsin, Parthasarathy, Evans, Beyer, Xia, Mustafa,
  et~al.]{tschannen2025siglip}
Michael Tschannen, Alexey Gritsenko, Xiao Wang, Muhammad~Ferjad Naeem, Ibrahim
  Alabdulmohsin, Nikhil Parthasarathy, Talfan Evans, Lucas Beyer, Ye Xia, Basil
  Mustafa, et~al.
\newblock Siglip 2: Multilingual vision-language encoders with improved
  semantic understanding, localization, and dense features.
\newblock \emph{arXiv preprint arXiv:2502.14786}, 2025.

\bibitem[Wald et~al.(2019)Wald, Avetisyan, Navab, Tombari, and
  Niessner]{Wald_2019_ICCV}
Johanna Wald, Armen Avetisyan, Nassir Navab, Federico Tombari, and Matthias
  Niessner.
\newblock {RIO}: 3d object instance re-localization in changing indoor
  environments.
\newblock In \emph{Proceedings of the IEEE/CVF International Conference on
  Computer Vision (ICCV)}, 2019.

\bibitem[Wang et~al.(2023)Wang, Xie, Jiang, Mandlekar, Xiao, Zhu, Fan, and
  Anandkumar]{wang2023voyager}
Guanzhi Wang, Yuqi Xie, Yunfan Jiang, Ajay Mandlekar, Chaowei Xiao, Yuke Zhu,
  Linxi Fan, and Anima Anandkumar.
\newblock Voyager: An open-ended embodied agent with large language models.
\newblock \emph{arXiv preprint arXiv:2305.16291}, 2023.

\bibitem[Wang et~al.(2024{\natexlab{a}})Wang, Chen, Huang, Ben, Wang, Mi,
  Huang, Zhao, Chen, Yang, Cao, Yu, Ye, Li, Long, Wang, Wang, Zhao, Tu, Qiao,
  Lin, and Pang]{wang2024grutopiadreamgeneralrobots}
Hanqing Wang, Jiahe Chen, Wensi Huang, Qingwei Ben, Tai Wang, Boyu Mi, Tao
  Huang, Siheng Zhao, Yilun Chen, Sizhe Yang, Peizhou Cao, Wenye Yu, Zichao Ye,
  Jialun Li, Junfeng Long, Zirui Wang, Huiling Wang, Ying Zhao, Zhongying Tu,
  Yu Qiao, Dahua Lin, and Jiangmiao Pang.
\newblock Grutopia: Dream general robots in a city at scale,
  2024{\natexlab{a}}.

\bibitem[Wang et~al.(2025)Wang, Chen, Karaev, Vedaldi, Rupprecht, and
  Novotny]{wang2025vggt}
Jianyuan Wang, Minghao Chen, Nikita Karaev, Andrea Vedaldi, Christian
  Rupprecht, and David Novotny.
\newblock Vggt: Visual geometry grounded transformer.
\newblock In \emph{2025 IEEE/CVF Conference on Computer Vision and Pattern
  Recognition (CVPR)}, pages 5294--5306. IEEE, 2025.

\bibitem[Wang et~al.(2024{\natexlab{b}})Wang, Ling, Yuan, Shridhar, Bao, Qin,
  Wang, Xu, and Wang]{wang2024gensim}
Lirui Wang, Yiyang Ling, Zhecheng Yuan, Mohit Shridhar, Chen Bao, Yuzhe Qin,
  Bailin Wang, Huazhe Xu, and Xiaolong Wang.
\newblock Gensim: Generating robotic simulation tasks via large language
  models.
\newblock In \emph{International Conference on Learning Representations}, pages
  4890--4924, 2024{\natexlab{b}}.

\bibitem[Wang et~al.(2026{\natexlab{a}})Wang, Wang, Li, Yang, Lin, Li, and
  Cheng]{wang2026scenecode}
Puyi Wang, Yuhao Wang, Linjie Li, Zhengyuan Yang, Kevin~Qinghong Lin, Yangguang
  Li, and Yu Cheng.
\newblock Scenecode: Executable world programs for editable indoor scenes with
  articulated objects.
\newblock \emph{arXiv preprint arXiv:2605.19587}, 2026{\natexlab{a}}.

\bibitem[Wang et~al.(2019)Wang, Zhou, Shi, Chen, Zhao, and
  Xu]{wang2019shape2motion}
Xiaogang Wang, Bin Zhou, Yahao Shi, Xiaowu Chen, Qinping Zhao, and Kai Xu.
\newblock Shape2motion: Joint analysis of motion parts and attributes from 3d
  shapes.
\newblock In \emph{2019 IEEE/CVF Conference on Computer Vision and Pattern
  Recognition (CVPR)}, pages 8868--8876. IEEE, 2019.

\bibitem[Wang et~al.(2026{\natexlab{b}})Wang, Liu, Ding, Choi, Huang, Zhao, Li,
  Jiang, Yu, Liu, et~al.]{wang2026embodiedgen}
Xinjie Wang, Liu Liu, Taojun Ding, Andrew Choi, Chaodong Huang, Mengao Zhao,
  Ziang Li, Jackson Jiang, Chunlei Yu, Shengxiang Liu, et~al.
\newblock Embodiedgen v2: An agentic, simulation-ready 3d world engine for
  embodied ai.
\newblock \emph{arXiv preprint arXiv:2607.07459}, 2026{\natexlab{b}}.

\bibitem[Wang et~al.(2024{\natexlab{c}})Wang, Qiu, Liu, Chen, Cai, Wang, Wang,
  Xian, and Gan]{wang2024architect}
Yian Wang, Xiaowen Qiu, Jiageng Liu, Zhehuan Chen, Jiting Cai, Yufei Wang,
  Tsun-Hsuan Wang, Zhou Xian, and Chuang Gan.
\newblock Architect: Generating vivid and interactive 3d scenes with
  hierarchical 2d inpainting.
\newblock \emph{Advances in Neural Information Processing Systems},
  37:\penalty0 67575--67603, 2024{\natexlab{c}}.

\bibitem[Wang et~al.(2024{\natexlab{d}})Wang, Xian, Chen, Wang, Wang,
  Fragkiadaki, Erickson, Held, and Gan]{wang2024robogen}
Yufei Wang, Zhou Xian, Feng Chen, Tsun-Hsuan Wang, Yian Wang, Katerina
  Fragkiadaki, Zackory Erickson, David Held, and Chuang Gan.
\newblock {RoboGen}: Towards unleashing infinite data for automated robot
  learning via generative simulation.
\newblock In \emph{Proceedings of the 41st International Conference on Machine
  Learning}, pages 51936--51983, 2024{\natexlab{d}}.

\bibitem[Wei et~al.(2022)Wei, Liu, Ling, and Su]{wei2022approximate}
Xinyue Wei, Minghua Liu, Zhan Ling, and Hao Su.
\newblock Approximate convex decomposition for 3d meshes with collision-aware
  concavity and tree search.
\newblock \emph{ACM Transactions on Graphics (TOG)}, 41\penalty0 (4):\penalty0
  1--18, 2022.

\bibitem[Wortsman et~al.(2022)Wortsman, Ilharco, Kim, Li, Kornblith, Roelofs,
  Lopes, Hajishirzi, Farhadi, Namkoong, et~al.]{wortsman2022wise}
Mitchell Wortsman, Gabriel Ilharco, Jong~Wook Kim, Mike Li, Simon Kornblith,
  Rebecca Roelofs, Raphael~Gontijo Lopes, Hannaneh Hajishirzi, Ali Farhadi,
  Hongseok Namkoong, et~al.
\newblock Robust fine-tuning of zero-shot models.
\newblock In \emph{2022 IEEE/CVF Conference on Computer Vision and Pattern
  Recognition (CVPR)}, pages 7949--7961. IEEE, 2022.

\bibitem[Wu et~al.(2023)Wu, Antonova, Kan, Lepert, Zeng, Song, Bohg,
  Rusinkiewicz, and Funkhouser]{wu2023tidybot}
Jimmy Wu, Rika Antonova, Adam Kan, Marion Lepert, Andy Zeng, Shuran Song,
  Jeannette Bohg, Szymon Rusinkiewicz, and Thomas Funkhouser.
\newblock Tidybot: Personalized robot assistance with large language models.
\newblock \emph{Autonomous Robots}, 47\penalty0 (8):\penalty0 1087--1102, 2023.

\bibitem[Wu et~al.(2024)Wu, Li, Yan, Shang, Sun, Wang, Cui, Liu, Sato, Li,
  et~al.]{wu2024blockfusion}
Zhennan Wu, Yang Li, Han Yan, Taizhang Shang, Weixuan Sun, Senbo Wang, Ruikai
  Cui, Weizhe Liu, Hiroyuki Sato, Hongdong Li, et~al.
\newblock Blockfusion: Expandable 3d scene generation using latent tri-plane
  extrapolation.
\newblock \emph{ACM Transactions on Graphics (ToG)}, 43\penalty0 (4):\penalty0
  1--17, 2024.

\bibitem[Xia et~al.(2026)Xia, Li, Li, Ma, Xu, Liu, Cui, Lin, Ma, Wang, Song,
  and Wei]{sage}
Hongchi Xia, Xuan Li, Zhaoshuo Li, Qianli Ma, Jiashu Xu, Ming-Yu Liu, Yin Cui,
  Tsung-Yi Lin, Wei-Chiu Ma, Shenlong Wang, Shuran Song, and Fangyin Wei.
\newblock Sage: Scalable agentic 3d scene generation for embodied ai.
\newblock In \emph{IEEE/CVF Conference on Computer Vision and Pattern
  Recognition (CVPR)}, 2026.

\bibitem[Xiang et~al.(2026)Xiang, Bao, Feng, Wu, Liu, Guan, and Liu]{colayout}
Chucheng Xiang, Ruchao Bao, Biyin Feng, Wenzheng Wu, Zhongyuan Liu, Yirui Guan,
  and Ligang Liu.
\newblock Co-layout: Llm-driven co-optimization for interior layout.
\newblock In \emph{Proceedings of the AAAI Conference on Artificial
  Intelligence}, pages 14371--14379, 2026.

\bibitem[Xiang et~al.(2020)Xiang, Qin, Mo, Xia, Zhu, Liu, Liu, Jiang, Yuan,
  Wang, et~al.]{xiang2020sapien}
Fanbo Xiang, Yuzhe Qin, Kaichun Mo, Yikuan Xia, Hao Zhu, Fangchen Liu, Minghua
  Liu, Hanxiao Jiang, Yifu Yuan, He Wang, et~al.
\newblock Sapien: A simulated part-based interactive environment.
\newblock In \emph{2020 IEEE/CVF Conference on Computer Vision and Pattern
  Recognition (CVPR)}, pages 11094--11104. IEEE, 2020.

\bibitem[Xiang et~al.(2025)Xiang, Lv, Xu, Deng, Wang, Zhang, Chen, Tong, and
  Yang]{xiang2025structured}
Jianfeng Xiang, Zelong Lv, Sicheng Xu, Yu Deng, Ruicheng Wang, Bowen Zhang,
  Dong Chen, Xin Tong, and Jiaolong Yang.
\newblock Structured 3d latents for scalable and versatile 3d generation.
\newblock In \emph{2025 IEEE/CVF Conference on Computer Vision and Pattern
  Recognition (CVPR)}, pages 21469--21480. IEEE, 2025.

\bibitem[Xie et~al.(2026)Xie, Xu, Zhao, and Song]{xie2026agentsynth}
Jingxu Xie, Dylan Xu, Xuandong Zhao, and Dawn Song.
\newblock Agentsynth: Scalable task generation for generalist computer-use
  agents.
\newblock In \emph{International Conference on Learning Representations}, pages
  155149--155177, 2026.

\bibitem[Xu et~al.(2026)Xu, Li, Lin, Xue, Xian, Xu, Wu, Zhang, Deng, Yu,
  et~al.]{xu2026deepseek}
Anyi Xu, B Li, Bangcai Lin, Bing Xue, BingCheng Xian, Bingzheng Xu, Bochao Wu,
  Bowei Zhang, Boyi Deng, CC Yu, et~al.
\newblock Deepseek-v4. 1-flash: Pushing the limits of kv cache compression.
\newblock \emph{arXiv preprint arXiv:2609.19969}, 2026.

\bibitem[Yadav et~al.(2023)Yadav, Ramrakhya, Ramakrishnan, Gervet, Turner,
  Gokaslan, Maestre, Chang, Batra, Savva, Clegg, and Chaplot]{Yadav_2023_CVPR}
Karmesh Yadav, Ram Ramrakhya, Santhosh~Kumar Ramakrishnan, Theo Gervet, John
  Turner, Aaron Gokaslan, Noah Maestre, Angel~Xuan Chang, Dhruv Batra, Manolis
  Savva, Alexander~William Clegg, and Devendra~Singh Chaplot.
\newblock Habitat-matterport 3d semantics dataset.
\newblock In \emph{Proceedings of the IEEE/CVF Conference on Computer Vision
  and Pattern Recognition (CVPR)}, pages 4927--4936, 2023.

\bibitem[Yang et~al.(2025)Yang, Li, Yang, Zhang, Hui, Zheng, Yu, Gao, Huang,
  Lv, et~al.]{qwen2025qwen3}
An Yang, Anfeng Li, Baosong Yang, Beichen Zhang, Binyuan Hui, Bo Zheng, Bowen
  Yu, Chang Gao, Chengen Huang, Chenxu Lv, et~al.
\newblock Qwen3 technical report.
\newblock \emph{arXiv preprint arXiv:2505.09388}, 2025.

\bibitem[Yang et~al.(2026{\natexlab{a}})Yang, Dagli, Zook, Hadfield, Goyal,
  Birchfield, Ramos, and Tremblay]{yang2026robolab}
Xuning Yang, Rishit Dagli, Alex Zook, Hugo Hadfield, Ankit Goyal, Stan
  Birchfield, Fabio Ramos, and Jonathan Tremblay.
\newblock {RoboLab: A High-Fidelity Simulation Benchmark for Analysis of Task
  Generalist Policies}.
\newblock In \emph{Proceedings of Robotics: Science and Systems}, Sydney,
  Australia, 2026{\natexlab{a}}.

\bibitem[Yang et~al.(2024{\natexlab{a}})Yang, Jia, Zhi, and
  Huang]{yang2024physcene}
Yandan Yang, Baoxiong Jia, Peiyuan Zhi, and Siyuan Huang.
\newblock Physcene: Physically interactable 3d scene synthesis for embodied ai.
\newblock In \emph{2024 IEEE/CVF Conference on Computer Vision and Pattern
  Recognition (CVPR)}, pages 16262--16272. IEEE, 2024{\natexlab{a}}.

\bibitem[Yang et~al.(2024{\natexlab{b}})Yang, Sun, Weihs, VanderBilt, Herrasti,
  Han, Wu, Haber, Krishna, Liu, Callison-Burch, Yatskar, Kembhavi, and
  Clark]{holodeck}
Yue Yang, Fan-Yun Sun, Luca Weihs, Eli VanderBilt, Alvaro Herrasti, Winson Han,
  Jiajun Wu, Nick Haber, Ranjay Krishna, Lingjie Liu, Chris Callison-Burch,
  Mark Yatskar, Aniruddha Kembhavi, and Christopher Clark.
\newblock Holodeck: Language guided generation of 3d embodied ai environments.
\newblock In \emph{Proceedings of the IEEE/CVF Conference on Computer Vision
  and Pattern Recognition (CVPR)}, pages 16227--16237, 2024{\natexlab{b}}.

\bibitem[Yang et~al.(2026{\natexlab{b}})Yang, Jia, Zhang, and
  Huang]{sceneweaver}
Yandan Yang, Baoxiong Jia, Shujie Zhang, and Siyuan Huang.
\newblock Sceneweaver: All-in-one 3d scene synthesis with an extensible and
  self-reflective agent.
\newblock \emph{Advances in neural information processing systems},
  38:\penalty0 140319--140351, 2026{\natexlab{b}}.

\bibitem[Yang et~al.(2026{\natexlab{c}})Yang, Luo, Ding, Lu, Gao, Yang,
  Sanchez, and Zheng]{optiscene}
Yixuan Yang, Zhen Luo, Tongsheng Ding, Junru Lu, Mingqi Gao, Jinyu Yang, Victor
  Sanchez, and Feng Zheng.
\newblock Optiscene: Llm-driven indoor scene layout generation via scaled
  human-aligned data synthesis and multi-stage preference optimization.
\newblock \emph{Advances in Neural Information Processing Systems},
  38:\penalty0 42499--42529, 2026{\natexlab{c}}.

\bibitem[Yeshwanth et~al.(2023)Yeshwanth, Liu, Nie{\ss}ner, and
  Dai]{Yeshwanth_2023_ICCV}
Chandan Yeshwanth, Yueh-Cheng Liu, Matthias Nie{\ss}ner, and Angela Dai.
\newblock Scannet++: A high-fidelity dataset of 3d indoor scenes.
\newblock In \emph{Proceedings of the IEEE/CVF International Conference on
  Computer Vision (ICCV)}, pages 12--22, 2023.

\bibitem[Zeng et~al.(2026)Zeng, Lv, Hou, Du, Zheng, Chen, Yin, Ge, Huang, Xie,
  et~al.]{zeng2026glm}
Aohan Zeng, Xin Lv, Zhenyu Hou, Zhengxiao Du, Qinkai Zheng, Bin Chen, Da Yin,
  Chendi Ge, Chenghua Huang, Chengxing Xie, et~al.
\newblock Glm-5: from vibe coding to agentic engineering.
\newblock \emph{arXiv preprint arXiv:2602.15763}, 2026.

\bibitem[Zhang et~al.(2025)Zhang, Lazuka, and Murag]{anthropic_2025}
Barry Zhang, Keith Lazuka, and Mahesh Murag.
\newblock Equipping agents for the real world with agent skills.
\newblock Anthropic Engineering Blog, 2025.

\bibitem[Zhang et~al.(2026{\natexlab{a}})Zhang, Li, Zhang, Zhang, Chua, and
  Tang]{zhang2026code2worlds}
Haoyu Zhang, Siming Li, Zeyu Zhang, Yi Zhang, Tat-Seng Chua, and Hao Tang.
\newblock Code2worlds-2: Physically plausible agent harness for embodied world
  simulation.
\newblock \emph{arXiv preprint arXiv:2610.17651}, 2026{\natexlab{a}}.

\bibitem[Zhang et~al.(2026{\natexlab{b}})Zhang, Sun, Zhao, Liu, Gao, and
  Bian]{zonemaestro}
Meisheng Zhang, Shizhao Sun, Yang Zhao, Ziyuan Liu, Zhijun Gao, and Jiang Bian.
\newblock Orchestrating spatial semantics via a zone-graph paradigm for
  intricate indoor scene generation.
\newblock In \emph{International Conference on Machine Learning (ICML)}, pages
  158519--158567, 2026{\natexlab{b}}.

\bibitem[Zhang et~al.(2018)Zhang, Isola, Efros, Shechtman, and
  Wang]{zhang2018unreasonable}
Richard Zhang, Phillip Isola, Alexei~A Efros, Eli Shechtman, and Oliver Wang.
\newblock The unreasonable effectiveness of deep features as a perceptual
  metric.
\newblock In \emph{2018 IEEE/CVF conference on computer vision and pattern
  recognition}, pages 586--595. IEEE, 2018.

\bibitem[Zhang et~al.(2026{\natexlab{c}})Zhang, Cai, Wang, Guo, Li, Lin, and
  Wang]{Zhang_2026_CVPR}
Yiheng Zhang, Zhuojiang Cai, Mingdao Wang, Meitong Guo, Tianxiao Li, Li Lin,
  and Yuwang Wang.
\newblock M3dlayout: A multi-source dataset of 3d indoor layouts and structured
  descriptions for 3d generation.
\newblock In \emph{Proceedings of the IEEE/CVF Conference on Computer Vision
  and Pattern Recognition (CVPR)}, pages 34217--34226, 2026{\natexlab{c}}.

\bibitem[Zhao et~al.(2026)Zhao, Sun, Zhang, Shi, Yang, and Bian]{scenerevis}
Yang Zhao, Shizhao Sun, Meisheng Zhang, Yingdong Shi, Xubo Yang, and Jiang
  Bian.
\newblock Scenerevis: A self-reflective vision-grounded framework for 3d indoor
  scene synthesis via multi-turn rl.
\newblock In \emph{Proceedings of the 34th ACM International Conference on
  Multimedia}, pages 15445--15453, 2026.

\bibitem[Zheng et~al.(2020)Zheng, Zhang, Li, Tang, Gao, and
  Zhou]{zheng2020structured3d}
Jia Zheng, Junfei Zhang, Jing Li, Rui Tang, Shenghua Gao, and Zihan Zhou.
\newblock Structured3d: A large photo-realistic dataset for structured 3d
  modeling.
\newblock In \emph{Proceedings of The European Conference on Computer Vision
  (ECCV)}, 2020.

\bibitem[Zheng et~al.(2023)Zheng, Chiang, Sheng, Zhuang, Wu, Zhuang, Lin, Li,
  Li, Xing, et~al.]{zheng2023judging}
Lianmin Zheng, Wei-Lin Chiang, Ying Sheng, Siyuan Zhuang, Zhanghao Wu, Yonghao
  Zhuang, Zi Lin, Zhuohan Li, Dacheng Li, Eric Xing, et~al.
\newblock Judging llm-as-a-judge with mt-bench and chatbot arena.
\newblock \emph{Advances in neural information processing systems},
  36:\penalty0 46595--46623, 2023.

\bibitem[Zhong et~al.(2025)Zhong, Cao, Jin, Li, Cai, Lin, Wang, Lyu, WANG, XU,
  Dai, and Pang]{zhong2026internscenes}
Weipeng Zhong, Peizhou Cao, Yichen Jin, Luo Li, Wenzhe Cai, Jingli Lin, Hanqing
  Wang, Zhaoyang Lyu, Tai WANG, Xudong XU, Bo Dai, and Jiangmiao Pang.
\newblock Internscenes: A large-scale simulatable indoor scene dataset with
  realistic layouts.
\newblock In \emph{Advances in Neural Information Processing Systems}. Curran
  Associates, Inc., 2025.

\bibitem[Zhou et~al.(2026)Zhou, Li, Lyu, Song, Huang, Zheng, Rupprecht,
  Vedaldi, and Wu]{zhou2026articraft}
Matt Zhou, Ruining Li, Xiaoyang Lyu, Zhaomou Song, Zhening Huang, Chuanxia
  Zheng, Christian Rupprecht, Andrea Vedaldi, and Shangzhe Wu.
\newblock Articraft: An agentic system for scalable articulated 3d asset
  generation.
\newblock \emph{arXiv preprint arXiv:2605.15187}, 2026.

\bibitem[Zhou et~al.(2025)Zhou, Nie, Du, Yin, Huang, Guo, Zhang, and
  Hu]{zhou2025il3d}
Wenxu Zhou, Kaixuan Nie, Hang Du, Dong Yin, Wei Huang, Siqiang Guo, Xiaobo
  Zhang, and Pengbo Hu.
\newblock Il3d: A large-scale indoor layout dataset for llm-driven 3d scene
  generation.
\newblock \emph{arXiv preprint arXiv:2510.12095}, 2025.

\bibitem[Çelen et~al.(2024)Çelen, Han, Schindler, Gool, Armeni, Obukhov, and
  Wang]{idesign}
Ata Çelen, Guo Han, Konrad Schindler, Luc~Van Gool, Iro Armeni, Anton Obukhov,
  and Xi Wang.
\newblock I-design: Personalized llm interior designer, 2024.

\end{thebibliography}
}
\startcontents[appendix]
\twocolumn[{%
\pdfbookmark[0]{Appendix}{appendix.contents}
\noindent\makebox[\textwidth]{\textbf{\method: Persistent Task-Driven World Expansion}}\par
\nointerlineskip\vspace{3pt}\hrule\vspace{14pt}
{\noindent\LARGE\bfseries Appendix\par}\vspace{12pt}
{\noindent\large\bfseries Contents\par}\vspace{6pt}
\begingroup
\hypersetup{linktoc=all,linkcolor=black}
\printcontents[appendix]{app}{1}[2]{}
\endgroup
\vspace{12pt}
}]
\appendix
\ifdefined\pdfsuppressptexinfo\pdfsuppressptexinfo=-1 \fi
\clubpenalties 2 10000 0\relax
\widowpenalties 2 10000 0\relax
\newenvironment{points}{\begin{itemize}\small\setlength{\itemsep}{1pt}\setlength{\parskip}{0pt}\setlength{\topsep}{2pt}}{\end{itemize}}

\section{Motivation and Comparison with Prior \mbox{Systems}}
\label{supp:motivation}

%
\begin{table*}[t]
\centering
\caption{\textbf{System comparison.} In the reported settings, only \method{} combines history-based task proposals, connected room additions as needed and preservation of previously manipulated objects' identities and states.}
\label{tab:supp:motivation:systems}
\definecolor{scTask}{HTML}{3A6CA6}
\definecolor{scWorld}{HTML}{B8661F}
\definecolor{scExec}{HTML}{2F7866}
\definecolor{scOursEdge}{HTML}{2F5F8F}
\definecolor{scOursFill}{HTML}{EEF4FB}
\colorlet{scNo}{black!38}
\colorlet{scRule}{black!22}
\colorlet{scOursRule}{scOursEdge!22}
\newlength{\scR}\setlength{\scR}{2.75pt}
\newcommand{\scCur}{neutralD}
\newcommand{\scBall}[1]{$\vcenter{\hbox{\tikz{%
  \path[use as bounding box] (-\scR,-\scR) rectangle (\scR,\scR); #1}}}$}
\newcommand{\scY}{\scBall{\fill[\scCur] (0,0) circle (\scR);}}
\newcommand{\scP}{\scBall{%
  \fill[\scCur] (0,0) -- (90:\scR) arc (90:270:\scR) -- cycle;
  \draw[\scCur, line width=.55pt] (0,0) circle ({\scR-.275pt});}}
\newcommand{\scN}{\scBall{%
  \draw[scNo, line width=.5pt] (0,0) circle ({\scR-.25pt});}}
\newsavebox{\scBox}
\newcommand{\scRot}[1]{\sbox{\scBox}{#1}\makebox[0pt][l]{%
  \hspace*{\dimexpr-\ht\scBox*433/1000\relax}\rotatebox{60}{\usebox{\scBox}}}}
\newcommand{\scSys}[2]{\multicolumn{1}{c}{%
  \raisebox{3pt}{\scRot{#1 {\scriptsize\cite{#2}}}}}}
\newlength{\scRowH}\setlength{\scRowH}{22pt}
\newlength{\scBandW}\setlength{\scBandW}{0.5cm}
\newlength{\scLab}\setlength{\scLab}{3.5cm}
\newcommand{\scL}[1]{\parbox[c][\scRowH][c]{\scLab}{\raggedright #1}}
\newcommand{\scGroup}[3]{\gdef\scCur{#2}%
  \parbox[c][\scRowH][t]{\scBandW}{\tikz[overlay]{%
    \draw[#2, line width=.9pt, line cap=round]
      (\scBandW-4pt,-3pt) -- (\scBandW-4pt,{3pt-#1*\scRowH});
    \node[rotate=90, text=#2, font=\scriptsize, align=center, inner sep=0pt]
      at (\scBandW*.5-3pt,{-#1*\scRowH*.5}) {#3};}}}
\newcommand{\scO}[1]{\gdef\scCur{#1}\parbox[c][\scRowH]{0pt}{}}
\footnotesize
\setlength{\tabcolsep}{0pt}
\setlength{\aboverulesep}{0pt}
\setlength{\belowrulesep}{0pt}
\setlength{\cmidrulekern}{3pt}
\renewcommand{\arraystretch}{1}
\newlength{\scOursW}\setlength{\scOursW}{1.2cm}
\newlength{\scGapCat}\setlength{\scGapCat}{5pt}
\newlength{\scGapOurs}\setlength{\scGapOurs}{0.7cm}
\newlength{\scCol}
\setlength{\scCol}{\dimexpr(\textwidth-\scBandW-\scLab-3\scGapCat
  -\scGapOurs-\scOursW-1pt)/16\relax}
\begin{tabular}[b]{@{}l@{}l@{}
  *{8}{w{c}{\scCol}}@{\hspace{\scGapCat}}
  *{4}{w{c}{\scCol}}@{\hspace{\scGapCat}}
  *{3}{w{c}{\scCol}}@{\hspace{\scGapCat}}
  w{c}{\scCol}@{}}
 & &
\scSys{ProcTHOR}{deitke2022} &
\scSys{Holodeck}{holodeck} &
\scSys{SceneSmith}{pfaff2026scenesmith} &
\scSys{SAGE}{sage} &
\scSys{EmbodiedGen V2}{wang2026embodiedgen} &
\scSys{MANSION}{Che_2026_CVPR} &
\scSys{WorldGrow}{li2026worldgrow} &
\scSys{BlockFusion}{wu2024blockfusion} &
\scSys{RoboGen}{wang2024robogen} &
\scSys{PARTNR}{ICLR2025_a3cf318f} &
\scSys{SimWorld Studio}{kang2026simworld} &
\scSys{Eurekaverse}{liang2024eurekaverse} &
\scSys{OMNI-EPIC}{faldor2025omni} &
\scSys{Voyager}{wang2023voyager} &
\scSys{AgentSynth}{xie2026agentsynth} &
\scSys{Ctrl-World}{guo2026ctrlworld} \\
\arrayrulecolor{neutralD}\cmidrule[.6pt](lr){3-10}\cmidrule[.6pt](lr){11-14}%
\cmidrule[.6pt](lr){15-17}\cmidrule[.6pt](lr){18-18}
\scGroup{3}{scTask}{Task generation} &
\scL{Conditions tasks on\newline world state and history}
 & \scP & \scN & \scN & \scN & \scN & \scN & \scN & \scN & \scN & \scP & \scP & \scP & \scP & \scY & \scY & \scN \\
 & \scL{Checks task feasibility\newline before building the scene}
 & \scP & \scN & \scN & \scN & \scP & \scP & \scN & \scN & \scN & \scY & \scP & \scP & \scP & \scN & \scP & \scN \\
 & \scL{Selects tasks by\newline target difficulty}
 & \scN & \scN & \scN & \scN & \scN & \scN & \scN & \scN & \scN & \scN & \scY & \scY & \scP & \scP & \scY & \scN \\
\arrayrulecolor{scRule}\cmidrule[.4pt]{2-18}
\scGroup{6}{scWorld}{World generation} &
\scL{Generates the room\newline a task needs}
 & \scN & \scN & \scY & \scY & \scY & \scP & \scN & \scN & \scY & \scP & \scP & \scP & \scY & \scN & \scN & \scN \\
 & \scL{Connected\newline multi-room worlds}
 & \scY & \scY & \scY & \scP & \scY & \scY & \scY & \scP & \scN & \scP & \scN & \scN & \scP & \scN & \scN & \scN \\
 & \scL{Incrementally extends\newline an existing scene}
 & \scN & \scN & \scN & \scN & \scN & \scN & \scY & \scY & \scN & \scN & \scN & \scN & \scN & \scN & \scN & \scN \\
 & \scL{Keeps geometry and object\newline IDs as the world grows}
 & \scN & \scN & \scN & \scN & \scP & \scP & \scP & \scP & \scN & \scN & \scP & \scN & \scN & \scN & \scN & \scN \\
 & \scL{Gives task-operated\newline objects working joints}
 & \scP & \scN & \scP & \scP & \scN & \scP & \scN & \scN & \scP & \scP & \scN & \scN & \scN & \scN & \scN & \scN \\
 & \scL{Checks physical validity\newline and robot access}
 & \scY & \scP & \scY & \scP & \scY & \scY & \scN & \scN & \scP & \scP & \scY & \scP & \scN & \scN & \scN & \scN \\
\arrayrulecolor{scRule}\cmidrule[.4pt]{2-18}
\scGroup{4}{scExec}{Persistence and execution} &
\scL{Keeps object states\newline across successive tasks}
 & \scN & \scN & \scN & \scN & \scN & \scN & \scN & \scN & \scN & \scN & \scN & \scN & \scN & \scY & \scP & \scN \\
 & \scL{Feeds execution results\newline into the next task}
 & \scN & \scN & \scN & \scN & \scN & \scN & \scN & \scN & \scN & \scN & \scY & \scY & \scY & \scY & \scY & \scN \\
 & \scL{Evaluates robot policies\newline in generated scenes}
 & \scP & \scP & \scY & \scY & \scY & \scY & \scN & \scN & \scY & \scP & \scY & \scP & \scY & \scN & \scN & \scP \\
 & \scL{Evaluates video world\newline models in generated scenes}
 & \scN & \scN & \scN & \scN & \scN & \scN & \scN & \scN & \scN & \scN & \scN & \scN & \scN & \scN & \scN & \scP \\
\arrayrulecolor{neutralD}\cmidrule[.6pt]{2-18}
\end{tabular}%
\hspace{\scGapOurs}%
\begin{lrbox}{\scBox}\begin{tabular}[b]{@{}l@{}w{c}{\scOursW}@{}}
 & \rotatebox{60}{\textcolor{scOursEdge}{\textbf{\method{}}}} \\[3pt]
\scO{scTask} & \scY \\
\scO{scTask} & \scY \\
\scO{scTask} & \scY \\
\arrayrulecolor{scOursRule}\cmidrule[.4pt]{2-2}
\scO{scWorld} & \scY \\
\scO{scWorld} & \scY \\
\scO{scWorld} & \scY \\
\scO{scWorld} & \scY \\
\scO{scWorld} & \scP \\
\scO{scWorld} & \scY \\
\arrayrulecolor{scOursRule}\cmidrule[.4pt]{2-2}
\scO{scExec} & \scY \\
\scO{scExec} & \scY \\
\scO{scExec} & \scY \\
\scO{scExec} & \scY\rlap{\textsuperscript{\,$\dagger$}} \\
\arrayrulecolor{scOursFill}\cmidrule[.6pt](lr){2-2}
\end{tabular}\end{lrbox}%
\tikz[baseline=(o.base)]\node[draw=scOursEdge, fill=scOursFill,
  line width=.6pt, rounded corners=4pt, inner xsep=0pt, inner ysep=4pt]
  (o) {\usebox{\scBox}};

\vspace{4pt}
\par\noindent
\parbox{\linewidth}{%
\raggedright
\scriptsize
\renewcommand{\scCur}{neutralD}%
\scY\ reported;\quad
\scP\ partial or restricted;\quad
\scN\ absent or unreported.\quad
$^{\dagger}$ Predictions are scored against simulator renders; world models do not modify the world.
}
\end{table*}

\paragraph{Task-driven expansion.}
RoboGen~\cite{wang2024robogen} proposes a task, builds a scene of
task-relevant assets for it, and learns a skill in that scene.
The next task gets a new scene.
SceneSmith~\cite{pfaff2026scenesmith}, SAGE~\cite{sage}, and EmbodiedGen
V2~\cite{wang2026embodiedgen} also build fresh scenes for each given task.
In this task-to-room pattern, no world accumulates, and no earlier room
or task effect constrains the next scene.
In \method{}, a task specifies any additions it needs to the current
world $W_t$ (Sec.~\ref{sec:task_specification}).
When a task needs a new room, the room attaches at a free doorway.
The task can use the new room, revisit objects that earlier tasks moved,
or combine both.

\paragraph{Keeping geometry and object identities.}
When a room is added, existing objects keep their geometry, pose, ID,
and state. Existing rooms retain their geometry apart from the opening edits of
Sec.~\ref{supp:world:integration}.
The interaction history $H_t$ refers to objects by ID
(Sec.~\ref{supp:task:generator}), so a later task can refer to the same instance an earlier task moved.
In Sec.~\ref{supp:task:online}, a bottle keeps its ID as it moves from the starting
house to a new entryway and then into the new kitchen's fridge.

\paragraph{One-shot generation.}
Holodeck~\cite{holodeck}, ProcTHOR~\cite{deitke2022}, SceneSmith, and SAGE
generate a whole scene from one prompt or specification.
EmbodiedGen V2, the editing agent of MANSION~\cite{Che_2026_CVPR}, and
SimWorld Studio~\cite{kang2026simworld} let authors edit individual
objects in an existing scene. Their reported workflows do not use the
state left by a robot to choose the next room.
WorldGrow~\cite{li2026worldgrow} uses the geometry of earlier blocks to guide each new block.

\paragraph{Without preservation.}
If adding a room generated the house again, $H_t$ would point to IDs that no
longer exist or to objects that have moved.
We measured this effect by furnishing the existing rooms again before
each round in the 30 symbolic sessions (Sec.~\ref{supp:world:preserve}).
Even after matching objects by class and room, only 1.8\% of untouched
objects kept their pose and bounding box. Among the 950 tasks that
reused an object moved earlier, only 0.8\% found every such object
on the support or in the container where the earlier task had left it.
We also checked the room additions saved by the three adapted proposers
in Table~\ref{tab:task_generation_expansion}, which use the same procedure
as \method{} to add a room. Across 4,013 such rounds, all 1,308,209
pre-existing object records matched their values before the addition,
and every newly assigned object ID was distinct from the existing IDs.
In the 3D worlds, each object ID is qualified by its room ID
(Sec.~\ref{supp:world:integration}).

\paragraph{System comparison.}
Table~\ref{tab:supp:motivation:systems} compares 16 prior systems across
task generation, room generation, and execution. The comparison includes
blockwise scene extension in BlockFusion~\cite{wu2024blockfusion},
adaptive course generation in Eurekaverse~\cite{liang2024eurekaverse},
and robot policy evaluation in the Ctrl-World video model~\cite{guo2026ctrlworld}.
Among the eight scene generators in the first group, ProcTHOR also samples
navigation targets that an agent can reach and see in the finished house.
SimWorld Studio combines physical verification and adaptive difficulty
in its navigation setting. Its curriculum generates new environment
batches as the agent reaches successive mastery thresholds.
Eurekaverse also feeds policy results back and sets a difficulty, but it
checks a course only against height thresholds, and each course is a new
height field for one fixed parkour task.
Of the 16 prior systems, WorldGrow and BlockFusion extend existing scenes
block by block. Neither reports using the outcome of a task to choose the
next block. Voyager~\cite{wang2023voyager} uses interaction history to
propose tasks in Minecraft. AgentSynth~\cite{xie2026agentsynth} uses
desktop state across the subtasks of one computer-use task.
In \method{}, a verified change to an object becomes part of the next
task's input. That task can request a new room and refer to the same
object there (Sec.~\ref{supp:task:online}).

\section{Task Generation}
\label{supp:task}
Room-to-Task annotates existing scenes and can supply the first task
of a session (Sec.~\ref{supp:task:roomtotask}). For subsequent tasks,
the generator proposes goals using the current world and the effects of
earlier tasks (Sec.~\ref{supp:task:generator}). We illustrate this process
in one 50-round symbolic session and describe the evaluation across houses
and seeds in Sec.~\ref{supp:task:online}.

\subsection{Room-to-Task}
\label{supp:task:roomtotask}

\begin{figure}[t]
    \centering
    \includegraphics[width=\columnwidth]{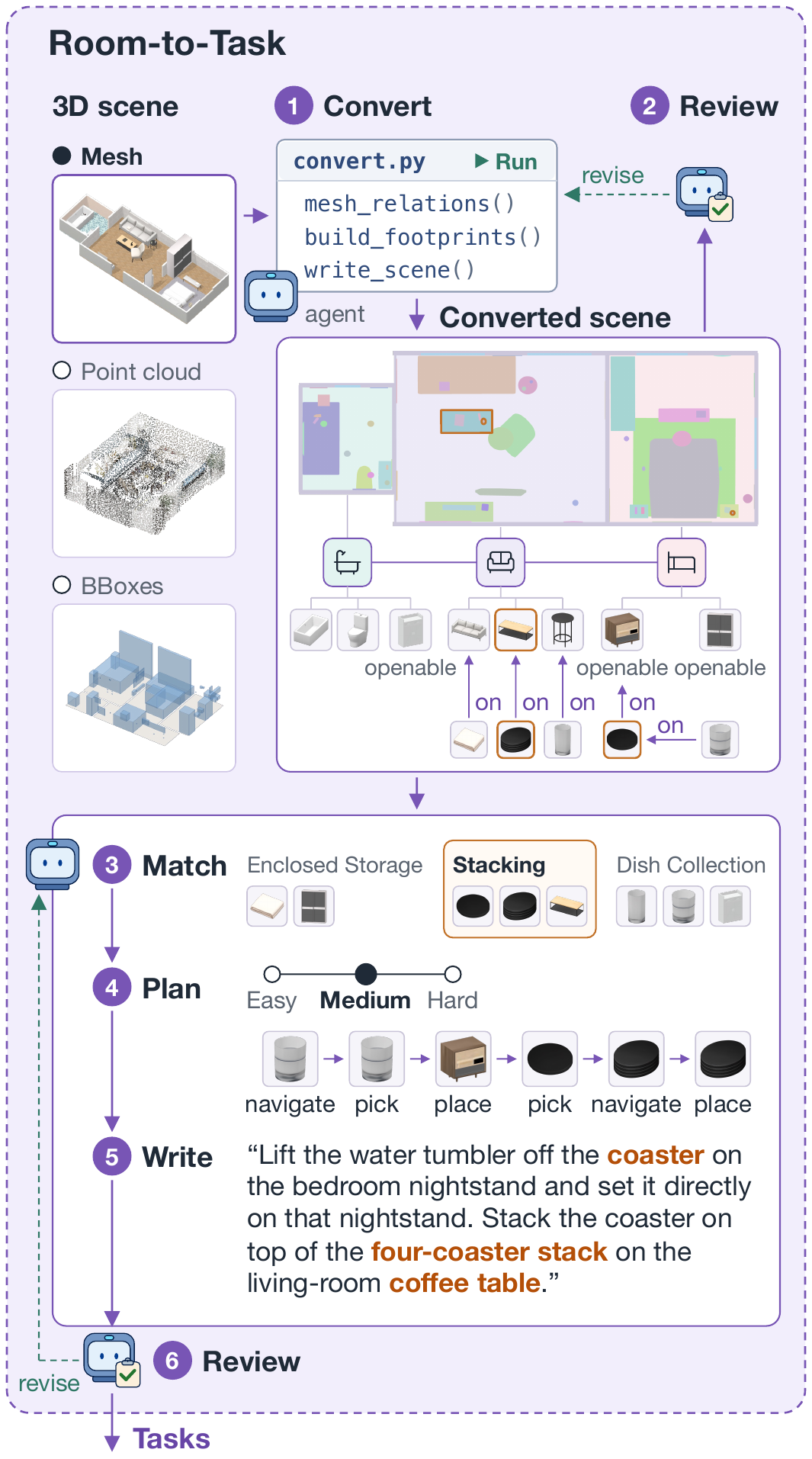}
    \caption{\textbf{Room-to-Task annotation workflow.}
        The first skill converts a scene from a source dataset to a common representation.
        The second skill matches activities to the scene and writes the action sequence and the instruction of each task.
        The reviewer agent checks the output of each skill.}
    \label{fig:room_to_task_pipeline}
\end{figure}

Room-to-Task annotates scenes from existing 3D datasets with tasks $\tau=(s_0,u,\mathbf a)$ as defined in Sec.~\ref{sec:experimental_setting}.
It consists of two agent skills~\cite{anthropic_2025}, each a set of workflow instructions and scripts (Fig.~\ref{fig:room_to_task_pipeline}).
The agent that executes them is GPT-6.1-sol~\cite{openai2026gpt61sol} running in the Codex CLI~\cite{openai_2025}.

\paragraph{Scene conversion.}
The 16 source datasets describe scenes in different forms, such as meshes, point clouds, and bounding boxes (Table~\ref{tab:task_library_sources}).
The first skill converts each scene to a common representation with a top-down layout and a description of the rooms, their connections, and the objects in each room.
For each scene, the agent writes a conversion program.
The conversion retains the object attributes, states, and spatial relations supplied by the source.
When metric geometry is available, it can also infer support from contact between surfaces within 5\,mm, or containment when the entire object fits inside an explicitly defined interior.
A reviewer agent checks the conversion against the source data in a fresh context. The agent that wrote the conversion program corrects errors it can verify.

\begin{table}[t]
    \centering
    \footnotesize
    \setlength{\tabcolsep}{4pt}
    \renewcommand{\arraystretch}{1.10}
    \caption{\textbf{Sources of 480 scenes for annotation.}
        BBoxes denotes bounding boxes and 3DGS denotes 3D Gaussian splats~\citep{kerbl20233d}.}
    \label{tab:task_library_sources}
    \begin{tabularx}{\columnwidth}{@{}lr>{\raggedright\arraybackslash}X@{}}
        \toprule
        Dataset & \#Scenes & Geometry \\
        \midrule
        3D-FRONT~\citep{Fu_2021_ICCV} & 4 & Mesh \\
        3RScan~\citep{Wald_2019_ICCV} & 15 & Mesh, BBoxes \\
        ASE~\citep{avetisyan2024scenescript} & 1 & Point cloud, BBoxes, Parametric layout \\
        HM3DSem~\citep{ramakrishnan2021habitat,Yadav_2023_CVPR} & 3 & Mesh \\
        HSSD-200~\citep{Khanna_2024_CVPR} & 8 & Mesh, Parametric layout \\
        IL3D~\citep{zhou2025il3d} & 46 & Mesh, BBoxes \\
        InteriorGS~\citep{miao2026towards} & 72 & 3DGS, BBoxes, Parametric layout \\
        InternScenes~\citep{zhong2026internscenes} & 26 & Mesh, BBoxes \\
        M3DLayout~\citep{Zhang_2026_CVPR} & 45 & Mesh, BBoxes \\
        MansionWorld~\citep{Che_2026_CVPR} & 28 & Mesh, Parametric layout, Asset bounds \\
        MultiScan~\citep{mao2022multiscan} & 6 & Mesh, BBoxes \\
        SAGE-10K~\citep{sage} & 72 & Mesh, Parametric layout \\
        ScanNet++~\citep{Yeshwanth_2023_ICCV} & 28 & Mesh, Point cloud, BBoxes \\
        SceneSmith~\citep{pfaff2026scenesmith} & 67 & Mesh \\
        SpatialLM~\citep{mao2026spatiallm} & 53 & Point cloud, BBoxes, Parametric layout \\
        Structured3D~\citep{zheng2020structured3d} & 6 & BBoxes, Parametric layout \\
        \bottomrule
    \end{tabularx}
\end{table}

\paragraph{Task annotation.}
The second skill matches scene objects to 22 activities in five families (Table~\ref{tab:room_to_task_activities}).
For each task, the agent selects one of the matched activities, chooses the objects for it, and writes the action sequence $\mathbf a$.
The number of actions determines the difficulty level (1--4 for easy, 5--10 for medium, and 11--20 for hard).

\begin{figure*}[tp]
    \vbox to \textheight{%
        \centering
        \includegraphics[width=\textwidth]{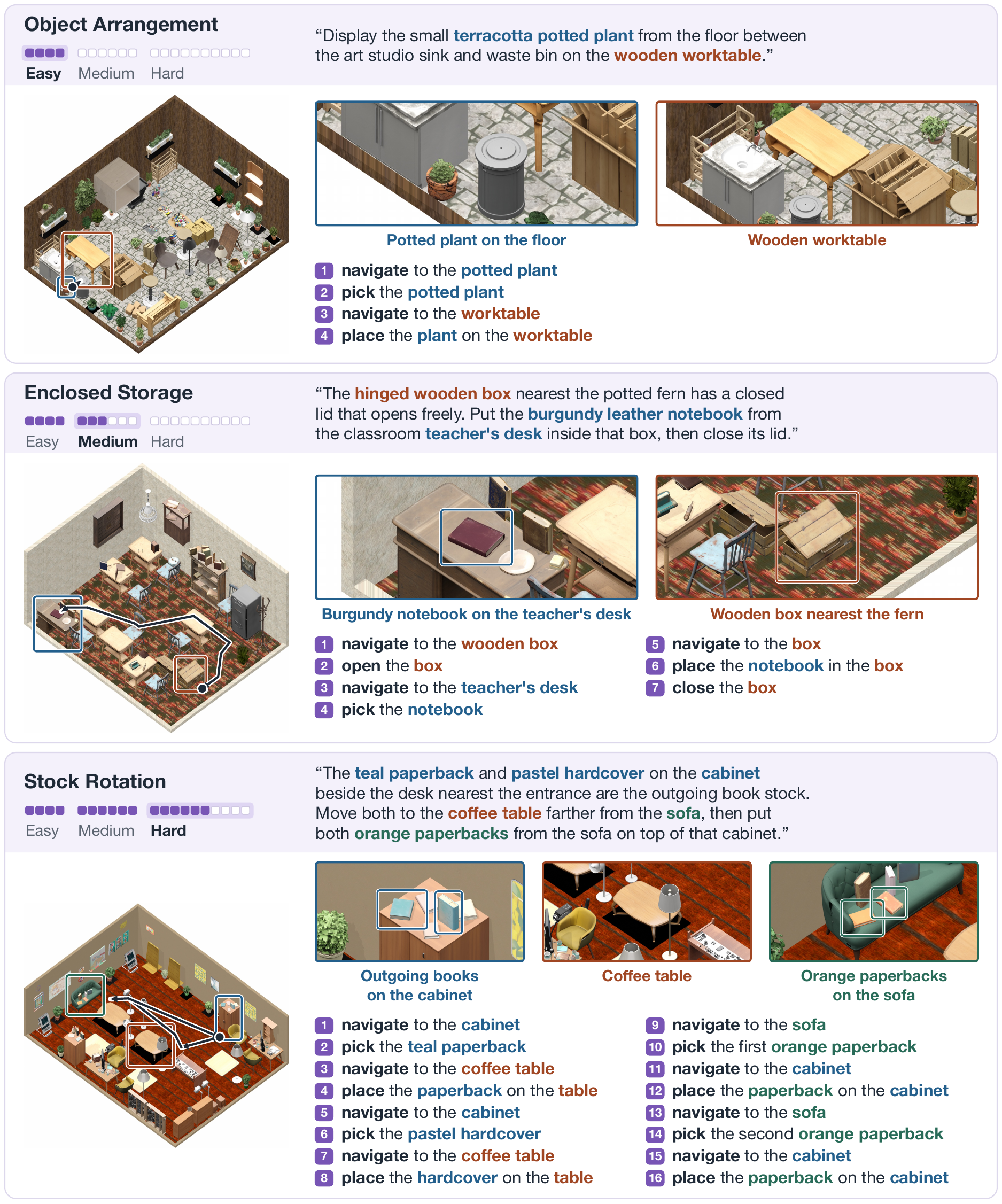}
        \caption{\textbf{Room-to-Task examples.} From top to bottom: Object Arrangement (easy), Enclosed Storage (medium), and Stock Rotation (hard). Each scene view traces the route, and colors link the actions to the objects.}
        \label{fig:room_to_task_generation_examples}
        \vfil
    }
\end{figure*}

The agent then writes the instruction $u$ for the robot, referring to objects by category, appearance, and location.
When a task depends on an object state, the instruction specifies that state as a start condition, such as the closed lid of the box in the medium example of Fig.~\ref{fig:room_to_task_generation_examples}.
The initial state $s_0$ comprises the converted scene, the start room that the agent chooses for the robot, and the start conditions given in the instruction.
Scripts check that each referenced object exists, routes follow connected rooms, and a pick precedes each placement.
A reviewer agent then checks that the sequence carries out the specified activity and agrees with the written instruction.
Tasks that fail either check are revised, replaced, or removed.

\paragraph{Task library.}
We derive 24 functional categories from the primary functions of the scenes in the source datasets.
The categories are bathroom, bedroom, cinema or \mbox{screening} room, corridor or stairwell, dining and drinking space, game room, gym or fitness studio, hair salon, hotel, kitchen, laundry room, library, living room, media production studio, museum or art gallery, office or meeting room, parking garage, patient care space, residence, retail store, school or classroom, scientific laboratory, storage room, and workshop or art studio.
We select 20 scenes per category, which gives 480 scenes from 16 datasets (Table~\ref{tab:task_library_sources}).
Of these scenes, 389 have a single room and 91 have several, the largest with 77 rooms on six floors.
The scene descriptions list 173,578 objects, a median of 69 per scene.

\begin{table*}[t]
    \centering
    \footnotesize
    \renewcommand{\arraystretch}{1.15}
    \setlength{\tabcolsep}{4pt}
    \caption{\textbf{List of activities used by Room-to-Task.}
        The 22 activities are grouped into five families and listed with their procedures.}
    \label{tab:room_to_task_activities}
    \begin{tabularx}{\textwidth}{@{}>{\raggedright\arraybackslash}p{0.125\textwidth}
        >{\raggedright\arraybackslash}p{0.195\textwidth}
        >{\raggedright\arraybackslash}X@{}}
        \toprule
        \textbf{Family} & \textbf{Activity} & \textbf{Procedure} \\
        \midrule
        \textbf{Cleaning} & Surface Wiping & Clear obstructions as needed, wipe, put away the tool, and restore displaced objects. \\
         & Dish Collection & Collect the listed items at the receiving point. \\
         & Machine-Wash Loading & Open the door if closed, load items one at a time, then close it. \\
         & Waste Collection & Open the lid if needed, put the waste inside, and close any lid opened. \\
         & Leftover Disposal & Open lids as needed, pour, set down the source as specified, and close any lids opened. \\
        \midrule
        \textbf{Organization and} & Enclosed Storage & Open the door or lid if closed, store the items, and close it. \\
        \textbf{Storage} & Open Storage & Move each item from its source to its assigned location. \\
         & Sorting & Transfer each item to the destination specified by the rule. \\
         & Stacking & Position the base if needed and stack from bottom to top. \\
         & Quota-Based Restocking & Retrieve and add enough stock to meet the quotas. \\
         & Stock Rotation & Remove all outgoing items, then restock the same location. \\
        \midrule
        \textbf{Food Preparation} & Food Plating & Plate the food as specified, then deliver the loaded dish if serving is required. \\
         & Ingredient Gathering & Gather the listed ingredients at the preparation location. \\
         & Beverage Portioning & Open as needed, pour into assigned cups, and set down the sources as specified. \\
        \midrule
        \textbf{Setup and} & Object Arrangement & Retrieve the listed objects and arrange them as specified. \\
        \textbf{Delivery} & Space Clearing & Move the occupying objects to their assigned destinations. \\
         & Workspace Repurposing & Remove listed old supplies, then arrange new ones on the same surface. Shared items may stay. \\
         & Supply Delivery & Load any missing items and deliver the carrier as a unit. \\
        \midrule
        \textbf{Environmental} & Lighting Configuration & Use the controls to set the requested state or brightness. \\
        \textbf{Control} & Window Opening and Closing & Open or close the panels as specified, using handles as needed. \\
         & Ventilation Configuration & Use the controls to set the requested state or speed. \\
         & Shading Configuration & Open or close each designated component as requested. \\
        \bottomrule
    \end{tabularx}
    \vspace{12pt}
\end{table*}
\begin{table*}[t]
    \centering
    \small
    \setlength{\tabcolsep}{4pt}
    \caption{\textbf{Comparison of task construction methods on 480 scenes.}}
    \label{tab:scene2task_comparison}
    \begin{tabular*}{\textwidth}{@{\extracolsep{\fill}}lrrrrr@{}}
        \toprule
        Method & \#Tasks & Feas. (\%) $\uparrow$ & Clar. (\%) $\uparrow$ & Valid. (\%) $\uparrow$ & Comp@9 (\%) $\uparrow$ \\
        \midrule
        Rule-based & 3,187 & 75.68 & 22.12 & 17.23 & 12.71 \\
        GPT-6.1-sol~\citep{openai2026gpt61sol} & 4,320 & \textbf{98.82} & 51.13 & 50.83 & 50.83 \\
        ManiTaskGen~\citep{dai2025manitaskgen} & 3,032 & 78.69 & 11.25 & 10.06 & 7.06 \\
        RoboLab~\citep{yang2026robolab} & 3,393 & 96.64 & 44.77 & 44.30 & 34.79 \\
        \midrule
        Room-to-Task & 3,581 & 93.44 & \textbf{98.21} & \textbf{92.38} & \textbf{76.57} \\
        \bottomrule
    \end{tabular*}
\par\smallskip
{\footnotesize\raggedright \#Tasks counts delivered tasks. Feas., Clar. and Valid. give the percentages of delivered tasks judged feasible, clear or both, respectively. Comp@9 gives the percentage of all requested tasks that were delivered and valid.\par}
\end{table*}
\begin{table*}[t]
    \centering
    \footnotesize
    \caption{\textbf{Judging feasibility and instruction clarity.} Inputs are a JSON scene description, an SVG top-down layout and a rendering.}
    \label{tab:task_judging_prompts}
    \begin{tabular}{>{\raggedright\arraybackslash}p{\dimexpr\textwidth-2\tabcolsep\relax}}
        \toprule
        \textbf{Feasibility prompt} \\
        \midrule
        \setlength{\parindent}{0pt}
        \setlength{\parskip}{3pt}
        Evaluate the physical feasibility of each listed task from the perspective of a mobile manipulation robot operating in the supplied scene, equipped with a single gripper and a single camera for visual perception. Each task is an independent assignment starting from the same initial scene.

        The scene JSON and SVG provide the main reference for objects, attributes, and spatial relations. Use the scene rendering as an approximate reference for the robot's visual field and how the scene appears; verify specific scene details against the JSON and SVG.

        Consider whether the robot could complete the task in the given scene. Pay particular attention to the objects and space available, the actions the robot would need to perform, and whether the intended result is physically reasonable.

        Assign exactly one label per task, either Yes or No:\par
        - Yes: the evidence supports completing the whole task within the stated capabilities.\par
        - No: a specific conflict with the scene, robot capabilities, or physical constraints prevents completion.

        Apply the same standard to every task independently; all tasks may receive the same label. Other task descriptions cannot establish scene facts or supply missing requirements.
        \\
        \midrule
        \textbf{Clarity prompt} \\
        \midrule
        \setlength{\parindent}{0pt}
        \setlength{\parskip}{3pt}
        Evaluate the clarity of each task from the perspective of a mobile manipulation robot equipped with a single gripper and a single camera for visual perception, relying on its visual observations to understand the instruction and recognise its intended objects. Each task is an independent assignment starting from the same initial scene.

        The scene JSON and SVG provide the main reference for objects, attributes, and spatial relations. Use the scene rendering as an approximate reference for the robot's visual field and how the scene appears; verify specific scene details against the JSON and SVG.

        Consider whether the robot can understand the instruction from its visual perspective in the scene. Being able to locate an object in the supplied scene data does not by itself establish that the robot can identify it from the instruction and observable scene cues. Pay particular attention to whether the robot can tell which objects to interact with, what to do with them, and what result is expected. Look for ambiguities that could lead to different understandings of the task.

        Assign exactly one label per task, either Yes or No:\par
        - Yes: the intended actions, targets or permitted choices, and satisfactory outcome are understandable.\par
        - No: missing or ambiguous wording leaves an essential target, action, or goal materially unclear.

        Apply the same standard to every task independently; all tasks may receive the same label. Evaluate each instruction as written, without borrowing information from another task to complete its meaning.
        \\
        \bottomrule
    \end{tabular}
\end{table*}

For each scene, we request nine tasks, three per difficulty level.
Of the 480 scenes, 375 yield all nine tasks, 55 yield one to eight, and 50 yield none.
For example, one bathroom contains a sink, a toilet, and a light, but the scene identifies neither a movable supply nor a control for the light. None of the 22 activities could be grounded in that inventory.
The resulting library holds 3,581 tasks: 1,234 easy, 1,187 medium, and 1,160 hard.
On average, easy, medium, and hard tasks have 3.6, 7.6, and 14.7 actions, and their instructions have 23, 35, and 48 words.
Easy tasks take place in a single room by design, while 195 medium and 229 hard tasks span several rooms, up to six.
All 22 activities occur in the library, and its actions use all nine operations (navigate, pick, place, open, close, press, rotate, pour, and wipe).
By family, Organization and Storage accounts for 47.0\% of the tasks, Setup and Delivery for 33.2\%, Cleaning for 10.9\%, Environmental Control for 6.3\%, and Food Preparation for 2.6\%.
Fig.~\ref{fig:room_to_task_generation_examples} shows one task per difficulty level.

\paragraph{Annotation evaluation.}
We compare Room-to-Task with four other methods on the 480 scenes (Table~\ref{tab:scene2task_comparison}).
We request nine tasks per scene, 4,320 per method.
All methods take the converted scene as input, and GPT-6.1-sol is the LLM in every method that uses one.
The rule-based baseline fills fixed instruction templates with objects matched by category.
The GPT-6.1-sol baseline prompts the LLM directly for nine instructions per scene.
ManiTaskGen~\cite{dai2025manitaskgen} fills its 200 outcome-based task templates with scene objects.
RoboLab~\cite{yang2026robolab} provides an agent skill that turns a user-given goal in a USD scene into a task definition.
We give this skill the converted scene, let the LLM propose nine goals, and take the default instruction from each resulting task definition.

For each scene, we pool the instructions from all methods, shuffle them, and hide the method names.
Following the LLM-as-a-judge approach~\cite{zheng2023judging}, GPT-6.1-sol judges each instruction using the scene description, top-down layout, and a rendering. The judge assumes a mobile manipulator with one gripper and one camera (Table~\ref{tab:task_judging_prompts}).
For feasibility, the judge asks whether the robot could perform the task with the objects and space available. For clarity, it asks whether the robot could identify the intended objects, actions, and outcome from the instruction and observable scene cues.
A task counts as valid if the judge answers yes to both questions.
Validity divides the valid count by the number delivered. Comp@9 divides that count by all 4,320 requests, including unanswered requests.

The judge rated 3,308 Room-to-Task instructions both feasible and clear, or 92.38\% of the 3,581 delivered and 76.57\% of all requests.
The next highest Comp@9 was 50.83\% for the GPT-6.1-sol baseline. That baseline scored higher on feasibility (98.82\% vs.\ 93.44\%). RoboLab also scored higher on feasibility (96.64\%).
For Room-to-Task, validity was 91.98\% on easy tasks, 92.17\% on medium tasks, and 93.02\% on hard tasks.

\subsection{Task Generator}
\label{supp:task:generator}

\paragraph{Task and world representation.}
The persistent world $W_t$ is a set of records.
An object record in $\mathcal O_t$ holds a unique ID, a class, a pose, a
bounding-box size, and six state fields.
These give the object's support, its container, and whether it is open,
held, wiped, or switched on.
The relations $\mathcal E_t$ are not stored separately; they are the
support and container fields.
A room record in $\mathcal R_t$ holds an ID, a type, a position, a
footprint, and four walls with door and window openings.
A doorway that is not yet connected, faces unoccupied space, and has
not been sealed after repeated refused growths is \emph{free} and can
receive the next room.
The agent state $\xi_t$ is the agent's room and position, and its single
hand holds at most one object.
A goal $\mathcal G_t$ is a conjunction of predicates over object IDs.
Each predicate states that an object is on a support or in a container,
that a surface is wiped, or that a lamp is on or off.

The interaction history $H_t$ is an append-only log with one entry per
committed task.
An entry stores the task identifier, the added room, the IDs of its
objects, and the record of every changed object before and after the task.
Discarded candidates leave no entry.

A proposal has the three fields shown under ``Next Task Script'' in
Fig.~\ref{fig:task_generation}.
The field \texttt{new\_room} is a room type, and \texttt{new\_objects}
lists object classes that the room will contain.
The field \texttt{relations} lists one to three relations (up to five in
a hard round) of the kinds \emph{on}, \emph{in}, \emph{wipe}, and
\emph{switch on/off}.
Each end of a relation names an existing object by its ID or an object
of the new room by its class, a \emph{class reference}.
Every relation is a goal, and none is an initial condition.

Once the room is staged, object-role binding maps each class reference to
an instance of that class in the new room.
It also fixes the role of every object as moved object, support,
container, wiping tool, wiped surface, or switched lamp.
For a placement, binding first tries pairs whose moved object already
rests on or in the target, then pairs within one room.
The first pair that the relation can use wins, so the moved object is
liftable and fits its target.
A class that two relations name resolved to one instance in every task
that the \method{} row of Table~\ref{tab:task_generation_expansion} delivered.
This is what ``that same counter'' in Fig.~\ref{fig:task_generation} requires.

The normalized task $\tau_t=(\Delta_t,\mathcal G_t,u_t)$ of Eq.~\ref{eq:method:task} then holds $\Delta_t$, the goals $\mathcal G_t$ over IDs, and
the instruction $u_t$.
Here $\Delta_t$ is the staged room with its new objects and their initial
placements.
The task also lists the objects it may move and the containers it may
open.
A template renders $u_t$ from the bound goals; the LLM does not write it.
The template describes each object by its class, support, and room at
the start of the task.
Where these do not tell the object apart, it adds one object on or in it.
In the dataset notation $(s_0,u,\mathbf a)$ of Sec.~\ref{sec:experimental_setting}, $s_0$ is the staged world with the robot at its start.
The sequence $\mathbf a$ is the plan of the symbolic solver.
Dataset tasks state their outcome only in $u$.
Table~\ref{tab:supp:task:record} follows a jar and a towel through one
task in the pilot session.
A persistent identifier records where each object originated. For an
object introduced during expansion, the identifier encodes the doorway
through which its room was added, its class, and an instance number.
It stays with the object when the object moves.
The proposer cannot add distractors, occupy supports, or close containers
through auxiliary instructions. Its written rationale does not affect
selection.

\begin{table*}[t]
\centering
\caption{\textbf{Pilot round~8 object bindings.} An existing jar goes in a new cabinet and a new towel goes on a workbench.}
\label{tab:supp:task:record}
\footnotesize
\begin{tabularx}{\textwidth}{@{}>{\raggedright\arraybackslash}p{2.35cm} >{\raggedright\arraybackslash}X@{}}
\toprule
Step & Worked example \\
\midrule
Candidate 2 &
Put the existing jar in a new workroom's cabinet.
Put a new towel on the workbench in that same room. \\
Room and bindings &
The workroom attaches to the library through its south doorway.
The cabinet, towel and workbench each refer to a specific object in the new room.
Furnishing adds a bookshelf and a rug. \\
Selection and plan &
All four candidates passed screening. The seeded order tried candidate~2 first.
The solver placed both objects in nine actions ($d=9$), within the medium target. \\
Committed changes &
The cabinet opened. The jar moved from the existing workshop's workbench into that cabinet. The towel moved from the new workroom floor onto its workbench. \\
\bottomrule
\end{tabularx}
\end{table*}

The initial conditions $\mathcal E_{\Delta,t}$ of new objects are set by
the system, not by the proposal.
In a staged room they are the placements chosen by a box-level furnishing
step, and containers start closed.
When the room is generated in 3D, the Room Designer chooses the supports
(Sec.~\ref{sec:room_design}).
For a given task, such as a Room-to-Task task, the start facts stated in
its instruction take this role.

A task given as free text, such as the user task of Fig.~\ref{fig:pipeline}, becomes the task specification by task parsing.
One LLM call, adapted from the task disassembly of EmbodiedGen
V2~\cite{wang2026embodiedgen}, returns the room type and size, the
objects the task names or cannot do without, their initial relations,
and the goals.
Program checks refuse a parse in which a goal already holds at the
start; a failed parse is refused, never replaced.
In the reported runs this call is the first call of the Room Designer
(Sec.~\ref{supp:world:design}).

The staged world $\widehat W_{t,i}$ of Eq.~\ref{eq:method:identity} is a copy
of $W_t$ with the room of candidate $i$ attached at a free doorway.
The staged room holds the proposed classes as typed boxes and the
furniture that the furnishing step adds for the room type.
The proposal does not choose the doorway; the first free one in a fixed
order is used, and the prompt names the room behind it.
The expanding-world sessions in Table~\ref{tab:task_generation_expansion} propose tasks that use a new
room, and expansion stops when the remaining doors cannot accept an
addition. Section~\ref{sec:task_specification} also defines in-place execution
with $\Delta_t=\varnothing$. The current online runner first considers
tasks on objects in the saved rooms. If none reaches the target tier
and a doorway is available, it also considers expansion and favors the
nearest feasible tier. A run with a specified room-count target defaults
to expansion; the operator can explicitly require either scope. The fixed-world
entry point can also serve a previously generated Room-to-Task list
(Sec.~\ref{supp:task:roomtotask}). Table~\ref{tab:task_generation_expansion} uses the expansion branch
throughout its scored rounds.
In both existing and newly added rooms, the physical online runner proposes
placing objects on supports or in containers. The symbolic proposer also
supports wiping surfaces and switching lamps.

In Eq.~\ref{eq:method:identity}, $\sqcup$ is the disjoint union and $\operatorname{id}(\cdot)$
the set of object IDs.
Two staging refusals enforce Eq.~\ref{eq:method:identity}: a new ID equals an existing one, or
an existing object record differs from $W_t$.
Every other refusal is part of the feasibility check below.
A room or doorway record may change only by two side effects of growth.
A doorway facing the new room becomes connected to it, and a window is
removed from a wall that the new room now shares.
The relations $\mathcal E_t$ carry over unchanged, and so does $\xi_t$
except under the start rule given with the tiers below.
In the symbolic experiment of Table~\ref{tab:task_generation_expansion}, the staged world replaces $W_t$
when its task is committed. An uncommitted candidate and its staged
room are discarded. Physical execution uses the state rules of
Sec.~\ref{supp:execution}.

The stored rounds match these rules.
Per-round worlds were kept only for the three adapted rows of Table~\ref{tab:task_generation_expansion}, which ran the same staging and commit code as the
\method{} row.
Their 90 sessions have 4,500 scheduled rounds, and 4,013 of them committed
a task.
In each of these rounds, the staged world matched $W_t$ in every existing
object record, 1,308,209 in all, and no new ID equaled an existing one.
Of 152,157 existing room records, 6,262 differed.
In 5,033 the only change was a doorway now connected to the new room; in
1,229 it was a window removed from a wall that the new room now shares.
Each commit listed exactly the objects that its round added or changed,
28,291 new objects and 6,225 changes of existing ones in all.
In every round without a commit, including the 56 that discarded every
examined candidate, the stored world remained identical to $W_t$.

\paragraph{History-conditioned proposals.}
Each round sends one text-only request to
GPT-6.1-sol~\cite{openai2026gpt61sol} at reasoning effort \emph{high} for
four candidates, the stacked cards of Fig.~\ref{fig:task_generation}.
The request is repeated only when the reply is empty or contains no task
list, never because a candidate fails a later check.

The prompt has five blocks in a fixed order (rules, world, history, tier,
and aim) and closes with a line that repeats the reply format.
The rules block states the reply schema and the task rules.
The rules list 67 room types plus those already in the house and the 90
object classes that the furnishing step can build.
They also name the 15 small classes that may be moved as new objects.
The 67 room types are the 22 room-level categories of Room-to-Task (its
24 functional categories without the two whole-building ones) and finer
room types from its room-type table and scenes.

The world block lists, for each room, the objects a task may move, with
their support or container, then the supports and the containers with
their open state.
The lamps and their power state follow.
An object is listed only if its description in instructions identifies
it uniquely; nothing else is sampled away or truncated.
The tier block states the target tier $d^\star$ as a task size, with the
action bands of Sec.~\ref{sec:experimental_setting}.
An easy task has one relation between new-room objects or one switch of
a new-room lamp, a medium task two or three relations, and a hard task
four or five.
The aim block calls for variety in object choice. Tasks should carry changed objects onward or use added rooms without undoing earlier changes.

The history block renders the ledger $H_t$ of Sec.~\ref{sec:execution_persistence} as net changes.
Each object that no longer stands where it stood before the first task
that moved it has one line.
The line gives its current place, its original place, and the tasks that
moved it.
Each switched lamp, wiped surface, and added room also has one line.
The block omits changes that later tasks undid. A separate line lists
every lamp ever switched, even if it returned to its original state.

The history block then orders what a new task can build on.
Changed objects that a task can name and carry come first, wiping tools
excepted.
Among them, objects moved by the fewest tasks come first, ties going to
the most recently moved.
Added rooms follow if no later task changed anything in them and they
still hold an object that the world block lists.
In medium and hard rounds, the first four entries are assigned one per
candidate.
The prompt asks each candidate to carry its object on to a new place or
to move an object into the room.
In easy rounds, whose tasks use only the new room, the block ends instead
with the arrangements that earlier tasks built in their rooms, so that
they are not repeated.
No screen checks that a candidate uses its assigned entry.
The symbolic experiment in Table~\ref{tab:task_generation_expansion} passes solver outcomes through the
committed world state and $H_t$.
In the physical loop in Sec.~\ref{supp:execution}, the next proposal
also receives the validator's execution outcome.
A compact feedback block lists validated first-attempt outcomes by
task feature and recent failures with their codes.
An optional log view shows what happened in the last failed rounds and
the last successful one.
The validator's measurements come first: the goal predicates before
and after, the objects that moved, and the failing step, if any.
The robot's own step log follows, quoted, truncated and marked as not
verified; the outcome is always the validator's.
Table~\ref{tab:task_generation_expansion} uses the world state and history but omits the detailed log.

Table~\ref{tab:supp:task:prompt} quotes one excerpt from each prompt block in
round~8 of a pilot on the first evaluation house (seed~0, \emph{medium}
reasoning effort). The scored sessions did not retain their prompts.
Round~6 put the jar on the workshop workbench. The history block calls
this task~7 because it counts the seed task as task~1. It assigns the
jar to the second candidate in round~8. The solver executed that candidate and put the jar in the new workroom's cabinet
(Table~\ref{tab:supp:task:record}).

\begin{table*}[t]
\centering
\caption{\textbf{Proposer input for the jar and towel task.} Excerpts from five blocks of the round~8 pilot prompt.}
\label{tab:supp:task:prompt}
\footnotesize
\begin{tabularx}{\textwidth}{@{}>{\raggedright\arraybackslash}p{2.35cm} >{\raggedright\arraybackslash\ttfamily}X@{}}
\toprule
\normalfont Block & \normalfont Prompt excerpt \\
\midrule
1 Rules & The new room opens off room\_19 (library), through its free doorway. \\
2 World & room\_21 (workshop): movers: room\_18+south/jar\_0 jar on room\_18+south/workbench\_0 [moved] \\
3 History & - room\_18+south/jar\_0 jar: now on room\_18+south/workbench\_0 (room\_21), was on floor (room\_21) (moved by task 7) \\
4 Tier & the task has 2-3 relations, at least one with a new-room end: about 5-10 actions. \\
5 Aim & Spread the tasks over different objects instead of moving the same one again and again. \\
\bottomrule
\end{tabularx}
\end{table*}

\paragraph{Task verification and selection.}
The screens (``Valid Task'' in Fig.~\ref{fig:task_generation}) run on $W_t$ in
the order valid references, required room use, and consistent goals.
The first failure rejects the candidate.
A candidate passes \emph{valid references} if the room type is known, a
free doorway exists, every ID exists in $W_t$, and every class reference
appears in \texttt{new\_objects}.
It passes \emph{required room use} if at least one relation involves an
object of the new room.
It passes \emph{consistent goals} if the relation count is within the
cap, no object receives two end states, and every relation respects the
class rules.
Under these rules, an \emph{on} target is a support and an \emph{in}
target is an openable container.
A moved new-room object is one of the 15 small classes, and a wipe uses
a wiping tool that the same task puts away.
An existing moved object or wiping tool is also liftable, and it rests
on or in another object or has a largest side of at most 0.5\,m.
Fit is checked here only when both ends already exist.
Every existing object has a unique description, and no relation between
existing objects already holds.
Under the no-repeat rule, a candidate identical to a task selected in an
earlier round is also dropped.

The remaining candidates are grouped by a six-field task descriptor.
It records the number of relations (capped at three), whether a relation
joins objects of two rooms, and whether a relation refers to an object's
earlier place (a history query).
It also records how many other objects of the same class share the room
of an existing object that a relation moves, wipes with, or switches
(capped at two).
The fifth field marks a required container recorded as closed. The physical
runner also marks a new insertion target that it initializes closed.
The sixth field records whether a target needs clearing first.
The history and clearing fields are always false in the \method{} row.
Its prompt offers no history query, and the parser drops a proposal that
uses one or that asks for a staged requirement such as a blocked target.
The ``diversity ranking'' in Sec.~\ref{sec:task_generation} (``Rank Tasks'') is a seeded
random order over these groups with a uniform draw inside each group.
It assigns no score, so each group present among the candidates is
equally likely to be examined first, however many candidates share it.

Candidates are examined in this order (``Feasibility Check'').
For each one, the room is staged, the class references are bound, and the
symbolic solver runs on $\widehat W_{t,i}$ under the action model of
Table~\ref{tab:supp:task:actions}.
An action whose precondition does not hold is refused, and the plan fails.
The solver is a fixed reference procedure, not a search.
It orders the goals (wipes, then placements in the listed order, then
switches) and executes one action pattern per goal.
For a placement it goes to the object's room, opens the object's
container if closed, picks the object, and closes the container again.
It then goes to the target's room, opens the target for an \emph{in}
relation, and places.
In the symbolic rounds, a place action searches free spots on a 5\,cm
grid over the top of the support's box or the interior of the container's
box.
A wipe fetches the tool in the same way before wiping, and a switch
presses the lamp.
The procedure stops at the first refusal or after 32 steps, including walks between rooms.

\begin{table*}[t]
\centering
\caption{\textbf{Action preconditions and effects in the symbolic solver.}}
\label{tab:supp:task:actions}
\footnotesize
\begin{tabularx}{\textwidth}{@{}l>{\raggedright\arraybackslash}X>{\raggedright\arraybackslash}p{0.33\textwidth}@{}}
\toprule
Action & Precondition & Effect \\
\midrule
go to room $r$ &
A door path connects the agent's room to $r$. &
The agent is in $r$; doors on the path are open; a held object moves with the agent. \\
open $c$, close $c$ &
$c$ is in the agent's room and belongs to an openable class. &
The open state of $c$ is set or cleared. \\
pick $o$ &
$o$ is in the agent's room; the hand is empty; $o$ is liftable (not a fixture, a floor covering, or a heavy or plumbed class; largest side at most 1\,m); nothing rests on or in $o$; no closed container encloses $o$. &
$o$ is held; its support and container are cleared. \\
place $o$ on $s$ &
$o$ is held; $s$ is in the agent's room and is not wall-mounted; no closed container encloses $s$; a surface of $s$ has a free spot that fits the footprint and height of $o$. &
$o$ is released with support $s$ and the pose of the free spot. \\
place $o$ in $c$ &
$o$ is held; $c$ is in the agent's room, belongs to an openable class, and is open; no closed container encloses $c$; the interior of $c$ has a free spot for $o$. &
$o$ is released with container $c$ and the pose of the free spot. \\
wipe $s$ &
A wiping tool is held; $s$ is in the agent's room and belongs to a support class; no closed container encloses $s$. &
$s$ is wiped; the tool stays in the hand. \\
press $l$ &
$l$ is in the agent's room and is a lamp; no closed container encloses $l$. &
The power state of $l$ toggles. \\
\bottomrule
\end{tabularx}

\vspace{2pt}
{\scriptsize $o$ object, $s$ support, $c$ container, $l$ lamp, each referenced by ID.\par}
\end{table*}

A candidate is feasible only if every step of this check succeeds.
For a feasible candidate, the growth step accepts the room, and binding
finds usable instances.
The staged world does not yet satisfy the goal, and at least one goal
that names a new-room object is still unmet.
After the plan, all goals hold, and only the objects that the task may
change have changed.
A container it may open, a surface it wipes, or a lamp it switches may
change only in that state.

Two box-level guards outside the action model run after staging and
again after the plan.
The first rejects a change that newly blocks a doorway or the front of an
openable object.
The blocker is a floor object or, at an openable object on a
support, an object on the same support.
In the new room, only an object that cannot be lifted counts at a front.
This guard also rejects a change after which a floor object that
was reachable from a doorway no longer is.
The second guard, the physical level, rejects an overlap of two boxes, or
of a box and the floor, that grew by more than 5\,mm.
It also rejects an object sunk into its support.
A settle of the rooms involved, as boxes for 2\,s in
MuJoCo~\cite{todorov2012mujoco}, then rejects an object that newly shifts
more than 0.2\,m or turns more than $8^\circ$.
This settle ran only in the three adapted rows of Table~\ref{tab:task_generation_expansion}, where it rejected none of the 4,839 probed candidates.
The other rows, \method{} included, ran without a simulator, so their
physical level checked box overlap and sinking only.

The solver is sound with respect to the action model, because every
action passes the preconditions of Table~\ref{tab:supp:task:actions}.
It is not complete, since it neither reorders goals nor backtracks.
For example, the tool stays in the hand after a wipe, so a task that
lists another placement before the tool's put-away fails at the next
pick.
Plans are not guaranteed to be shortest.
The action model reasons over boxes and rooms.
It does not model grasps, arm reach, paths inside a room, door swing,
collisions, or stability.
The guards approximate the last four only with the box-level checks
above, and no check uses meshes.
The pour and rotate operations of Sec.~\ref{sec:experimental_setting} have no
counterpart in it.

Difficulty $d$ uses the operation vocabulary of Sec.~\ref{sec:experimental_setting}.
Each pick, place, open, close, wipe, and press of the plan $\pi_{t,i}$ is
one action.
Each object the robot acts on adds one navigate action, unless that
object is beside the one last navigated to and the robot has not changed
room since.
Here \emph{beside} means the same object, its support or container, or
an object resting on or in it.
Walking between rooms is not counted.
A plan of 1--4 actions is easy only if it stays in one room, the start
room included.
A short plan that crosses rooms forms a class of its own between easy and
medium.
Medium is 5--10 actions, hard is 11--20, and a longer plan is out of
range.
Under the start rule, a task whose relations involve only new-room
objects starts with the robot moved into the new room, which lets an
easy task stay in one room.

The target tier $d^\star$ of a round comes from a seeded tier schedule
that cycles through the three tiers in shuffled blocks.
The generator returns the first feasible candidate whose tier equals
$d^\star$ (``Match Difficulty'').
If there is none, it returns the feasible candidate whose tier is
nearest, the lower one on a tie.
In the reported symbolic runs, out of range counted as one step above
hard, allowing an out-of-range plan on a hard target when neither hard
nor medium was feasible. The current generator excludes out-of-range
plans before choosing among the remaining tiers.
The returned task is executed and committed; the round counts as
delivered only if the hand is then empty.
Probing stops at the first match, so the feasible set of Eq.~\ref{eq:method:solver} is not enumerated in full.
For the runs in Table~\ref{tab:task_generation_expansion}, Table~\ref{tab:supp:task:tierhit}
reports how often a delivered task matched $d^\star$.

\paragraph{Optional adaptive selection.}
By default, the physical online runner sends execution feedback to the
proposer and retains the seeded tier schedule.
A competence estimate is the estimated probability that the executing
agent completes a task of one descriptor group.
It is kept as a Beta posterior over the first-attempt outcomes recorded
for that group, with a prior pooled from groups that share descriptor
values.
When adaptive selection is enabled, execution outcomes change two things
and leave the LLM, its prompt, and the screens untouched.
The first is the target tier (the feedback arrow to ``Target'' in
Fig.~\ref{fig:task_generation}).
It moves one step up when the success rate over the last six
first-attempt results at the current tier reaches 0.8, and one step down
when it falls to 0.2.
When the selected task misses its target tier, the outcome enters the
window for the requested tier.
A move needs at least four results and comes at most once every two
rounds, and every sixth round takes a seeded random tier.
The second is the order in which the groups are visited (the arrow to
``Rank Tasks'').
The order is random and favors groups with a high SFL learnability score
$4p(1-p)$~\cite{sfl}, with $p$ sampled from the estimate, and groups with
few recorded outcomes.
Adaptive selection is not used in any experiment in Sec.~\ref{sec:experiments}.
In Table~\ref{tab:task_generation_expansion}, ``fixed competence estimates'' means that
no estimate is updated or consulted.
\subsection{Symbolic Task Generation in Expanding Worlds}
\label{supp:task:online}

\paragraph{Object reuse across tasks.}
In one symbolic session, the seed task moved a bottle from the dressing-room
floor onto a cabinet. Round~3 moved the same bottle to a console table in
the new entryway. Round~4 put it in the new kitchen's fridge and left a pan
on an island in the starting house. Round~5 picked up that pan and placed
it in a cabinet in the new kitchenette (Table~\ref{tab:supp:task:session}).
This session used the first evaluation house and seed~0. After the unscored
seed task, all 50 scheduled rounds met their target difficulty and each
added one room, bringing the total from 16 to 66.

The longest chain of object reuse in this session connects 14 rounds, from
the seed task to round~50 (Table~\ref{tab:supp:task:chain}). Each link
joins a task to the most recent earlier task that moved the same object.
In 10 of the 13 links, the later task found the object in the room added
by the earlier task. All 13 later tasks met the \emph{Prior effect}
criterion when evaluated against the reset world
(Table~\ref{tab:supp:task:metrics}).

\begin{table*}[t]
\centering
\caption{\textbf{Bottle and pan placements in selected rounds of one symbolic session.} Action counts cover each full task.}
\label{tab:supp:task:session}
\small
\begin{tabularx}{\textwidth}{@{}rl>{\raggedright\arraybackslash}p{0.16\textwidth}r>{\raggedright\arraybackslash}X@{}}
\toprule
Round & Target tier & Room added & Actions & Recorded outcome \\
\midrule
0 & Seed & None & 4 & The bottle moved from the dressing-room floor onto a cabinet. \\
3 & Medium & Entryway & 10 & The same bottle moved from the cabinet to the new entryway's console table. \\
4 & Hard & Kitchen & 17 & The bottle moved into the new fridge. A pan from the new kitchen moved to an island in the starting house. \\
5 & Medium & Kitchenette & 9 & The same pan moved from the island into the new kitchenette's cabinet. \\
\bottomrule
\end{tabularx}
\end{table*}

\begin{table*}[t]
\centering
\caption{\textbf{Object reuse in one symbolic session.} Thirteen links connect 14 selected rounds.}
\label{tab:supp:task:chain}
\small
\setlength{\tabcolsep}{5pt}
\renewcommand{\arraystretch}{1.18}
\begin{tabularx}{\textwidth}{@{}>{\centering\arraybackslash}p{1.55cm}>{\raggedright\arraybackslash}p{2.25cm}>{\raggedright\arraybackslash}X>{\raggedright\arraybackslash}X@{}}
\toprule
Rounds & Object & Earlier state & Later state \\
\midrule
$0\to3$   & Bottle          & On cabinet (R11)             & On console table (R18) \\
$3\to4$   & Bottle          & On console table (R18)       & In fridge (R19) \\
$4\to8$   & Bowl            & On counter (R19)             & In microwave (R23) \\
$8\to9$   & Food box        & In cabinet (R23)             & In cabinet (R24) \\
$9\to12$  & Jar             & On shelf (R24)               & On side table (R27) \\
$12\to14$ & Desk lamp       & On desk (R27)                & On desk (R29) \\
$14\to24$ & Book            & On bookshelf (R29)           & On bookshelf (R39) \\
$24\to25$ & Cup             & On desk (R29)                & On kitchen counter (R40) \\
$25\to28$ & Bottle          & In fridge (R35)              & In fridge (R43) \\
$28\to31$ & Fruit container & On kitchen island (R43)      & On kitchen counter (R46) \\
$28\to31$ & Bottle          & In fridge (R43)              & In fridge (R46) \\
$31\to47$ & Fruit container & On kitchen counter (R46)     & On kitchen island (R62) \\
$47\to48$ & Pot             & On kitchen island (R62)      & In cabinet (R63) \\
$48\to50$ & Pan             & On counter with microwave (R63) & In cabinet (R65) \\
\bottomrule
\end{tabularx}
\par\smallskip
{\footnotesize\raggedright R$k$ denotes room $k$ in the session. Both the fruit container and the bottle move again between rounds~28 and~31.\par}
\end{table*}

\paragraph{Protocol.}
The experiment of Sec.~\ref{sec:next_task_proposal} uses ten 16-room evaluation
houses of typed boxes from an earlier symbolic room-growth run.
Each house has a first task for every seed and a free doorway at the
start.
A session starts from a supplied box-level world, not from an empty
$W_0$ (Sec.~\ref{sec:task_specification}).
It copies its house and adds seeded small objects on supports and in
containers, giving start worlds of 113--228 objects.
Round 0 executes an easy Room-to-Task task
(Sec.~\ref{supp:task:roomtotask}) re-instantiated in one room; it is
committed, adds no room, and enters no column.
Rounds 1--50 are scheduled and scored, and the tier schedule of
Sec.~\ref{supp:task:generator} gives them 17 easy, 17 medium, and 16 hard
rounds.
A seed fixes the added objects, first task, shared tier schedule and
selection order. LLM replies are unseeded. We do not retry empty rounds.

\paragraph{Proposer rows.}
All rows share the screens, the no-repeat rule, selection, the symbolic
solver, the commit rule, and four candidate slots per round.
A failed slot is not refilled.
Every LLM call of every row uses the model and reasoning effort of
Sec.~\ref{supp:task:generator}.
Template uses no LLM.
It moves the target of each placement, wipe, or switch template into a
new room of the target's room type and draws four, ignoring history and
tier.
The four prompt rows request four candidates in one call and share the
task rules, world description and target tier
(Table~\ref{tab:supp:task:prompt}). Hard rounds allow up to five
relations. The aim block in \emph{Direct prompt} reads ``Make them
varied.'' \emph{+ Aim} also asks for tasks that reuse earlier changes
and vary the objects they move.

\emph{+ Task list} adds the delivered instructions in chronological
order and adapts AgentSynth's instructions for proposing follow-up
tasks~\cite{xie2026agentsynth}. Its \texttt{PREVIOUS} target restores an
object to the support or container it occupied before its last move.
This adaptation omits AgentSynth's persona, verifier, reviser and
summarizer.
\method{} adds the history of object and room changes to \emph{+ Aim}
and marks moved objects in the world description. On easy rounds it
lists earlier room arrangements so the proposer can avoid repeats.

\paragraph{Adapted rows.}
Voyager~\cite{wang2023voyager}, OMNI-EPIC~\cite{faldor2025omni}, and
SEAgent~\cite{sun2025seagent} receive the task rules, the world and tier
blocks, and the aim block of + Aim in their own prompts, but not the
history block.
A shared compiler call turns each proposal into the task schema or
refuses it.
Voyager keeps its curriculum prompt with completed and failed tasks and
its task-context questions.
Once 15 tasks are completed, its self-questioning context is added with
probability 0.8.
The skill library, critic, and code learning are omitted.
OMNI-EPIC keeps its archive of committed and failed tasks with a weighted
anchor and retrieves up to five similar successes and five failures.
Its model-of-interestingness judgment follows compilation, and a
rejection uses up the slot.
Environment code, RL, and reflection are omitted.
SEAgent keeps one call per round that updates a persistent guidebook and
returns four episodes.
The call sees the previous guidebook, the house, and the last attempt's
instruction and outcome.
It also sees a symbolic caption of how the house then differed from its
start state.
The GUI actor and all training are omitted.
The shared compiler repeats a call after an empty LLM reply, as the
prompt rows do.
The adapted rows do not repeat their own calls after an empty reply, so
such a reply leaves the round undelivered and counts against Yield.

\begin{table*}[t]
\centering
\caption{\textbf{Metric definitions and task subsets for Table~\ref{tab:task_generation_expansion}.}}
\label{tab:supp:task:metrics}
\footnotesize
\begin{tabularx}{\textwidth}{@{}l>{\raggedright\arraybackslash}X>{\raggedright\arraybackslash}p{0.2\textwidth}@{}}
\toprule
Column & Definition & Tasks \\
\midrule
Yield &
Share of scheduled rounds that deliver a task: a candidate passes the screens, repeats no earlier task, is completed by the symbolic solver with the hand empty, is committed, and uses its new room. Its tier need not match the target. A round stays undelivered when no candidate passes the screens, every valid candidate repeats an earlier task, the solver completes none, or the LLM reply is empty. &
All 50 scheduled rounds \\
Prior effect &
Share of tasks whose non-restoring relations cannot start, or receive a different plan from the symbolic solver, on a reset world. The reset world keeps every room, the open states of containers, the task's wiping tool, and the new room; objects of the start house return to their state before the first task, and objects of earlier added rooms to their state as built. Plans are compared by how they stop and as multisets of (action, object, room, source), where the source of a pick is its support or container and that of a place its target, so a changed pick source alone counts. &
Medium/hard tasks \\
Prior room &
Share of tasks in which an object of a non-restoring relation, wiping tools excluded, stands before the task in a room added by an earlier round. &
Medium/hard tasks, once an added room holds objects \\
Top object &
Largest share of tasks that name one object existing before the round, in any role; objects of the task's own new room are not counted. This is the Berger--Parker dominance~\cite{berger1970diversity} with tasks, not namings, as the denominator. &
Medium/hard tasks; at least 8 per session \\
Distinct@16 &
Expected number of distinct structures among 16 tasks drawn without replacement, $\sum_c\bigl[1-\binom{n-n_c}{16}\big/\binom{n}{16}\bigr]$ for $n$ tasks of which $n_c$ have structure $c$~\cite{hurlbert1971nonconcept}. &
Medium/hard tasks; at least 16 per session \\
\bottomrule
\end{tabularx}
\end{table*}

\paragraph{Metrics.}
Each column of Table~\ref{tab:supp:task:metrics} is computed per session.
For the last four columns, we consider tasks delivered in the 33 rounds per
session with a medium or hard target.
Easy rounds are left out because an easy target asks for a task inside
the new room.
Every session supplied enough tasks to compute all four metrics.
A cell averages the seeds of each house and then the ten house means.
The 95\% interval has half-width $t_{0.975,9}\,s/\sqrt{10}$, where $s$ is
the standard deviation of the house means.

A relation is restoring if it held at an earlier point of the session,
such as putting an object back or switching a lamp back, or if it uses
the history query.
The put-away of a wiping tool never counts as restoring.
Prior effect does not count how many earlier tasks a dependence spans.

A structure, the unit of Distinct@16, is a task's canonical graph with
one node per object.
Each node is labelled by the object's age (start house, added in an
earlier round, or new in this round) and by its roles (moved object,
support, container, wiping tool, wiped surface, lamp).
A node also records whether its object lies inside a container.
Directed edges carry the relations, and the number of distinct rooms
among the objects (1, 2, 3 or more) completes the structure.
Object IDs, class names, and room types are not part of it.

\begin{figure*}[t]
\centering
\includegraphics[width=\textwidth]{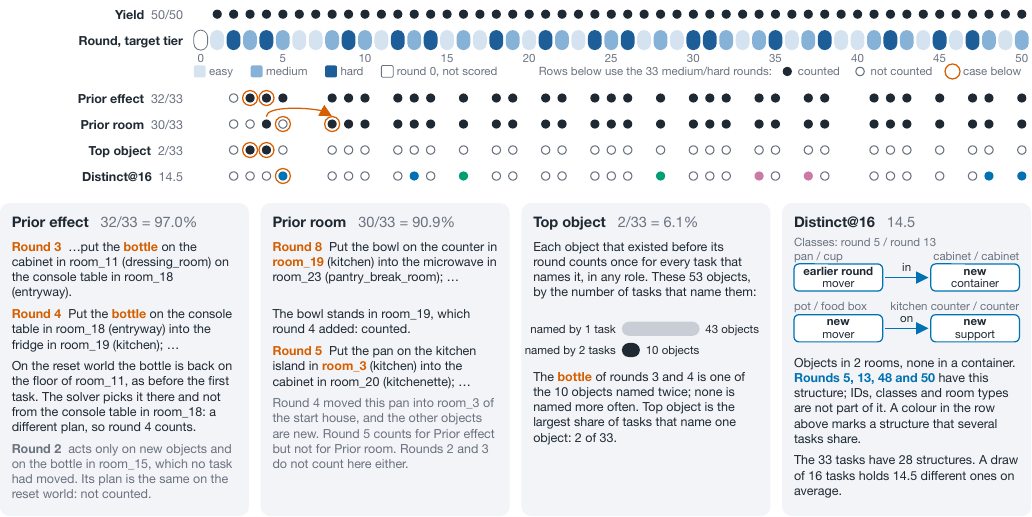}
\caption{\textbf{Task metrics across 50 rounds} of the session in Table~\ref{tab:supp:task:session}. The lower panels show the four metrics for medium and hard targets.}
\label{fig:supp:task:metrics}
\end{figure*}

\paragraph{Reading the columns.}
Fig.~\ref{fig:supp:task:metrics} scores the session of
Table~\ref{tab:supp:task:session}. We use the recorded Prior effect for
each task and reconstruct the other marks from the round records.
An open mark in the Distinct@16 row denotes a structure seen once.
This session delivered 50 of 50 scheduled tasks. Of the 33 tasks with a
medium or hard target, 32 counted for Prior effect and 30 for Prior room.
Round~4 counted for both. In the persistent world the bottle was on the
console table in the entryway added in round~3. In the reset world the
solver had to pick it up from the dressing-room floor. Round~5 counted
only for Prior effect. It retrieved the pan from the original kitchen
where round~4 had left it. The other objects in round~5 belonged to the
new kitchenette.

The most frequently named earlier objects each appeared in two of the
33 tasks (Top object, $2/33=6.1\%$). Rounds 5, 13, 48, and 50 named
four different earlier objects but shared one arrangement of roles and
relations. The 33 tasks contained 28 distinct structures, with 14.5
expected in a sample of 16 (Distinct@16).

\paragraph{Tier hit.}
Tier hit is the share of scheduled rounds that deliver a task in the
target tier.
An undelivered round counts as a miss, so tier hit is at most Yield.
\method{} reached a tier hit of 98.8\%
(Table~\ref{tab:supp:task:tierhit}).
Only one task it delivered fell outside its target tier, a hard round
whose plan exceeded 20 actions.
The other LLM rows reached 86.1--92.3\%, and Template reached 35.5\%.
Except for Voyager, the other LLM rows missed mostly on easy rounds, where
they reached 60.8--83.3\% against 99.8\% for \method{}.
On hard rounds, \method{} (96.7\%) stayed below the other prompt rows and
SEAgent (97.5--98.8\%).
For the first four rows of the table only the per-tier means were kept,
so their per-tier cells have no interval.

\begin{table}[t]
\centering
\caption{\textbf{Tier hit in Table~\ref{tab:task_generation_expansion}.} Percentage of scheduled rounds delivering a task at the target tier.}
\label{tab:supp:task:tierhit}
\footnotesize
\setlength{\tabcolsep}{0.5pt}
\renewcommand{\arraystretch}{1.15}
\begin{tabular*}{\columnwidth}{@{\extracolsep{\fill}}lcccc@{}}
\toprule
Proposer & All & Easy & Medium & Hard \\
\midrule
Template & 35.5$_{\pm 3.7}$ & 27.3 & 77.3 & 0.0 \\
\midrule
Direct prompt & 91.3$_{\pm 1.3}$ & 77.1 & 99.6 & 97.5 \\
\;+ Aim & 90.4$_{\pm 1.6}$ & 73.5 & 99.4 & 98.8 \\
\;+ Task list~\cite{xie2026agentsynth} & 92.3$_{\pm 1.9}$ & 79.2 & 99.4 & 98.8 \\
\midrule
Voyager$^\dagger$~\cite{wang2023voyager} & 88.1$_{\pm 4.0}$ & 88.2$_{\pm 4.3}$ & 87.8$_{\pm 4.7}$ & 88.3$_{\pm 5.8}$ \\
OMNI-EPIC$^\dagger$~\cite{faldor2025omni} & 91.0$_{\pm 1.9}$ & 83.3$_{\pm 3.3}$ & 96.3$_{\pm 2.2}$ & 93.5$_{\pm 3.1}$ \\
SEAgent$^\dagger$~\cite{sun2025seagent} & 86.1$_{\pm 1.8}$ & 60.8$_{\pm 6.3}$ & 99.8$_{\pm 0.4}$ & 98.3$_{\pm 1.5}$ \\
\midrule
\method{} (Ours) & 98.8$_{\pm 0.9}$ & 99.8$_{\pm 0.4}$ & 99.8$_{\pm 0.4}$ & 96.7$_{\pm 2.6}$ \\
\bottomrule
\end{tabular*}

\vspace{2pt}
{\scriptsize Percent; mean $\pm$ 95\% CI over houses. $^\dagger$Adapted proposers.\par}
\end{table}

\paragraph{In-place proposals on saved worlds.}
For each of 30 archived final worlds, we gave GPT-6.1-sol the recorded
outcome and the net object changes between the initial and final states.
One call per world returned four candidate tasks using its existing rooms.
All 120 candidates parsed, and the symbolic solver completed 117.
Of those 117, 53 involved more than one room and 115 referred to an
object whose placement differed between the two states.

\section{Task-Conditioned World Generation}
\label{supp:world}
When a task requires a new room, \method{} furnishes it and connects it to
the existing world. We describe the five stages, evaluate the resulting
rooms (Sec.~\ref{supp:world:table1}), and measure how refurnishing
earlier rooms affects later tasks (Sec.~\ref{supp:world:preserve}).
\subsection{Room Design}
\label{supp:world:design}

The Room Designer (Sec.~\ref{sec:room_design}) turns one task into a
room design, the scene script of Fig.~\ref{fig:pipeline}.
It runs a fixed sequence of LLM calls, with a conditional revision
after critique (Table~\ref{tab:supp:world:designpipeline}).
Every call uses GPT-6.1-sol~\cite{openai2026gpt61sol} at reasoning
effort xhigh, through its API without fine-tuning.
A checking program tests every answer before the next call.
The Room Designer produces a text plan; object sizes, poses and meshes are
supplied by the later stages. Table~\ref{tab:supp:world:designpipeline}
follows design (a) of Table~\ref{tab:supp:world:design}.

\begin{table*}[t]
\centering
\caption{\textbf{Room Designer workflow for placing a wineglass on a living room side table with Medium furnishing.}}
\label{tab:supp:world:designpipeline}
\small
\setlength{\tabcolsep}{5pt}
\renewcommand{\arraystretch}{1.12}
\begin{tabularx}{\textwidth}{@{}>{\raggedright\arraybackslash}p{1.85cm}>{\raggedright\arraybackslash}X>{\raggedright\arraybackslash}p{5.15cm}@{}}
\toprule
Stage & Main operation & Example / outcome \\
\midrule
\textbf{Task analysis}
& Identify required objects, dependencies, initial relations and goals.
& Wineglass and side table: required. Coffee table: initial support. \\
\addlinespace[4pt]
\textbf{Furnishing}
& Run stages up to the requested level; preserve task objects and initial states.
& Low: +6 objects. Medium: +2 objects. High: skipped. \\
\addlinespace[4pt]
\textbf{Critique}
& Accept if all six scores reach 7 out of 10, with no major/critical issues.
& Six scores of 10; accepted. \\
\addlinespace[4pt]
\textbf{Revision}\newline (if needed)
& Revise optional objects and task-object descriptions once.
Keep the revision if the total score does not decrease; otherwise keep the original.
& Not invoked. \\
\addlinespace[4pt]
\textbf{Scene script}
& Assemble the validated scene script for WorldLayout.
& 3 task objects + 8 optional objects. Geometry is generated later. \\
\bottomrule
\end{tabularx}
\par\smallskip
\begin{minipage}{\textwidth}
\small
\textbf{After each call:} validate the answer; allow at most two repairs before failure.
Full checks and call specifications: Table~\ref{tab:supp:world:designcalls}.
\end{minipage}
\end{table*}

\paragraph{Input.}
The Room Designer receives the normalized task as text (room type,
instruction and the objects the room is to contain) and the furnishing
level.
The analyst sees only the text; the level decides which designers run
and appears in the later messages.
The available doorway of an added room is not shown to the Room Designer and
enters only when the design is bound to the world.

\paragraph{Prompts.}
Each call has its own system prompt and a JSON user message, the
``Sees'' column of Table~\ref{tab:supp:world:designcalls}.
At Medium the calls are the analyst, Low, Medium and the critic, followed
by the reviser and a second critic call when the first design is not
accepted.
The analyst prompt extends the task disassembly of EmbodiedGen
V2~\cite{wang2026embodiedgen}.
The designer, critic and reviser prompts apply the furniture agent of
SceneCode~\cite{wang2026scenecode} to a plan without sizes or positions.
Later messages give the task, room and each object's role, stage, support and moving-part flag.

Each prompt fixes a JSON answer schema (``Returns'') and states its rules
in plain words with household examples
(Table~\ref{tab:supp:world:designprompts}).
The prompts call the Low, Medium and High levels tiers easy, medium and
hard, and the moving-part flag ``articulation''.
The program checks supports and room bounds. The prompt requests at most 15 words per object description.
The program rejects descriptions over 20 words. The critic rates style, density and walkways.

The program also tests a plan contract.
It requires no object sizes or positions, task-relation endpoints among
the task objects, and one start support per object with no support cycle.
Open and closed states agree with the moving-part flags, an operational
task has goals and none of them holds at the start, and no object starts
\emph{in} a container.
A preflight of the mesh stage checks that the room shell and its door can
be built.

\begin{table*}[!tbp]
\centering
\caption{\textbf{Room Designer LLM calls in execution order.}
Each row lists inputs, outputs, prompt rules and output checks.}
\label{tab:supp:world:designcalls}
\scriptsize
\setlength{\tabcolsep}{3pt}
\renewcommand{\arraystretch}{1.12}
\begin{tabularx}{\textwidth}{@{}>{\raggedright\arraybackslash}p{1.15cm}>{\raggedright\arraybackslash}p{2.35cm}>{\raggedright\arraybackslash}p{2.75cm}>{\raggedright\arraybackslash}X>{\raggedright\arraybackslash}p{3.65cm}@{}}
\toprule
Call & Sees & Returns & Key rules of the prompt & Program checks on the answer \\
\midrule
Task analyst
& The task text.
& Task ID, kind (operational or furnishing), task type, robot; room type, style, width, depth, height and one door (wall, offset); prompt type (tight or open); task objects (ID, category, description, role, reason, moving-part flag and initial state, FLOOR or ON with parent); initial relations (on, open, closed); goals (on, in, open, closed); assumptions. Or an error.
& No object sizes or positions. A rectangle sized as a real room of its type with walkways for a 0.5\,m wide robot, usually 3--6\,m a side and 2.6--3.2\,m high; one door at least 0.6\,m from the corners. Required: every object the task mentions and every goal endpoint. Dependency: an object the task cannot do without, with its causal reason. No furniture that only completes the room. A description of at most 15 words, of the object alone. The moving-part flag only where the task operates a door, drawer, lid or flap. Two support levels; a start inside a container returns an error; no wall or ceiling objects. Goals of an operational task are false at the start. Tight when the task restricts the room. For the rooms of Table~\ref{tab:task_readiness_expansion}, the room-only addendum of Table~\ref{tab:supp:world:designprompts}.
& Field types and values; sides 2.0--12.0\,m, height 2.2--4.5\,m; a door nearer than 0.6\,m to a corner is moved to 0.6\,m; at most 20 words per description; a reason for every task object; each initial \emph{on} relation equals the object's ON edge; moving-part evidence; relation graph; plan contract and preflight. \\
\midrule
Low designer
& Task text, kind, level, prompt type, room with floor area and door, every object so far (ID, category, description, role, stage, support, moving-part flag, reason of task objects), initial relations, goals, stage.
& New objects (ID, category, description, FLOOR or ON with parent, reason). An empty list is valid.
& Every stage: add only objects of its stage and never change, repeat or remove one; design for the room type and style; nothing a tight prompt does not allow; the stage follows kind and size, never relabelled; nothing on an object the task moves, opens or closes; at most 2 added objects on a task surface; no second object of a required object's category; 0.7--1\,m walkways for a 0.5\,m robot; descriptions of at most 15 words in one style; identical items share category and description; no wall or ceiling objects, people, animals or room structure. Low: floor-standing furniture, large appliances and rugs, all FLOOR; about 3--6 pieces below 10\,m$^2$, 7--12 up to 20\,m$^2$ and 12 or more above, floor-standing task objects included; complete functional groups.
& Field types; at most 20 words; unique IDs; moving-part flag set to false; relation graph; Low objects FLOOR, Medium and High objects ON a floor-standing object; nothing on a goal subject and at most 2 added objects on a goal target or on a goal subject's start support (both read from the goals, which the runs of Table~\ref{tab:task_readiness_expansion} had removed); no category of a required object; plan contract (an optional object's level not below its support's) and preflight. \\
\midrule
Medium designer
& As Low.
& As Low.
& Objects of about 0.3\,m or more on top of a floor-standing piece with a usable top, such as a table, desk, counter, dresser, nightstand or shelf unit; usually 1--2 on a suitable surface and not on every surface; room left on a small task surface for the High stage.
& As Low. \\
\midrule
High designer
& As Low.
& As Low.
& Clutter under about 0.3\,m on top of a floor-standing piece, never on a Medium or High object; some on the task surfaces within the limit, the rest a few per surface. Not run at Medium, the level of every room of Table~\ref{tab:task_readiness_expansion}.
& As Low. \\
\midrule
Critic
& As the designers, with the stages that ran.
& Six scores from 0 to 10; issues (severity critical, major or minor; object IDs; problem; fix); verdict accept or revise.
& Judge only what the plan decides, since sizes and positions do not exist yet. Feasibility: goals false at the start and achievable, plausible supports, nothing on a moved object, no second object of a required category. Completeness for the level: Low the large furniture, Medium also objects on some surfaces, High also clutter; over-filling a small room is as wrong as an empty large one. Robot space: 0.5\,m walkways from the door to every task object. List only critical issues while one exists, then major ones, minor ones only when nothing else; a concrete minimal fix that names the object; task objects cannot be removed. Accept when every score is at least 7 and no issue is critical or major.
& Six scores in 0--10; severities and verdict from their sets; acceptance recomputed from the scores and severities, the verdict itself unused. \\
\midrule
Reviser
& As the critic, with the critique verbatim.
& The complete list of optional objects with their stages; new descriptions for task objects; one line per change.
& Address the critical and major issues as the critique proposes; leave correct objects as they are; act on a minor issue only when it costs nothing; a kept object keeps its ID; the stage rules hold; task objects cannot be removed or renamed and keep their relations; a moving part stays named in the description.
& Stages among those that ran; the designers' checks on the new list; new descriptions only for task objects, at most 20 words each; task objects otherwise unchanged. The critic scores the revision, which is kept if the six scores do not drop in total. \\
\bottomrule
\end{tabularx}
\end{table*}

\paragraph{Task analysis.}
The first call, the task analyst, returns the room (type, style, width,
depth, height and one door) and only the objects the task needs.
Required objects are named by the task or appear in a goal.
Dependencies are objects the task does not name but cannot do without,
such as the surface an object starts on, and each comes with its reason.
Every object has a category, a visual description and a support.
The analyst also labels the task operational or furnishing and lists
what it inferred without evidence as assumptions.
It marks the prompt tight when the task restricts the room (``only'',
``exactly''); the designers then add nothing the task does not allow.

\begin{table*}[!tbp]
\centering
\caption{\textbf{Verbatim analyst and critic prompts.} Excerpts include the analyst addendum used for Table~\ref{tab:task_readiness_expansion}.}
\label{tab:supp:world:designprompts}
\fontsize{6.5}{7.3}\selectfont\frenchspacing
\setlength{\tabcolsep}{3pt}
\begin{tabularx}{\textwidth}{@{}>{\raggedright\arraybackslash}p{1.45cm}>{\raggedright\arraybackslash\ttfamily}X@{}}
\toprule
\normalfont Prompt & \normalfont Text \\
\midrule
\normalfont Task analyst & \textrm{[\ldots]} Given one household task, decide the room the task happens in and every object the task NEEDS. Three designers then add the rest of the room by difficulty tier, and \textrm{[WorldLayout]} decides every size and position later: give neither sizes nor positions of objects. \newline \textrm{[\ldots]} \newline If the task cannot be resolved into such a room, return \{"error": "<reason>"\}; never switch to another task. \newline \textrm{[\ldots]} \newline - The room type comes from task evidence; record a guess in assumptions. Size the room as a real room of that type that holds the task's furniture with walkways for a 0.5 m wide mobile robot: usually 3-6 m a side, 2.6-3.2 m high. \newline - One door, its centre at least 0.6 m from both corners of its wall. The robot enters through it. \newline Objects: what the task needs, nothing else \newline - required: every object the task mentions and every endpoint of a goal. \newline - dependency: an object the task does not mention but cannot do without, with its causal reason (the counter a cup starts on, the shelf a book is taken from). Infer dependencies; never leave one out. \newline - Do NOT add furniture that only makes the room complete: the designers do that, tier by tier. \newline \textrm{[\ldots]} \newline - articulation: true only for an object with functional moving parts (hinged door, drawer, lid, flap) that the task itself opens or closes, or puts something into or takes something out of through that part (the fridge the robot opens, the drawer a spoon goes into). An open container (bowl, basket, tray, open box, sink, open shelf) is false. \textrm{[\ldots]} An articulated object's description names the moving part the task uses (a white wooden cabinet with two closed hinged doors; a nightstand with one drawer). \newline - Relation graph (EmbodiedGen V2's scene graph): every object rests on exactly one thing. "FLOOR" with parent "floor" for an object standing on the floor; "ON" with the parent's id for an object resting on top of another object, and that parent stands on the floor (at most two levels: room \textrightarrow{} furniture \textrightarrow{} object). "INSIDE" (in a cabinet, on an inner shelf) cannot be prepared: if the task needs an object to START inside a container, return an error. No wall-mounted or ceiling objects. \newline \textrm{[\ldots]} \newline - operational: every goal is FALSE before execution (a cup destined for the table starts elsewhere, e.g. on the counter; a door the robot opens starts closed). furnishing: goals may already hold after layout. \\
\midrule
\normalfont Room-only addendum (rooms of Table~\ref{tab:task_readiness_expansion}) & This run builds the room for evaluating room generation; no robot executes the task. Never return an error for something the task's room cannot express: build the closest room you can and record every simplification in assumptions. \newline - The robot does not run, but write the goals as for a robot (on, in, open, closed; an action such as wipe, pour or press has none). They show which objects the task opens or puts things into, and the program drops them before layout. \newline \textrm{[\ldots]} \newline - An object that must start inside a container rests ON that container only if it fits on the container's top (a book on a box, a cup on a basket); anything wider (a plant in a pot) stands beside it ON the same furniture or FLOOR. \\
\midrule
\normalfont Critic & \textrm{[\ldots]} You apply SceneCode's furniture critic to a room plan before layout. \textrm{[\ldots]} Sizes and positions do not exist yet (\textrm{[WorldLayout]} decides them): judge only what the plan decides. \newline \textrm{[\ldots]} \newline - task\_feasibility: the goals are false before execution (operational) and achievable; supports are plausible (the cup starts on the counter); nothing rests on an object the task moves or opens; no second object of a required object's category. \newline \textrm{[\ldots]} \newline - completeness: the room feels finished FOR ITS TIER. easy: the large furniture of the room type; medium: plus medium objects on some surfaces; hard: plus small clutter. Over-filling a small room is as wrong as an empty large one; a tight prompt is complete with what it allows. \newline \textrm{[\ldots]} \newline - robot\_space: for this floor area, the floor-standing furniture still leaves a 0.5 m wide robot walkways from the door to every task object. \newline Rules (SceneCode's critic) \newline - While a critical issue exists, list only critical issues; then major ones; minor ones only when nothing else. \newline - A fix is concrete and minimal: name the object id and the change. Near capacity, prefer replacing to adding; "X or Y" means one of them. \newline - Required and dependency objects cannot be removed: fix around them (their descriptions may change). \newline - verdict "accept" when every score is at least 7 and no issue is critical or major, else "revise". \\
\bottomrule
\end{tabularx}

\vspace{2pt}
{\scriptsize [\ldots]\ text left out; hard-wrapped lines are joined; [WorldLayout] replaces the code name of that stage.\par}
\end{table*}

\paragraph{Moving parts and initial relations.}
An object whose door, drawer, lid or flap the task opens, closes, or uses
to put something in or take it out carries a moving-part flag and its
initial state.
An open container such as a bowl or a basket does not.
The checking program reads this evidence from the relations.
An object that a relation opens or closes needs the flag, and so does
closed furniture, such as a cabinet or a fridge, that a goal fills.
A flagged object's description names its moving part.
An answer that breaks either rule fails.
The initial relations (on, open, closed) describe the start, in which no
goal of a manipulation task may hold yet.
Later answers can only add, replace or remove optional objects and
rewrite the descriptions of task objects, so task objects keep their
roles, supports and initial states.

\paragraph{Staged furnishing.}
Three designer calls add optional objects, following the solver layers
of EmbodiedGen V2~\cite{wang2026embodiedgen}.
Low adds floor-standing furniture, large appliances and rugs, about
3--6 pieces in all below 10\,m$^2$, 7--12 up to 20\,m$^2$ and 12 or more
above.
Medium adds objects of about 0.3\,m or more on furniture tops, and High
adds smaller clutter; each level includes the stages below it.
The designers are told to keep the task possible.
Nothing rests on an object the task moves, a task surface holds at most
two added objects, and no added object shares the category of a required
object.
The program tests the first two rules against the goals.
In the runs of Table~\ref{tab:task_readiness_expansion} the goals were already removed.
These two rules then rested on the prompt and the task text, and only
the category rule was tested.
The level is set independently of the task's difficulty.
Every room of Table~\ref{tab:task_readiness_expansion} was designed at Medium.

\paragraph{Critique and revision.}
A critic following SceneCode~\cite{wang2026scenecode} scores six aspects
from 0 to 10: prompt following, task feasibility, realism, completeness
for the level, moving-part flags, and space for the robot.
It accepts when every score is at least 7 and no issue is critical or
major.
The program applies this rule to the scores and severities and does not
use the critic's own verdict.
Otherwise a reviser may replace optional objects and rewrite the
descriptions of task objects.
It returns the complete list of optional objects, and the list goes
through the designers' checks.
The critic reviews the revision, which is kept and delivered only if the
total of the six scores does not drop.
If the total drops, the revision is undone and the first design is
delivered; no second revision follows.
The critic accepted the initial design in 29 of 150 single-room tasks
(Panel~A) and 253 of 460 room expansions (Panel~B)
(Table~\ref{tab:supp:world:designcritic}).
After revision, another 6 and 53 designs met the acceptance criteria,
bringing the totals to 35 and 306. Designs that passed the structural
checks could still proceed with unresolved critic concerns.

\begin{table}[!tbp]
\centering
\caption{\textbf{Critic outcomes from the design records for all rooms in Table~\ref{tab:task_readiness_expansion}.}}
\label{tab:supp:world:designcritic}
\footnotesize
\setlength{\tabcolsep}{4pt}
\begin{tabular}{@{}lrr@{}}
\toprule
Outcome & Panel~A & Panel~B \\
\midrule
Designs                                 & 150 & 460 \\
Accepted in the first review            & 29  & 253 \\
Revision proposed                       & 121 & 207 \\
\quad Kept                              & 74  & 148 \\
\qquad accepted in the second review    & 6   & 53  \\
\quad Undone                            & 47  & 59  \\
\midrule
Accepted at delivery                    & 35  & 306 \\
Not accepted at delivery                & 115 & 154 \\
\bottomrule
\end{tabular}
\end{table}

\begin{table*}[!tbp]
\centering
\footnotesize
\setlength{\tabcolsep}{4pt}
\begin{tabularx}{\textwidth}{@{}lX@{}}
\toprule
\multicolumn{2}{@{}l}{\textbf{(a) The wineglass task of Fig.~\ref{fig:pipeline}: accepted at the first critique}} \\
\midrule
Task (input) & ``A living room. Put the wineglass in the living room on the side table in the living room. The room must contain: wineglass, side table.'' Furnishing level: Medium. \\
Analyst & Kind operational, task type ``pick and place'', prompt type open. Living room, ``warm contemporary'' style, 4.5\,$\times$\,4.0\,$\times$\,2.8\,m, one door on the south wall, its centre 1.0\,m from the west corner. Task objects: wineglass (required, on the coffee table), side table (required, floor), coffee table (dependency, floor). Initial relation: wineglass on coffee table. Goal: wineglass on side table. Assumptions: ``The wineglass initially rests on a coffee table because its starting support is unspecified and the destination must initially be unoccupied by it.'' ``The room style, dimensions, and door location are inferred.'' \\
Checking program & Found no object that needs moving parts. For the room-only run, the goal was then removed and the kind set to furnishing. \\
Low, Medium & Low added the sofa, armchair, media console, bookcase, floor lamp and rug. Medium added the television on the media console and the potted plant on the bookcase. \\
Critic & All six scores 10 and no issue: accepted, so no revision ran. Each of the four calls passed the checks with its first answer. \\
Room (bound) & 4.5\,$\times$\,4.0\,$\times$\,3.0\,m, entrance on the north wall (the attach point), free doorways on the south, east and west walls. \\
\end{tabularx}
\begin{tabularx}{\textwidth}{@{}llllX@{}}
\midrule
Object & Role & Added by & Rests on & Description \\
\midrule
wineglass & required & analysis & coffee table & Clear glass wineglass with a slender stem and classic rounded bowl. \\
side table & required & analysis & floor & Dark walnut wood side table with a contemporary frame and empty tabletop. \\
coffee table & dependency & analysis & floor & Natural oak wood coffee table with minimalist legs and an unadorned tabletop. \\
sofa & optional & Low & floor & Warm beige woven-fabric sofa with rounded arms and walnut legs in contemporary style. \\
armchair & optional & Low & floor & Caramel leather armchair with walnut legs and softly curved contemporary lines. \\
media console & optional & Low & floor & Low walnut media console with matte cream doors and clean contemporary lines. \\
bookcase & optional & Low & floor & Slender walnut bookcase with open shelves and a warm contemporary silhouette. \\
floor lamp & optional & Low & floor & Straight vertical brushed-brass floor lamp with an ivory linen shade and contemporary detailing. \\
rug & optional & Low & floor & Rectangular ivory and terracotta wool rug with a subtle contemporary geometric pattern. \\
television & optional & Medium & media console & Black flat-screen television with slim bezels and a freestanding contemporary pedestal base. \\
potted plant & optional & Medium & bookcase & Leafy green plant in a matte terracotta pot with a warm contemporary silhouette. \\
\end{tabularx}
\begin{tabularx}{\textwidth}{@{}lX@{}}
\midrule
\multicolumn{2}{@{}l}{\textbf{(b) A bedroom task of another Panel~B sequence: revised once, revision kept}} \\
\midrule
Task (input) & ``A bedroom. Put the vase in the bedroom on the nightstand in the bedroom. The room must contain: vase, nightstand.'' Furnishing level: Medium. \\
Analyst & Bedroom, ``warm minimalist'' style, 4.0\,$\times$\,3.6\,$\times$\,2.8\,m. Task objects: vase (required, floor), nightstand (required, floor). Goal, removed before the designers ran: vase on nightstand. Assumption: ``The vase initially stands on the floor rather than on the nightstand.'' \\
Low, Medium & Low added the bed, wardrobe, rug and floor lamp. Medium returned an empty list. \\
Critic, design & Completeness scored 6 and the other five criteria scored 10. Every object stood on the floor. The critic recommended replacing the floor lamp with a compact brass table lamp on the nightstand, leaving space for the vase. \\
Reviser & Placed the table lamp on the nightstand and revised the nightstand description to account for the lamp on its surface. \\
Critic, revision & All six scores 10 and no issue: accepted. The total rose from 56 to 60, so the revision was kept. \\
Scene script & Vase, nightstand, bed, wardrobe and rug on the floor; the table lamp (Medium) on the nightstand. \\
\bottomrule
\end{tabularx}
\caption{\textbf{Two Panel~B designs from task to scene script.} Tasks and quotations are verbatim. Other entries summarize design decisions.}
\label{tab:supp:world:design}
\end{table*}

\paragraph{Supports and checks.}
The relation graph has two supports, the floor and the top of a
floor-standing object (Sec.~\ref{sec:room_design}).
No object can start inside a container, hang on a wall or ceiling, or
rest on a third level.
The checking program enforces these supports, the room bounds (sides of
2.0--12.0\,m, height 2.2--4.5\,m) and the moving-part flags.
A failed answer returns to its step with its errors and the previous
answer, at most twice, and the design then fails with no fallback.

\paragraph{Unsupported starts.}
For a room meant for robot execution, the analyst is told to return an
error when the task requires an object to start in a container.
The design then ends without a room.
The rooms of Table~\ref{tab:task_readiness_expansion} were built for evaluation without a robot.
For them, the analyst was told instead to build the closest start the
graph can express and to record each simplification as an assumption.
An object then rests on its container if it fits and beside it
otherwise.
A hanging object stands on the floor or on its furniture, and a deeper
stack becomes a row.

\paragraph{Room structure.}
A room built on its own, as in Panel~A of Table~\ref{tab:task_readiness_expansion}, keeps its designed
size and its 0.9\,m\,$\times$\,2.1\,m door.
An added room attaches at a free doorway of the existing world
(Sec.~\ref{supp:task:generator}), the ``Attach Point / Available
Doorway'' of Fig.~\ref{fig:world_generator}.
It keeps the designed type and, under the grid rules of
Sec.~\ref{supp:world:integration}, the designed extent along the
direction of growth.
The footprint, height and doorways set by the world replace the designed
rectangle, height and door.
Its openings become the doors of the design: the attach point as
entrance, openings to adjoining rooms, and new free doorways.
The first room of a Panel~B sequence is the initial room of its world,
and the design is bound to it in the same way.

\paragraph{Worked example.}
Table~\ref{tab:supp:world:design} follows two Panel~B designs from the
task text to the scene script.
Both used the Medium furnishing level and the prompts in
Table~\ref{tab:supp:world:designprompts}.
The wineglass design was made for the fourteenth task of one sequence.
Its analyst added the coffee table as the start support, so that the goal
did not hold before execution.
The recorded reason reads ``Provides the wineglass's initial support,
separate from its destination.''
Low and Medium added eight optional objects, and the critic accepted the
first design.
In the bedroom design, Medium added nothing, and the critic scored
completeness 6 for a Medium room without objects on surfaces.
The reviser replaced the floor lamp with a table lamp on the nightstand,
as the critique proposed.
The revision scored 10 on all six aspects and was kept.

\paragraph{Hand-over to WorldLayout.}
WorldLayout receives the room type, the floor rectangle, the wall height
and the door openings, the entrance among them.
Each object comes with its ID, category, description, support, role and
moving-part flag.
WorldLayout keeps every required and dependency object and may leave out
only optional ones.
The initial \emph{on} relations become hard placement constraints.
No goal becomes a placement constraint, and in the runs of Table~\ref{tab:task_readiness_expansion} the
goals were removed after the moving-part check.
The design holds no object size, position or orientation; sizes come
from the size model of WorldLayout and poses from its layout agent
(Sec.~\ref{supp:world:layout}).

\subsection{Layout with Physical and Access Constraints}
\label{supp:world:layout}

\subsubsection{Related Work}
\label{sec:layout_related}
Table~\ref{tab:related} compares WorldLayout with 32 indoor layout generators in three groups: one-stage direct generation (I), agentic generation (II), and constraint solving (III).
Checks, circles and crosses indicate yes, partial and no, respectively.

\emph{Input $\to$ output} gives the main condition of the generation, an intermediate form where there is one, and the output; the tags are listed below the table.
\emph{Train} gives how the model is obtained: without training (ICL), by supervised training on layouts (GT), by fine-tuning (SFT, DPO, GRPO, its zone-aware variant ZGRPO, and rejection-sampling fine-tuning, RFT), or from hand-written rules (Rules).
\emph{Source, scale} names the data and its size, and \emph{Real} marks real floor plans or scans (circle: partly real, or real data used only for evaluation).
\emph{Fixed geometry} marks what a method takes as given input: doors and windows (D/W), columns and partitions (Col.), and stairs (St.); a circle means that the method generates them itself or handles them implicitly.
Under \emph{Completion}, \emph{Partial} marks a method that takes a partially furnished scene (circle: only its own output), and \emph{Layered} a method that places floor objects before the objects they support (circle: an order by size only).
Under \emph{Hardening}, \emph{Yaw} marks a discrete yaw (check: integer degrees; circle: classes of $90^\circ$), \emph{Snap} wall snapping (check: with an explicit tolerance; circle: by wall-relative placement), and \emph{Check} a collision or bounds check at inference.
\emph{Mechanism} describes how each method handles collisions and room boundaries: a hard constraint (Hard), rejection of infeasible outputs (Check), a solver or rules (Solver), scored candidates (Best-of-$K$), a reward (RL), a training loss or guidance (Loss), or none. Bar lengths show our qualitative ordering of these mechanisms.

WorldLayout (last row) takes the room outline, its doors and windows, and the object list as input.
Its sizes come from a fine-tuned model and its poses from a language model without training; a verifier checks every layout, and a repair search moves objects to remove the violations it reports (Sec.~\ref{sec:layout_method}).

\begin{table*}[t]
\centering
\caption{Comparison of indoor layout generators.}\label{tab:related}
\vspace{2pt}
\begingroup
\fontsize{7}{8.1}\selectfont
\setlength{\tabcolsep}{1.5pt}
\renewcommand{\arraystretch}{1.1}
\arrayrulecolor{neutralD}
\resizebox{\textwidth}{!}{%
\begin{tabular}{@{}>{\columncolor{cA1!14}}>{\centering\arraybackslash}p{1.25cm} @{\hspace{4pt}} l l l L{3.1cm} L{2.6cm} L{1.6cm} @{\hspace{5pt}} L{3.3cm}
  >{\columncolor{cA1!8}}c
  >{\columncolor{cA2!8}}c >{\columncolor{cA2!8}}c >{\columncolor{cA2!8}}c
  >{\columncolor{cA3!8}}c >{\columncolor{cA3!8}}c
  >{\columncolor{cA4!8}}c >{\columncolor{cA4!8}}c >{\columncolor{cA4!8}}c
  @{\hspace{5pt}} l@{}}
\toprule
\multicolumn{1}{@{}c}{} & \multicolumn{7}{l}{} & \rtab{1}{cA1}{\faDatabase\ Data} & \rtab{3}{cA2}{\faDoorOpen\ Fixed geometry} & \rtab{2}{cA3}{\faLayerGroup\ Completion} & \rtab{3}{cA4}{\faCogs\ Hardening} & \\[1pt]
\multicolumn{1}{@{}c}{} & \# & Method & Venue & Input $\to$ output & Arch & Train & Source, scale & Real & D/W & Col. & St. & Partial & Layered & Yaw & Snap & Check & Mechanism \\
\midrule
 & \textcolor{neutralM}{1} & LayoutGPT\href{https://github.com/weixi-feng/LayoutGPT}{\ghub}~\cite{layoutgpt} & NeurIPS~'23 & \iTxt\lnk \oThree & GPT-4 & ICL & 3D-Front exemplars & \xmark & \xmark & \xmark & \xmark & \ym{cA3} & \xmark & \xmark & \xmark & \xmark & \tNone \\
 & \textcolor{neutralM}{2} & InstructScene\href{https://github.com/chenguolin/InstructScene}{\ghub}~\cite{lin2024instructscene} & ICLR~'24 & \iTxt\lnk \oRel\lnk \oThree & Graph+diff. & GT & 3D-Front & \xmark & \xmark & \xmark & \xmark & \ym{cA3} & \xmark & \xmark & \xmark & \xmark & \tNone \\
 & \textcolor{neutralM}{3} & DiffuScene\href{https://github.com/tangjiapeng/DiffuScene}{\ghub}~\cite{tang2024diffuscene} & CVPR~'24 & \iNoise\lnk \oThree & Diffusion & GT & 3D-Front & \xmark & \xmark & \xmark & \xmark & \ym{cA3} & \xmark & \xmark & \xmark & \xmark & \tLoss \\
 & \textcolor{neutralM}{4} & PhyScene\href{https://github.com/PhyScene/PhyScene}{\ghub}~\cite{yang2024physcene} & CVPR~'24 & \iRoom\lnk \oThree & Diff.+guidance & GT & 3D-Front & \xmark & \xmark & \xmark & \xmark & \xmark & \xmark & \xmark & \xmark & \hm{cA4} & \tLoss \\
 & \textcolor{neutralM}{5} & SceneTeller\href{https://github.com/sceneteller/SceneTeller}{\ghub}~\cite{sceneteller} & ECCV~'24 & \iTxt\lnk \oThree & GPT-4 & ICL & 3D-Front exemplars & \xmark & \xmark & \xmark & \xmark & \xmark & \xmark & \xmark & \xmark & \xmark & \tNone \\
 & \textcolor{neutralM}{6} & OptiScene\href{https://github.com/PolySummit/OptiScene}{\ghub}~\cite{optiscene} & NeurIPS~'25 & \iTxt\lnk \oRel\lnk \oThree & Qwen3-8B & SFT, 2$\times$DPO & 3DFront+Holodeck 17k & \xmark & \xmark & \xmark & \xmark & \hm{cA3} & \xmark & \xmark & \xmark & \xmark & \tRL \\
 & \textcolor{neutralM}{7} & DirectLayout\href{https://github.com/rxjfighting/DirectLayout}{\ghub}~\cite{directlayout} & NeurIPS~'25 & \iTxt\lnk \oTwo\lnk \oThree & Qwen2.5-32B+72B & SFT+DPO & 3D-Front, 6.5k & \xmark & \xmark & \xmark & \xmark & \xmark & \xmark & \xmark & \xmark & \xmark & \tRL \\
 & \textcolor{neutralM}{8} & ReSpace\href{https://github.com/gradientspaces/respace}{\ghub}~\cite{respace} & arXiv~'25 & \iTxt\lnk \oThree & Qwen2.5-1.5B & SFT+RFT & 3D-FRONT, 12k & \xmark & \xmark & \xmark & \xmark & \ym{cA3} & \xmark & \hm{cA4} & \xmark & \xmark & \tBoK \\
 & \textcolor{neutralM}{9} & MiDiffusion\href{https://github.com/MIT-SPARK/MiDiffusion}{\ghub}~\cite{hu2026mixed} & WACV~'26 & \iRoom\lnk \oThree & Mix. diff. & GT & 3D-Front 5.7k & \xmark & \xmark & \xmark & \xmark & \ym{cA3} & \xmark & \xmark & \xmark & \xmark & \tNone \\
 & \textcolor{neutralM}{10} & FP-RLVR\href{https://github.com/ludolara/floor-plan-rlvr}{\ghub}~\cite{fprlvr}  & ACL Find.~'26 & \iTxt\lnk \oTwo & Llama-3.3-70B & SFT+GRPO & RPLAN, 80k & \ym{cA1} & \xmark & \xmark & \xmark & \xmark & \xmark & \xmark & \xmark & \xmark & \tBoK \\
\multirow{-11}{*}{\parbox{1.2cm}{\centering\fontsize{6.3}{6.8}\selectfont\textcolor{cA1}{\fontsize{8.5}{8.5}\selectfont\textbf{I}}\\[-1pt]\textcolor{bluemain}{One-stage\\direct}}} & \textcolor{neutralM}{11} & ZoneMaestro~\cite{zonemaestro}   & ICML~'26    & \iTxt\lnk \oRel\lnk \oThree & Qwen3-8B & SFT+ZGRPO & InternScenes+3DFront & \hm{cA1} & \xmark & \xmark & \xmark & \xmark & \xmark & \xmark & \xmark & \xmark & \tRL \\
\specialrule{0.4pt}{1.5pt}{1.5pt}
 & \textcolor{neutralM}{12} & ARCHITECT\href{https://github.com/wangyian-me/architect_official_code}{\ghub}~\cite{wang2024architect} & NeurIPS~'24 & \iTxt\lnk \oThree \chip{cRel}{$\circlearrowleft$} & LLM+inpainting & ICL & none & \xmark & \xmark & \xmark & \xmark & \hm{cA3} & \ym{cA3} & \xmark & \xmark & \xmark & \tNone \\
 & \textcolor{neutralM}{13} & SceneWeaver\href{https://github.com/Scene-Weaver/SceneWeaver}{\ghub}~\cite{sceneweaver} & NeurIPS~'25 & \iTxt\lnk \oThree \chip{cRel}{$\circlearrowleft$} & LLM agent & ICL & none & \xmark & \hm{cA2} & \xmark & \xmark & \ym{cA3} & \ym{cA3} & \xmark & \hm{cA4} & \ym{cA4} & \tSolver \\
 & \textcolor{neutralM}{14} & WorldGrow\href{https://github.com/world-grow/WorldGrow}{\ghub}~\cite{li2026worldgrow} & AAAI~'26 & \iTxt\lnk \oThree \chip{cRel}{$\circlearrowleft$} & Diff.+inpainting & GT & 3D-Front & \xmark & \xmark & \xmark & \xmark & \xmark & \xmark & \xmark & \xmark & \xmark & \tNone \\
 & \textcolor{neutralM}{15} & SAGE\href{https://github.com/NVlabs/sage}{\ghub}~\cite{sage}     & CVPR~'26    & \iTxt\lnk \oThree \chip{cRel}{$\circlearrowleft$} & LLM 120B & ICL & none & \xmark & \hm{cA2} & \xmark & \xmark & \ym{cA3} & \ym{cA3} & \xmark & \xmark & \ym{cA4} & \tCheck \\
 & \textcolor{neutralM}{16} & Edit-As-Act\href{https://github.com/SeongRaeNoh/Edit-As-Act}{\ghub}~\cite{Noh_2026_CVPR} & CVPR~'26 & \iScene\iTxt\lnk \oOps\lnk \oThree \chip{cRel}{$\circlearrowleft$} & LLM+PDDL & ICL & none & \xmark & \xmark & \xmark & \xmark & \ym{cA3} & \xmark & \xmark & \xmark & \ym{cA4} & \tSolver \\
 & \textcolor{neutralM}{17} & MANSION\href{https://github.com/AgibotGeneral/MANSION}{\ghub}~\cite{Che_2026_CVPR} & CVPR~'26 & \iTxt\lnk \oThree \chip{cRel}{$\circlearrowleft$} & MLLM agents & ICL & MansionWorld 1k+ & \xmark & \hm{cA2} & \xmark & \ym{cA2} & \xmark & \ym{cA3} & \xmark & \hm{cA4} & \hm{cA4} & \tSolver \\
 & \textcolor{neutralM}{18} & SceneSmith\href{https://github.com/nepfaff/scenesmith}{\ghub}~\cite{pfaff2026scenesmith} & ICML~'26  & \iTxt\lnk \oThree \chip{cRel}{$\circlearrowleft$} & GPT-5.2 & ICL & none & \xmark & \hm{cA2} & \xmark & \xmark & \hm{cA3} & \ym{cA3} & \xmark & \hm{cA4} & \ym{cA4} & \tSolver \\
 & \textcolor{neutralM}{19} & SceneReVis\href{https://github.com/Runder-sun/SceneReVis}{\ghub}~\cite{scenerevis} & ACM~MM~'26   & \iScene\iTxt\lnk \oOps\lnk \oThree \chip{cRel}{$\circlearrowleft$} & Qwen2.5-VL-7B & SFT+GRPO & SceneChain-12k & \xmark & \xmark & \xmark & \xmark & \ym{cA3} & \hm{cA3} & \xmark & \xmark & \hm{cA4} & \tRL \\
 & \textcolor{neutralM}{20} & EmbodiedGen V2\href{https://github.com/HorizonRobotics/EmbodiedGen}{\ghub}~\cite{wang2026embodiedgen} & arXiv~'26 & \iTxt\lnk \oThree\oMesh \chip{cRel}{$\circlearrowleft$} & LLM agents & ICL & none & \xmark & \hm{cA2} & \xmark & \xmark & \ym{cA3} & \ym{cA3} & \xmark & \hm{cA4} & \hm{cA4} & \tCheck \\
\multirow{-10}{*}{\parbox{1.2cm}{\centering\fontsize{6.3}{6.8}\selectfont\textcolor{cA1}{\fontsize{8.5}{8.5}\selectfont\textbf{II}}\\[-1pt]\textcolor{bluemain}{Agentic}}} & \textcolor{neutralM}{21} & EnvDreamer~\cite{swain2026envdreamerlargescalemultimodaltoenvironmentgeneration} & arXiv~'26 & \iTxt\lnk \oThree\oMesh \chip{cRel}{$\circlearrowleft$} & LLM+UE5 & ICL & EnvDreamer-20k & \xmark & \hm{cA2} & \xmark & \xmark & \xmark & \xmark & \xmark & \xmark & \hm{cA4} & \tCheck \\
\specialrule{0.4pt}{1.5pt}{1.5pt}
 & \textcolor{neutralM}{22} & ProcTHOR\href{https://github.com/allenai/procthor}{\ghub}~\cite{deitke2022} & NeurIPS~'22 & \iRules\lnk \oThree\oMesh & Proc.+solver & Rules & none & \xmark & \ym{cA2} & \xmark & \xmark & \xmark & \ym{cA3} & \hm{cA4} & \ym{cA4} & \ym{cA4} & \tSolver \\
 & \textcolor{neutralM}{23} & Infinigen Indoors\href{https://github.com/princeton-vl/infinigen}{\ghub}~\cite{infinigenindoors} & CVPR~'24 & \iRules\lnk \oThree\oMesh & Proc.+solver & Rules & none & \xmark & \hm{cA2} & \xmark & \hm{cA2} & \xmark & \ym{cA3} & \xmark & \hm{cA4} & \ym{cA4} & \tSolver \\
 & \textcolor{neutralM}{24} & Holodeck\href{https://github.com/allenai/Holodeck}{\ghub}~\cite{holodeck} & CVPR~'24    & \iTxt\lnk \oRel\lnk \oThree & LLM+solver & ICL & none & \xmark & \hm{cA2} & \xmark & \xmark & \xmark & \hm{cA3} & \hm{cA4} & \hm{cA4} & \ym{cA4} & \tSolver \\
 & \textcolor{neutralM}{25} & FlairGPT\href{https://github.com/gabriellelittle1/FlairGPT}{\ghub}~\cite{flairgpt} & EG~'25      & \iTxt\lnk \oRel\lnk \oThree & LLM+SLSQP & ICL & none & \xmark & \ym{cA2} & \xmark & \xmark & \xmark & \ym{cA3} & \xmark & \hm{cA4} & \ym{cA4} & \tSolver \\
 & \textcolor{neutralM}{26} & LayoutVLM\href{https://github.com/sunfanyunn/LayoutVLM}{\ghub}~\cite{layoutvlm} & CVPR~'25   & \iTxt\iImg\lnk \oRel\lnk \oThree & VLM+optim. & ICL / SFT & 3D-Front, 9k & \xmark & \xmark & \xmark & \xmark & \hm{cA3} & \xmark & \xmark & \hm{cA4} & \ym{cA4} & \tSolver \\
 & \textcolor{neutralM}{27} & TreeSearchGen\href{https://github.com/dw-dengwei/TreeSearchGen}{\ghub}~\cite{gltreesearch} & CVPR~'25 & \iTxt\lnk \oThree & VLM (GPT-4o) & ICL & none & \xmark & \xmark & \xmark & \xmark & \xmark & \ym{cA3} & \hm{cA4} & \hm{cA4} & \hm{cA4} & \tSolver \\
 & \textcolor{neutralM}{28} & Co-Layout\href{https://github.com/xccElephant/co-layout}{\ghub}~\cite{colayout} & AAAI~'26   & \iTxt\lnk \oRel\lnk \oTwo & LLM+IP & ICL & none & \xmark & \hm{cA2} & \xmark & \xmark & \xmark & \xmark & \hm{cA4} & \xmark & \ym{cA4} & \tHard \\
 & \textcolor{neutralM}{29} & HSM\href{https://github.com/3dlg-hcvc/hsm}{\ghub}~\cite{pun2026hsm} & 3DV~'26 & \iTxt\lnk \oThree \chip{cRel}{$\circlearrowleft$} & VLM+motifs & ICL & none & \xmark & \hm{cA2} & \xmark & \xmark & \xmark & \ym{cA3} & \xmark & \hm{cA4} & \ym{cA4} & \tSolver \\
 & \textcolor{neutralM}{30} & PolyLayout\href{https://github.com/ghanning/PolyLayout}{\ghub}~\cite{polylayout}    & arXiv~'26   & \iRoom\iDoor\iTxt\lnk \oRel\lnk \oThree & Gemini 2.5 Flash & ICL & none & \hm{cA1} & \ym{cA2} & \xmark & \xmark & \ym{cA3} & \xmark & \xmark & \hm{cA4} & \ym{cA4} & \tSolver \\
 & \textcolor{neutralM}{31} & RoboLayout\href{https://github.com/alishams21/robolayout}{\ghub}~\cite{robolayout} & arXiv~'26 & \iTxt\iImg\lnk \oRel\lnk \oThree & VLM+optim. & ICL / SFT & 3D-Front, 9k & \xmark & \xmark & \xmark & \xmark & \hm{cA3} & \xmark & \xmark & \hm{cA4} & \ym{cA4} & \tSolver \\
\multirow{-11}{*}{\parbox{1.2cm}{\centering\fontsize{6.3}{6.8}\selectfont\textcolor{cA1}{\fontsize{8.5}{8.5}\selectfont\textbf{III}}\\[-1pt]\textcolor{bluemain}{Constraint\\solving}}} & \textcolor{neutralM}{32} & SceneCode\href{https://github.com/wangpuyi/SceneCode}{\ghub}~\cite{wang2026scenecode} & arXiv~'26 & \iTxt\lnk \oProg\lnk \oThree\oMesh \chip{cRel}{$\circlearrowleft$} & LLM+code & ICL & none & \xmark & \hm{cA2} & \xmark & \xmark & \ym{cA3} & \xmark & \xmark & \xmark & \ym{cA4} & \tCheck \\
\specialrule{0.6pt}{2pt}{0pt}
\rowcolor{oursbg}\cellcolor{oursbg} & 33 & \textbf{Ours (WorldLayout)} & -- & \iTxt\iRoom\iDoor\lnk \oRel\lnk \oThree \chip{cRel}{$\circlearrowleft$} & Qwen3+GPT-6.1+solver & SFT+ICL & \textbf{SAGE-10k+IL3D, 29.6k} & \xmark & \ym{cA2} & \xmark & \xmark & \ym{cA3} & \ym{cA3} & \xmark & \hm{cA4} & \ym{cA4} & \tSolver \\
\bottomrule
\end{tabular}
}
\endgroup
\vspace{4pt}
\par
\begingroup\fontsize{6.2}{7.4}\selectfont\color{neutralD}\setlength{\parskip}{0pt}
{\fontsize{5.5}{5.5}\selectfont\textcolor{neutralM}{\faGithub}}\ public code.\quad
\ym{neutralD}\ yes; \hm{neutralD}\ partial; \xmark\ no.\quad
T text, I image, R room boundary, D doors and windows, S partial scene, N noise; rel, relations or scene graph; prog, program; ops, edit operations; $\circlearrowleft$ iterative.\quad
The columns are defined in Sec.~\ref{sec:layout_related}.
\endgroup
\end{table*}


\subsubsection{Method Details}
\label{sec:layout_method}

\begin{figure*}[t]
\centering
\includegraphics[width=\textwidth]{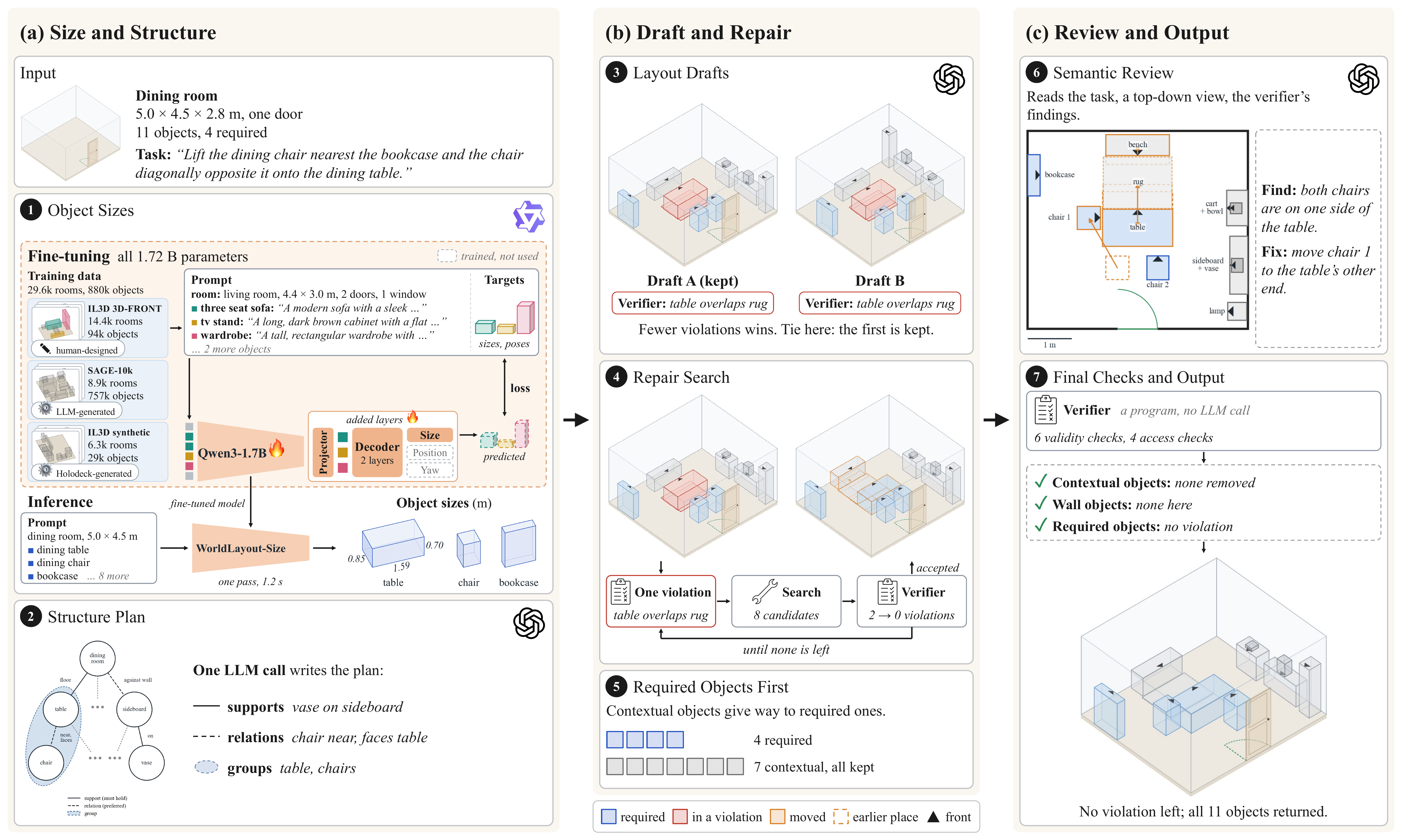}
\caption{\textbf{WorldLayout in a dining room.} (a) A fine-tuned model, WorldLayout-Size, predicts the object sizes, and one LLM call writes the structure plan. (b) The verifier checks two drafts; the kept draft is repaired by search, and contextual objects give way to required ones. (c) A semantic review and the final checks give the output layout. In the layouts, blue marks required objects, red marks objects in a violation, and orange marks moved objects; dashed outlines mark earlier positions and triangles mark object fronts. In step~1, dashed boxes mark outputs that are trained but not used.}
\label{fig:layout_pipeline}
\end{figure*}

WorldLayout first fixes each object's dimensions (step~1). A language model
assigns supports and functional groups (step~2) and drafts two layouts
(step~3). The verifier ranks the drafts, and a bounded search moves
objects in the selected layout to repair violations (step~4). If
contextual objects still obstruct a required object, the system tries
to relocate or remove them (step~5). A language model reviews the layout
for functional problems (step~6). Final checks return the layout or a
fallback (step~7). Only steps 2, 3, and 6 use a language model.
Figure~\ref{fig:layout_pipeline} follows a dining room with 11 objects, four required by the task, from the run in Table~\ref{tab:panel_a_sceneeval}.
Both drafts place the table over the rug. WorldLayout keeps draft A and moves the table, rug, and bench to clear the overlap. All contextual objects stay. The final review moves one of the two required chairs to the opposite end of the table.
Each box has an oriented rectangular footprint, a bottom height and a height. Its local $+x$ axis points forward.

\paragraph{Task inputs.}
WorldLayout receives the room outline and height, the dimensions and positions of fixed doorways,
an object list, and the task. Each object has a category and description. The input specifies its
role and, where applicable, its support and whether the task requires an articulated part.
Object sizes and poses are determined later. The task record includes the instruction,
initial relations, reasons for including each object, and the robot. Required objects stay in
the layout. The solver may remove contextual furnishings that it cannot place without a violation.

\begin{figure}[t]
\centering
\includegraphics[width=\columnwidth]{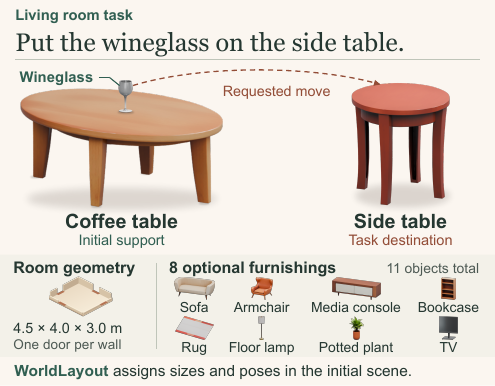}
\caption{\textbf{WorldLayout input for the wineglass task.} The meshes illustrate the object list. The wineglass and side table are required. The coffee table is a dependency because it supports the glass initially.}
\label{fig:layout_input}
\end{figure}

Figure~\ref{fig:layout_input} follows the wineglass example in Table~\ref{tab:supp:world:designpipeline}.
The instruction names the side table as the destination. The Room Designer supplies the coffee table
as the starting support, and the accepted layout places the glass there. The room-only request
keeps this initial relation and omits the execution goal.
Access checks use the same base dimensions~across~scenes.

\paragraph{Size model.}
The size model is Qwen3-1.7B~\cite{qwen2025qwen3}, fine-tuned and used as an encoder and not as a text generator.
The size model receives the room geometry, constraints and object list in a single JSON prompt. The task text is omitted.
The model reads the prompt in one pass.
Its last hidden states (2{,}048 dimensions) are projected to 128 dimensions.
Each object gets one query: the mean of the projected states over the tokens of its own JSON entry, plus a learned index embedding.
A two-layer Transformer decoder (width 128, four heads) lets the queries attend to each other and to all tokens.
A linear head maps each query to the logarithm of the object's size (depth, width, and height in metres).
Two more heads predict a position (a $16\times16$ grid over the room with an offset per cell, and a height) and a yaw (12 bins of $30^\circ$ with an offset per bin).
At test time the model makes one forward pass per scene without sampling (1.2\,s per scene on average in our evaluation), and its sizes are used as predicted, without clamping or rounding.

The size model trains on rooms with detailed object descriptions from SAGE-10k~\cite{sage} and two IL3D~\cite{zhou2025il3d} subsets: designer-authored 3D-FRONT rooms~\cite{Fu_2021_ICCV} and synthetic rooms.
In these sources, 88--95\% of the objects have a description of at least 12 words that names at least two of
colour, material, shape, and style. The data has 29{,}611 training rooms with 879{,}857 objects, 1{,}785
validation rooms, and 1{,}667 test rooms; all rooms of one house are in the same split. All training rooms are
used. A training room gives the room outline, its openings and fixed objects, and the object list with
descriptions; the targets are the objects' sizes and poses.
All 1.72\,B parameters are trained for five epochs (4{,}627 updates of 32 rooms).
We use AdamW~\cite{loshchilov2019decoupled} with a learning rate of $10^{-5}$ for the backbone and $10^{-4}$ for the added layers, a linear warm-up over the first 3\% of the updates, a cosine decay to 10\% of the peak rate, gradient clipping at 1.0, and bfloat16 mixed precision.
The loss is
\begin{equation}
\mathcal{L} = \mathcal{L}_\text{pos} + 0.6\,\mathcal{L}_\text{size} + 0.08\,\mathcal{L}_\text{bin} + 2.0\,\mathcal{L}_\text{off},
\end{equation}
where $\mathcal{L}_\text{size}$ is the mean absolute error of the log size over the three axes, $\mathcal{L}_\text{pos}$ sums the cross-entropy over the grid cells (weight 0.5), the L1 error of the offset in the true cell (0.4), and the L1 error of the height (1/3), $\mathcal{L}_\text{bin}$ is the cross-entropy over the yaw bins, and $\mathcal{L}_\text{off}$ is the smooth-L1 error of the yaw offset.
Objects with identical entries in the prompt are exchangeable; inside such a group, predictions are assigned to targets by Hungarian matching~\cite{kuhn1955hungarian}.
For augmentation, the room is turned by a random multiple of $90^\circ$, the object order is shuffled, constraints are dropped with probability 0.3 and given supports with probability 0.2, and with probability 0.5 the room is reduced to its type and bounding rectangle.
Training takes 5.3 hours on four GH200 GPUs.
We use the checkpoint after 4{,}500 updates, selected on the validation rooms.

We take the sizes from this model, and not from the language model that writes the drafts, for two reasons.
First, WorldLayout fixes all sizes before any object is placed, and no later step changes them: a violation can then only be removed by moving an object or leaving out a contextual one, never by changing a size.
Second, on held-out rooms the sizes of this model agree better with the sizes of the datasets' assets than the sizes written by the language model (Table~\ref{tab:size_accuracy}).
The position and yaw heads are trained with the size head, but WorldLayout uses only the sizes.
The size of an object follows from its description and its room, which the model reads.
Its pose has to follow the task, leave the robot access to the required objects, and pass the verifier, and the model sees no task text, no robot description, and no image. The poses therefore come from the drafts and the repair.
When an object comes with a retrieved asset, the measured size of the asset replaces the predicted one; no object in our evaluation scenes has one, so all 2{,}282 sizes come from the size model.
Of the 150 evaluation tasks, 37 were written from scenes in the training split of the size model; the layout stage does not see these scenes, and none of the 2{,}282 object descriptions occurs in the training data.
The evaluation scenes have no reference sizes, so size accuracy is measured on held-out rooms of the test split (Sec.~\ref{sec:layout_ablation}).

\paragraph{Structure plan.}
The plan call receives the room and the object list, without sizes and without the task.
For each object it returns a support (the floor, a wall, or another object) and a placement priority. For the scene it returns functional groups (an anchor and its members) and relations of three types: near (with a distance), faces, and against a wall.
The validator accepts the floor, a wall, or an existing object as a support
and rejects cycles. An invalid entry retains its rule-based default
(for example, a wall for a wall-hung object).
A support given in the request is locked: neither the plan, the repair, nor the review can change it.
Relations from the plan are soft and enter only $E_\text{soft}$; constraints given in the request stay hard.
In the 149 scenes with a saved plan, the call changes the support that the rules of the program alone give for 70 of the 1{,}572 objects whose support the request leaves open. Of these 70, 40 are hung on a wall and 28 are put on another object (for example, a book on a table) where the rules leave them on the floor, and two are put on the floor.
Its other output, 2{,}002 soft relations and 584 groups, enters only the draft prompt and $E_\text{soft}$.

\begin{figure*}[t]
\centering
\includegraphics[width=\textwidth]{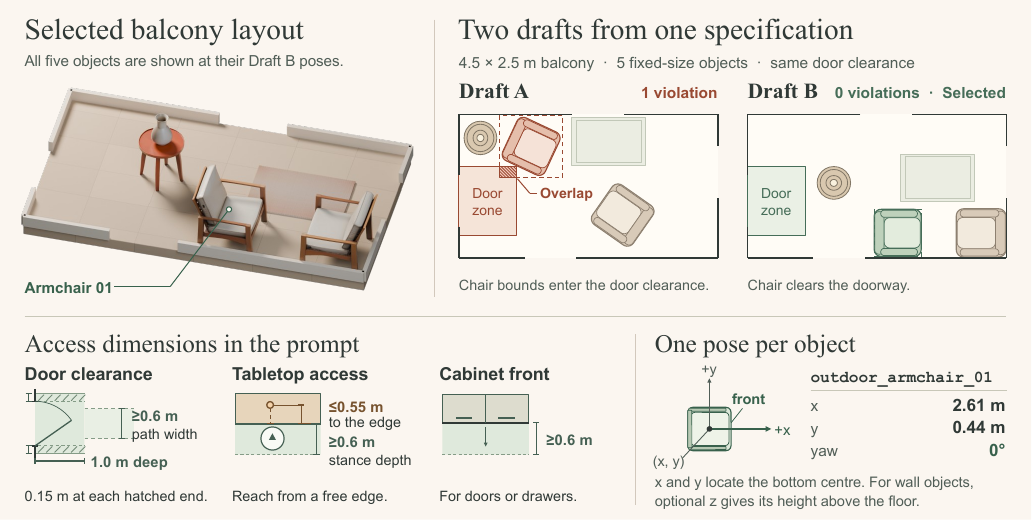}
\caption{\textbf{WorldLayout draft prompt.} Hatching marks where the armchair's axis-aligned bounding box (dashed) enters the left door zone in A. The verifier selects B with zero violations. The rendering and armchair pose both come from B. Yaw is counterclockwise from world $+x$.}
\label{fig:layout_prompts_draft}
\end{figure*}

\paragraph{Layout drafts.}
The draft call (Fig.~\ref{fig:layout_prompts_draft}) receives the room, the constraints, the plan's groups and relations, and each object with its fixed size, support, and priority.
Three marks are added by the program: \emph{required} for the objects the task needs, a front clearance (0.6\,m) for furniture with doors or drawers, and a mark for furniture that carries an object.
The task text itself is not given; it is used only in the review.
The prompt asks the model to keep objects inside the room, avoid collisions, and leave 1.0\,m clear in front of every door. The robot needs a path at least 0.6\,m wide from the doors to the marked furniture and space to stand beside it. If space is limited, required objects take priority.
The answer gives $x$, $y$, and a yaw in degrees for every object.
The program completes the pose: a floor object stands on the floor, an object on a support takes the height of the support's top, and a wall object is moved to the nearest wall with its back 2\,mm from it, facing the room, by default 1.15\,m above the floor.
An object that the answer leaves out is put on its support or at the room centre; this did not occur in a chosen draft of our run.

We sample $K{=}2$ drafts in parallel from one prompt and rank by violation count (validity and access), then total magnitude, then draft index.
In our run 230 of the 300 drafts (76.7\%) have no violation, and the chosen draft has none in 132 of the 150 scenes (88.0\%).

\paragraph{Verifier.}
Table~\ref{tab:layout_checks} lists the checks.
The access checks use a robot base 0.6\,m wide (the 0.5\,m base plus a margin of 0.1\,m); these are our settings for the robot.
The free floor is built as in the reachability check of SceneSmith~\cite{pfaff2026scenesmith}, which leaves the width to be set for the robot. It is the floor polygon shrunk by half this width, minus every object standing on the floor grown by half this width. Floor coverings lower than 0.1\,m are ignored.
The \emph{main region} of the free floor is the connected part that touches the most doors.
The checks are computed exactly on polygons, without a grid.
A collision between two movable objects is counted once for each of them.

The distances of the checks have the following sources.
The door keep-out (1.0\,m deep, 0.15\,m to each side) is the admission rule of the pipeline that receives the layout: a room that breaks it is refused.
The depth of the door access strip (0.8\,m) is the door clearance in the released configuration of SceneSmith, and the window check follows its window clearance.
The 0.85\,m of the object-on-furniture check is the largest distance between the robot base and an object that EmbodiedGen V2~\cite{wang2026embodiedgen} allows when it places objects.
The tolerances of the boundary and support checks (1\,mm and 4\,mm) are set inside those of the SceneEval metric code~\cite{sceneeval}, which counts an object as out of bounds when fewer than 99\% of its sampled points are inside the room and accepts a support within 1\,cm.
The other two are our settings: the standing strip of 0.7\,m is half the base width plus a working distance of 0.4\,m, and the opening clearance takes the depth of the furniture, limited to 0.35--0.6\,m, as the sweep of a door leaf or a drawer.
\begin{table*}[t]
\centering
\footnotesize
\setlength{\tabcolsep}{4pt}
\begin{tabular}{@{}lp{0.62\textwidth}l@{}}
\toprule
Check & Violated when & Magnitude \\
\midrule
\multicolumn{3}{@{}l}{\emph{Validity} ($V_\ell$)} \\
Collision & the footprints of two objects, or of an object and a fixed object, overlap by more than $10^{-5}$\,m$^2$ and their height ranges overlap & $\sqrt{A_{\mathrm{ov}}}$ \\
Boundary & the footprint is not inside the floor polygon shrunk by 1\,mm, or the box is below the floor or above the ceiling & $\sqrt{A_{\mathrm{out}}}$ or distance \\
Support & floor object: its bottom is not on the floor. Wall object: its back is more than 4\,mm from a wall, or it does not face the room. Object on a support: its bottom is not at the height of the support's top, or its footprint centre is not on that top & distance \\
Door keep-out & any object, at any height, enters the area 1.0\,m deep in front of a door (the door width plus 0.15\,m on each side); floor coverings lower than 0.1\,m are exempt & $\sqrt{A_{\mathrm{ov}}}$ \\
Window & a wall object hangs over a window & $\sqrt{A_{\mathrm{ov}}}$ \\
Input constraint & a hard constraint of the request does not hold & its gap \\
\midrule
\multicolumn{3}{@{}l}{\emph{Access} ($V_a$)} \\
Door access & the main region does not touch the area 0.8\,m deep in front of a door & distance \\
Standing place & for furniture that is required, carries an object, or opens: no usable side has a strip 0.7\,m deep that touches the main region (usable sides: the front; for beds also the long sides; for tables and seats all four) & distance \\
Opening clearance & for furniture with doors or drawers, or that the task opens: more than 10\% of the rectangle in front of it (as wide as its front, as deep as the furniture, between 0.35 and 0.6\,m) is outside the room or covered & covered share \\
Object on furniture & the centre of an object standing on furniture is more than 0.85\,m from the main region & excess distance \\
\bottomrule
\end{tabular}
\caption{\textbf{WorldLayout validity and access checks.}}
\label{tab:layout_checks}
\par\smallskip{\footnotesize\raggedright $A_{\mathrm{ov}}$ measures overlap area and $A_{\mathrm{out}}$ measures footprint area outside the floor polygon. Taking the square root converts area to length. Magnitudes are in metres except for opening clearance, which uses the covered fraction. In Eq.~(\ref{eq:method:layout}), $M_\ell$ and $M_a$ sum validity and access magnitudes, respectively, to rank layouts within a scene.\par}
\end{table*}

The soft cost $E_\text{soft}$ is a weighted sum of terms, each limited to 1: unmet soft constraints of the request (weight 1.0) and unmet relations of the plan (0.5); 62 category-pair patterns (near, faces) and a back-to-wall rule for 47 categories (0.5); furniture with less than half of a 0.5\,m strip free on its best side (0.5); furniture that a person cannot walk up to from a door (1.0); and floor objects in front of a window (0.25).
We estimated the pair patterns from 18{,}857 rooms and the wall rules from 18{,}880 rooms sampled from the same 125{,}787-room training split, which includes designed, generated, and scanned layouts.
As in SceneSmith, a \emph{faces} relation holds when a ray from the front of the object intersects the target's box.
The weights are our settings.
\paragraph{Repair search and acceptance rule.}
The repair takes the violations one at a time: validity before access, violations of required objects first, larger ones first.
For one violation it searches new poses for a set of objects that grows in three levels until a candidate is accepted: (1) the objects of the violation, the objects that collide with them, their relation partners, and everything standing on these; (2) their functional groups; (3) all objects.
Other objects stay fixed.

The search is a beam search (width 24) that places the objects of the set one after another, each at one of a fixed list of candidate poses: 33 poses around the current one (shifts of 0.05, 0.1, 0.2, and 0.4\,m in eight directions), positions along the walls facing the room, a floor grid of about 0.6\,m with four yaws, and positions beside relation partners. An object on a support is tried at its old place and at spots in the middle and along the edges of the support's top.
A pose is kept only if it adds no validity violation against the objects already placed; the kept poses are ranked by the distance moved and the soft relations met, and the best 48 are expanded.
Sizes are never changed.
Up to eight complete layouts are scored by the verifier, and the one with the smallest $k(L)$ is the candidate $L'$.

$L'$ replaces the current layout $L$ by the following rule.
The tiers $t\in\{\ell,a\}$ are examined in this order, and the first tier that decides is final:
(i) if $L$ has no violation in tier $t$ and $L'$ has one, reject;
(ii) if $L'$ has a violation in tier $t$ that $L$ does not have, with a magnitude above 0.05, reject;
(iii) if $|V_t(L')|<|V_t(L)|$, accept; if it is larger, reject;
(iv) if $M_t(L')<M_t(L)-0.005$, accept; if $M_t(L')>M_t(L)+0.005$, reject.
If neither tier decides, $L'$ is accepted only if $E_\text{soft}(L')<E_\text{soft}(L)-0.01$.
These margins follow $k(L)$, with guard (ii) against new violations.

If no candidate resolves a door access, standing place, or opening clearance violation, the repair searches the whole room for new positions, first for up to three obstructing objects, then for the furniture itself.
The repair makes at most three passes over the violations, at most $24+3n$ attempts per pass for $n$ violations at the start, and stops after 180\,s of search time.
It uses no random numbers.
The beam width, the list of candidate poses, the margins of the rule, and these limits are our settings.
In our run it takes 0.6\,s per scene on average (28.4\,s at most).

After the repair, contextual objects give way to required ones (step 5).
A required object is never removed.
If a required object is still in a violation, the contextual objects in that violation give way. The one involved in the most such violations is searched for a new place anywhere in the room, and if no place without a violation is found it is removed together with the objects on it.
If violations of required objects remain, the required objects are laid out alone and the contextual objects return one at a time, each only where it adds no violation to a required object. The result is kept only if the required objects then have fewer violations (tried in three scenes of our run, kept in none).
After the review, the final checks (step 7) place a contextual object that is still in a validity violation again on its support or remove it, and slide a wall object whose front is covered along the wall to a free place (11 objects in 11 scenes).
If a required object is still in a violation at the end, the required objects are laid out alone by the program, and this layout is returned if it has fewer such violations (the \emph{fallback}).

\paragraph{Semantic review.}
One review round has three kinds of calls.
(a) The \emph{review} call receives the layout as JSON, a top-down drawing of the boxes with their ids and front directions, the task context (the task, the role of each object with its reason, the initial relations, and the goals), and the soft findings of the verifier.
It returns at most six problems, each with the objects concerned, a severity, and a suggested correction; it is told not to report collisions, boundaries, or blocked doors, which the program handles.
(b) For each severe problem (at most three per round), a \emph{correction} call proposes new poses for the objects of the problem and the objects standing on them.
It cannot change a size, add or remove an object, or change a locked support.
Repair is restricted to the affected objects. In each tier, the revised
layout can have at most as many violations as the current layout, with
at most the same total magnitude. Each of the two access terms of
$E_\text{soft}$ may rise by at most 0.01, and $E_\text{soft}$ itself
by at most 0.1.
(c) A \emph{comparison} call receives the two layouts as JSON and as two drawings, labelled only as layout 1 and layout 2.
It is told neither the problem nor which layout is the original, and answers which one is better or that they are the same.
The original appears first. If the reviewer prefers the correction, we swap the order. We accept the correction only if it wins both comparisons.

Each call is a separate request without shared history.
A second round is run if the first round attempted a correction (102 of the 150 scenes); its calls also receive the reasons for the rejections.
No later round is run.
The limits of a round and the two values of the gate are our settings.

\paragraph{Prompts.}
Figure~\ref{fig:layout_prompts_draft} illustrates the draft prompt and selection rule on a balcony from Panel~B.
The model receives the room, objects, constraints, and structure plan in one JSON message, without an example layout.
The 0.55\,m reach limit comes from subtracting half the 0.6\,m navigation width from the 0.85\,m used by the object-on-furniture check.

\paragraph{Run statistics.}
All language-model calls use GPT-6.1-sol~\cite{openai2026gpt61sol} at its highest reasoning setting, with no temperature or seed set.
The values of the verifier, the repair search, the acceptance rule, and the review gate were fixed before WorldLayout was first run on the 150 evaluation scenes, and none of them was changed afterwards.
Between that first run and the reported run, only the review was changed: it now receives the task context and the drawings and may run a second round.
A scene takes 7.41 calls on average (4 to 17): one plan, two drafts, 1.68 reviews, 1.52 corrections, and 1.21 comparisons.
Nearly all of the 479.5\,s per scene is spent waiting for the language model: 24.7\,s for the plan, 292.9\,s for the two drafts, and 151.3\,s for the review. The size model takes 1.2\,s and the repair 0.6\,s.

The number of scenes without any violation is 132 for the chosen draft, 143 after the repair, 145 after contextual objects give way, and 146 at the end (97.3\%).
Of the other four scenes, three keep one violation of the smallest magnitude the verifier reports (a chair whose centre is at the edge of its platform, two balls that touch, and a wall mirror at the room boundary), and one is the fallback scene, where two jars on a shelf still fail the support check and touch each other.
WorldLayout places 2{,}271 of the 2{,}282 objects (99.52\%) and all 863 required ones.
The 11 removed objects are contextual: two gave way to required objects, one did not fit on its support, and eight belong to the fallback scene.

The review raised 406 problems in 134 scenes, 232 of them severe.
It attempted 228 corrections in 102 scenes, and 80 were accepted, in 62 scenes.
Of the others, 98 were rejected by the program gate, 22 were not preferred in the comparison, and 28 gave no usable proposal.
Every correction that was preferred in the first order was also preferred in the swapped order.

\begin{figure*}[p]
\begin{minipage}[c][0.995\textheight][t]{\textwidth}
\centering
\includegraphics[width=\textwidth]{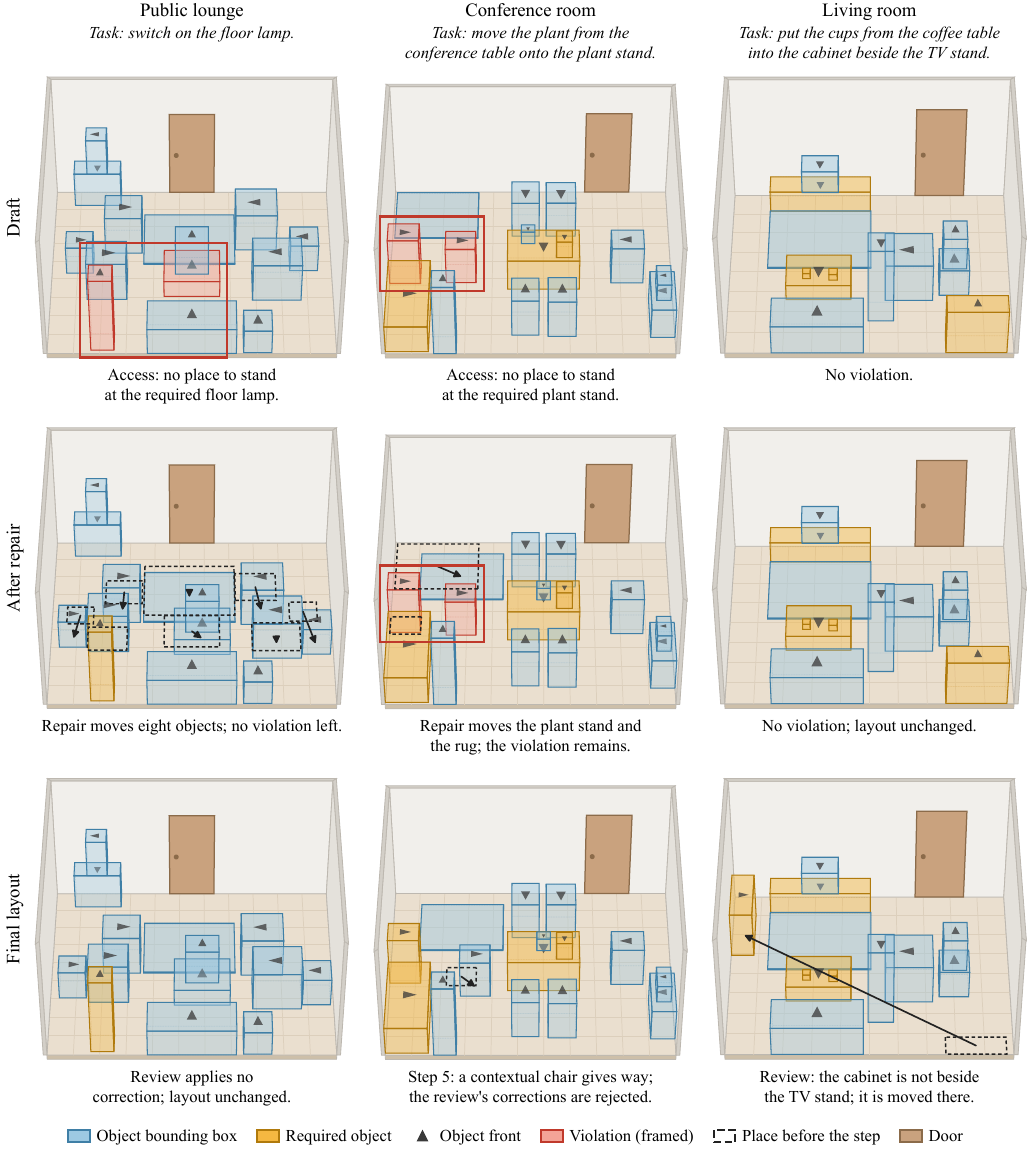}
\caption{\textbf{WorldLayout drafts, repairs, and final layouts} (rows) for three tasks (columns). Yellow marks required objects, red marks objects in a flagged violation, and triangles indicate fronts. Dashed outlines and arrows trace object movement.}
\label{fig:layout_stages}
\end{minipage}
\end{figure*}

Figure~\ref{fig:layout_stages} shows the saved layouts of three scenes of this run.
In the first scene, the draft leaves the robot no place to stand at a required object, and the repair removes the violation.
In the second, the repair does not remove such a violation. A contextual chair then gives way to the required object, and the two corrections proposed by the review are rejected, one for a new access violation and one for a higher access term of $E_\text{soft}$.
In the third, the draft passes, and the review moves a cabinet to its specified place.

\subsubsection{Ablation}
\label{sec:layout_ablation}

\paragraph{Steps.}
WorldLayout saves the layout after each step, and each step starts from the layout of the step before it. The
layout after a step is therefore what the method returns when all later steps are switched off.
Table~\ref{tab:layout_stages} scores these saved layouts of the run reported in
Table~\ref{tab:panel_a_sceneeval}, with the code and the rule of that table; no new run is made.

\begin{table*}[t]
\centering
\small
\setlength{\tabcolsep}{4pt}
\begin{tabular}{@{}lrrrrrrrrr@{}}
\toprule
Layout after & Changed & Placed$\uparrow$ & COL$_{ob}\downarrow$ & COL$_{sc}\downarrow$ & SUP$\uparrow$ & NAV$\uparrow$ & ACC$\uparrow$ & OOB$\downarrow$ & Valid$\uparrow$ \\
\midrule
One draft & -- & 100.00 & 0.53 & 3.33 & 99.42 & 98.63 & 94.33 & 0.40 & 78.0 \\
Better of two drafts (step 3) & 19 & 100.00 & 0.23 & 1.33 & 99.38 & 98.62 & 94.51 & 0.24 & 88.0 \\
+ Repair (step 4) & 12 & 100.00 & 0.00 & 0.00 & 99.38 & 98.62 & 94.42 & 0.00 & 95.3 \\
+ Required first (step 5) & 3 & 99.91 & 0.23 & 1.33 & 99.47 & 98.62 & 94.46 & 0.00 & 96.7 \\
+ Review and final checks (full) & 65 & 99.52 & 0.53 & 2.00 & 99.49 & 98.67 & 94.43 & 0.00 & 97.3 \\
\bottomrule
\end{tabular}
\caption{\textbf{WorldLayout scores after each successive stage on 150 scenes.}}
\label{tab:layout_stages}
\par\smallskip{\footnotesize\raggedright The final row reproduces Table~\ref{tab:panel_a_sceneeval} under the same scoring rule. Changed counts scenes whose layout changed at that stage. Valid is the percentage of scenes with no verifier violation. One fallback scene lacks intermediate layouts. Rows three and four use the last layout from its full attempt.\par}
\end{table*}

A single draft, with no second draft and no repair, has object and scene collision rates of 0.53\% and 3.33\% and
places every object. The lowest rates among the baselines of Table~\ref{tab:panel_a_sceneeval} are 3.82\% and
18.67\% (the baselines run at a lower reasoning setting, Sec.~\ref{sec:layout_eval}). Most of the difference to the
baselines is thus present before any step uses the verifier. Choosing the better of two drafts and the repair then
remove the remaining collisions and out-of-bounds objects (0.00 in the three columns) and raise the share of
scenes without a violation from 78.0\% to 95.3\%. The verifier also checks collisions and boundaries, so this
decrease is expected. SUP, NAV, and ACC stay within 0.2 points over all rows.

After repair, the reported collision rates rise mainly because of how unplaced objects are scored. In
step 5, one contextual object gives way in each of two scenes; in the final layouts, 11 objects in four scenes are
left out. An object that is left out is scored at the room centre (Sec.~\ref{sec:layout_eval}), where it
overlaps other objects. The final rates of 0.53\% and 2.00\% come from three scenes: the two scenes above and the
fallback scene, which leaves out eight objects and has one collision between placed objects. On placed objects
only, the rates of the last three rows are 0.00 / 0.00, 0.00 / 0.00, and 0.10 / 0.67.

We do not ablate the fourth row (step 5) further. It is not a check but a priority that follows from the purpose of
the layout: the room is laid out for a task, so an object that the task uses comes before a contextual one. The
scores of Table~\ref{tab:panel_a_sceneeval} do not know which objects a task uses, so they cannot show this priority; they
count a contextual object that gives way like any other object. The second scene of Fig.~\ref{fig:layout_stages}
is an example. The task moves a plant onto a plant stand, and after the repair a contextual chair still leaves the
robot no place to stand at the stand. In step 5 the chair is moved to another place in the room; the layout then has
no violation and keeps all 15 objects.

The last row changes the layout in 65 scenes; the semantic review accepts a correction in 62 scenes. Apart from
the collision rates discussed above, the scores move by at most 0.05 points.
They do not measure the functional arrangement that the review corrects.

\paragraph{Inputs.}
Each variant repeats the run of Table~\ref{tab:panel_a_sceneeval} with one input removed and all other settings
unchanged, once per variant. \emph{W/o size model}: the language model writes the size of each object together
with its pose in the draft, and these sizes are kept. \emph{W/o robot terms}: no robot is given, so the access
checks, the access terms of Eq.~(\ref{eq:method:layout}), and the robot in the prompts are off. \emph{W/o task information}: the task
text, the list of required objects, and the list of objects that the task opens are not given, so all objects are
treated alike. Table~\ref{tab:layout_ablation} gives the SceneEval scores and, in the last three columns, the
result of the full method's verifier (Table~\ref{tab:layout_checks}) applied to the final layout of each variant
with the robot and the required objects of the task.

\begin{table*}[t]
\centering
\small
\setlength{\tabcolsep}{4pt}
\begin{tabular}{@{}lrrrrrrrrrr@{}}
\toprule
Variant & Placed$\uparrow$ & COL$_{ob}\downarrow$ & COL$_{sc}\downarrow$ & SUP$\uparrow$ & NAV$\uparrow$ & ACC$\uparrow$ & OOB$\downarrow$ & Access$\uparrow$ & Required$\uparrow$ & All$\uparrow$ \\
\midrule
WorldLayout (full) & 99.52 & 0.53 & 2.00 & 99.49 & 98.67 & 94.43 & 0.00 & 100.00 & 97.33 & 97.33 \\
w/o size model & 99.12 & 0.00 & 0.00 & 99.76 & 97.83 & 93.67 & 0.00 & 98.67 & 99.33 & 98.67 \\
w/o robot terms & 99.96 & 0.14 & 0.67 & 99.63 & 98.53 & 93.45 & 0.00 & 69.33 & 75.33 & 66.67 \\
w/o task information & 100.00 & 0.00 & 0.00 & 99.59 & 98.01 & 93.32 & 0.04 & 91.33 & 90.67 & 90.00 \\
\bottomrule
\end{tabular}
\caption{\textbf{WorldLayout input ablations on 150 scenes.}}
\label{tab:layout_ablation}
\par\smallskip{\footnotesize\raggedright Placed and the six SceneEval scores follow Table~\ref{tab:panel_a_sceneeval}. Access, Required and All give the percentages of scenes with no access violations, no required-object violations and no violations of any kind under the full verifier.
Without the size model, one scene produced no layout. Its objects count as unplaced and it fails all three verifier checks. SceneEval averages for this variant cover 149 scenes.\par}
\end{table*}

The SceneEval scores do not separate the variants. NAV and ACC are 0.1 to 1.1 points lower without each input.
Among scenes that produced a layout, the variants omit 0, 1, and 0 objects, respectively, versus 11 for the full method. Without the size model, one scene produced no layout and is excluded from the SceneEval averages.
One run per variant cannot establish an effect this small.

The last three columns separate two of the variants. Without robot terms, 46 of the 150 final layouts have an
access violation (no standing place in 32 scenes, an object on furniture too far from the free floor in 24, an
opening blocked in 4), and the share of scenes without any violation falls from 97.33\% to 66.67\%. Without
task information, 13 layouts have an access violation and 14 have a violation on a required object, since the
method does not know which objects the task uses; the share falls to 90.00\%.
The full method's verifier found no access violations in its own layouts. It flagged access violations more often when either robot terms or task information was removed.
Without the size model, the three columns differ from those of the full method by one to three scenes, in both directions; what the size model changes is the sizes themselves (Table~\ref{tab:size_accuracy}).

\paragraph{Size accuracy.}
The evaluation scenes have no reference sizes, so size accuracy is measured on rooms of the test split of the
size model's data (Sec.~\ref{sec:layout_method}), where each object has the size of its asset. The error of one
object is the mean of $|\ln(\hat{s}/s)|$ over its three dimensions, with $\hat{s}$ the predicted and $s$ the
reference size. We compare three sources of sizes on 30 test rooms with 716 objects: the size model; the
language model (GPT-6.1-sol, medium reasoning setting), which writes the sizes together with its draft when the
size model is removed; and the median size of the object's category in training scenes.

\begin{table}[ht]
\centering
\small
\setlength{\tabcolsep}{3pt}
\begin{tabular}{@{}lrrrrr@{}}
\toprule
 & All & Unseen & Floor & Other & Per room \\
Sizes from & (716) & (627) & (277) & (439) & (30) \\
\midrule
Category median & 0.481 & 0.515 & 0.288 & 0.603 & 0.345 \\
Language model & 0.556 & 0.594 & 0.305 & 0.715 & 0.373 \\
Size model & \textbf{0.333} & \textbf{0.367} & \textbf{0.173} & \textbf{0.434} & \textbf{0.192} \\
\bottomrule
\end{tabular}
\caption{\textbf{Object size error in 30 held-out rooms.}}
\label{tab:size_accuracy}
\par\smallskip{\footnotesize\raggedright Lower is better. Error averages $|\ln(\hat{s}/s)|$ across three dimensions. Parentheses give sample counts. Unseen denotes descriptions absent from training. Floor objects rest on the floor. Other objects are on a wall or another object. Per room averages the 30 room means.\par}
\end{table}

Table~\ref{tab:size_accuracy} gives the errors. The size model has an error of 0.333, against 0.556 for the
language model and 0.481 for the category median; these correspond to factors of 1.40, 1.74, and 1.62 per
dimension. The size model is more accurate than the language model in 28 of the 30 rooms; the difference of the
per-room errors is $-0.181$ (95\% bootstrap interval $-0.231$ to $-0.128$).

The reference is the size of the dataset assets, and the test rooms come from the same sources
as the training rooms; the numbers therefore measure agreement with these datasets and not transfer to the
evaluation scenes.

\subsubsection{Evaluation Details}
\label{sec:layout_eval}

\paragraph{Evaluation scenes.}
The layout evaluation (Table~\ref{tab:panel_a_sceneeval}) uses the 150 Panel~A tasks: 50 per difficulty
level, covering all 24 functional categories. For each task, the Room Designer is run once, and its design is the
input of every layout method. A design gives a rectangular room (sides 2.0--6.0\,m, floor area 9--36\,m$^2$,
height 2.7--3.2\,m) with one 0.9\,m door, and a list of objects. Each object has a category and a text
description of 6--15 words, but no size and no pose. The 150 designs contain 2{,}282 objects of 651
categories; Table~\ref{tab:layout_data} gives the statistics per scene. The tasks leave eight of the 863 required objects unnamed and call for opening or closing 43 objects across 37 scenes.

\begin{table}[h]
\centering
\small
\begin{tabular}{lrrrr}
\toprule
Per scene & Total & Min & Median & Max \\
\midrule
Objects                      & 2{,}282 & 5 & 15 & 42 \\
\quad required by the task   & 863     & 1 & 5  & 31 \\
\quad contextual             & 1{,}419 & 2 & 9  & 18 \\
\quad on another object      & 696     & 1 & 4  & 29 \\
Floor area (m$^2$)           & --      & 9.0 & 22.5 & 36.0 \\
\bottomrule
\end{tabular}
\caption{\textbf{Layout statistics for the 150 evaluation scenes.}}
\label{tab:layout_data}
\end{table}

\paragraph{Metrics.}
Every layout is converted to a SceneEval~\cite{sceneeval} scene and scored with SceneEval's released metric code
and default configuration. COL$_{ob}$ and COL$_{sc}$ are the shares of objects and of scenes with a collision.
SUP is the share of objects that rest on the support of their type, and NAV the share of the free floor that lies
in its largest connected region. ACC is the free share of the space in front of an object's functional sides, and OOB the share of objects
out of bounds. Our protocol differs from SceneEval's in its inputs.
(i) Objects are their bounding boxes, since the compared methods output boxes. A rug is therefore a solid box in
COL, NAV, and ACC, for every method.
(ii) SceneEval obtains support types and functional sides from a VLM on renders. We take the support type from
the scene plan where it names a parent (696 of the 2{,}282 objects) and query it from the object description
otherwise; functional sides are queried with SceneEval's prompt. Both are obtained once per scene and are the same
for all methods.
(iii) Scores are computed per scene and averaged over the scenes for which a method returns a layout.
(iv) An object that a method does not place is put at the room centre, on the floor, with the size the method
used, and is scored like any other object. This rule is ours and is discussed below.

\paragraph{Common setting.}
All methods receive the same 150 rooms and object lists: the category and description of each object and, where
the scene plan names one, the object it rests on (696 objects). The baselines are not given the task text or the list of required objects;
WorldLayout is. All baselines call GPT-6.1-sol at its medium reasoning setting, with text only unless the method
makes its own images (LayoutVLM~\cite{layoutvlm}, RoboLayout~\cite{robolayout}, DirectLayout~\cite{directlayout}). WorldLayout calls the same model at its highest
reasoning setting and shows it images of the layout in the review.
To separate the method from the reasoning setting, we ran WorldLayout once more on the 150 scenes with every call at the medium setting and nothing else changed.
At the medium setting it places 99.91\% of the objects and scores 0.54 / 2.67 (COL$_{ob}$ / COL$_{sc}$), 99.56 (SUP), 98.61 (NAV), 94.15 (ACC), and 0.03 (OOB), in 152\,s per scene. Its row in Table~\ref{tab:panel_a_sceneeval}, at the highest setting, is 99.52, 0.53 / 2.00, 99.49, 98.67, 94.43, and 0.00, in 479\,s.
The verifier reports no violation in 96.00\% of the scenes at the medium setting and in 97.33\% at the highest.
With one run per setting, we do not read these differences as effects of the setting.
The order of WorldLayout and the baselines in the columns COL$_{ob}$, COL$_{sc}$, SUP, NAV, and ACC is the same at both settings.

Object sizes belong to each pipeline. WorldLayout's sizes come from its size model; the baselines' come from a GPT
size call or from the method itself (LayoutGPT~\cite{layoutgpt}, DirectLayout, I-Design~\cite{idesign}). Floor objects cover 25\% of the floor on
average for WorldLayout and 33--38\% for the baselines. Per object, the median footprint of the baselines is
0.9--1.0 times ours; the largest part of the difference is rugs. Each pipeline predicts its own sizes, which may explain some of the difference in collisions.

\paragraph{Baseline settings.}
We run each method's released code up to the object boxes, with the following changes. We omit asset retrieval and simulation.
\begin{itemize}
\item \textbf{LayoutGPT}: released prompt conventions; in-context examples are retrieved from the training rooms
of the size model (Sec.~\ref{sec:layout_method}) instead of 3D-FRONT.
\item \textbf{Holodeck}~\cite{holodeck}: released floor and wall solvers. Its small-object stage needs AI2-THOR~\cite{kolve2017ai2}; we sample
non-overlapping positions on the parent's top face instead (10 attempts per item, its setting).
\item \textbf{I-Design}: three attempts per scene, each limited to 300\,s of backtracking and 600\,s without a
call to the language model. Without limits, its search was still running after 25 minutes in 4 of the first 11
scenes. 135 of the 150 scenes obtain a layout; 13 of the other 15 ended on these limits.
\item \textbf{LayoutVLM, RoboLayout}: both use the released code and default settings, including RoboLayout's one-shot mode.
\item \textbf{DirectLayout}: its fine-tuned generator is not released; GPT generates, judges, and refines, with
box drawings in place of renders.
\item \textbf{OptiScene}~\cite{optiscene}: weights retrained by us on its released data and not validated on its own benchmark;
greedy decoding; sizes from Holodeck's size call. Its training scenes have 3--14 floor objects of four room
types (32 of our 150 scenes), while 83 of our scenes have more than 14 objects. It is given the room area only,
and we take its $x$ and $y$ axes as the room's; rotating its layouts so that the long sides agree gives an OOB of
19.73\% and leaves its collision rates unchanged.
\item \textbf{SAGE}: one pass of its placement stage in place of its agent loop, without its critics and physics
simulation; objects on other objects are lowered by the 1\,cm that its sampler leaves above the parent.
\item \textbf{MANSION}~\cite{Che_2026_CVPR}: its floor and wall generators on a single room, with matrix and paired placement; small
objects as for Holodeck (100 attempts, its setting).
\end{itemize}
LayoutGPT, LayoutVLM, and RoboLayout have no input for the object that an object rests on; they receive it as a
text constraint for the 343 of the 696 objects for which the scene plan also states an \emph{on} relation.

\paragraph{Unplaced objects.}
Holodeck, I-Design, SAGE, MANSION, OptiScene, and WorldLayout return no position for 165, 401, 233, 285, 3, and 11
objects. For I-Design, the 401 are in its 135 scenes; the 246 objects of the 15 scenes without a layout also
count as not placed in the Placed column, but have no scores. The causes differ. Holodeck's solvers, and our sampling
for its small objects, find no position. I-Design removes nearly all of them itself when it corrects size
conflicts. SAGE's search stops at the first
object without a position, so 58 objects are never tried; its agent loop, which we do not run, may place
more. MANSION does not generate rugs: 61 of the 82 rugs, mats, and carpets of our lists stay unplaced.
WorldLayout leaves out 11 contextual objects in four scenes, and no required object.

The object list is given, so leaving an object out is a failure of the layout, and a score on placed objects only
would favour methods that drop the objects that are hard to place. We therefore score an unplaced object at the
room centre. The rule has a cost: an object at the centre overlaps what stands there, so most collisions of
Holodeck, I-Design, SAGE, and MANSION in Table~\ref{tab:panel_a_sceneeval} come from this rule and not from
their placed objects (Table~\ref{tab:panel_a_placed_only}). The same holds for WorldLayout: its 0.53\% and 2.00\%
come from three scenes with unplaced objects, one of which also has a collision between placed objects. As a check that does not use the centre, we count each unplaced
object as one colliding object without a position. COL$_{ob}$ / COL$_{sc}$ are then 0.43 / 2.67 for WorldLayout,
against 3.82 / 18.67 for DirectLayout, 7.90 / 40.00 for Holodeck, 12.65 / 79.33 for MANSION, and 13.09 / 69.33 for
SAGE.

\paragraph{Scores on placed objects only.}
Table~\ref{tab:panel_a_placed_only} scores the same layouts with the unplaced objects left out. On placed
objects, MANSION has no collision (0.00 / 0.00) and a higher NAV (98.95), and Holodeck has a higher NAV (98.75),
than WorldLayout (0.10 / 0.67 and 98.67). They place 87.5\% and 92.8\% of the objects, against 99.5\%. WorldLayout
has the highest ACC and the second-highest SUP under both ways of scoring. Its collision rates are the lowest
under the centre rule and under the count above, and the second lowest on placed objects (COL$_{sc}$ equal to Holodeck's). What holds under every
treatment is that WorldLayout is the only method that places over 99\% of the objects with an object collision
rate below 1\%.

\begin{table}[h]
\centering
\fontsize{7}{8.5}\selectfont
\setlength{\tabcolsep}{1.35pt}
\renewcommand{\arraystretch}{1.1}
\resizebox{\columnwidth}{!}{%
\begin{tabular}{@{}lrrrrrrr@{}}
\toprule
Method & Placed$\uparrow$ & COL$_{ob}\downarrow$ & COL$_{sc}\downarrow$ & SUP$\uparrow$ & NAV$\uparrow$ & ACC$\uparrow$ & OOB$\downarrow$ \\
\midrule
LayoutGPT~\cite{layoutgpt}       & 100.00 & 22.15 & 68.00 & 93.16 & 96.89 & 78.82 & 0.09 \\
Holodeck~\cite{holodeck}         & 92.77 & 0.17 & 0.67 & 96.81 & 98.75 & 93.43 & 0.67 \\
I-Design~\cite{idesign}          & 71.65 & 11.96 & 41.48 & 99.64 & 97.00 & 83.96 & 0.00 \\
LayoutVLM~\cite{layoutvlm}       & 100.00 & 8.00 & 30.67 & 99.28 & 95.56 & 88.06 & 22.77 \\
DirectLayout~\cite{directlayout} & 100.00 & 3.82 & 18.67 & 95.96 & 98.01 & 84.61 & 0.00 \\
OptiScene~\cite{optiscene}       & 99.87 & 69.93 & 98.00 & 92.02 & 96.70 & 72.02 & 25.16 \\
SAGE~\cite{sage}                 & 89.79 & 2.28 & 16.11 & 97.36 & 96.99 & 90.24 & 3.23 \\
RoboLayout~\cite{robolayout}     & 100.00 & 12.37 & 51.33 & 92.31 & 96.44 & 89.58 & 25.01 \\
MANSION~\cite{Che_2026_CVPR}     & 87.51 & 0.00 & 0.00 & 96.02 & 98.95 & 92.47 & 0.18 \\
\midrule
WorldLayout                      & 99.52 & 0.10 & 0.67 & 99.57 & 98.67 & 94.52 & 0.00 \\
\bottomrule
\end{tabular}%
}
\caption{\textbf{SceneEval scores for placed objects.}}
\label{tab:panel_a_placed_only}
\par\smallskip{\footnotesize\raggedright We score the layouts of Table~\ref{tab:panel_a_sceneeval} after excluding unplaced objects. Placed is the percentage of requested objects positioned. SAGE placed no objects in one scene, so its SceneEval averages cover 149 scenes.\par}
\end{table}

\paragraph{Out-of-bounds tolerance.}
SceneEval counts an object as out of bounds when fewer than 99\% of the points sampled on it lie over the floor.
LayoutVLM and RoboLayout accept 1\,cm beyond the wall and leave many objects on the wall line. With the floor
enlarged by 5\,mm, their OOB falls from 22.8\% and 25.0\% to 0.4\% and 1.4\%; their high OOB in
Table~\ref{tab:panel_a_sceneeval} is thus a matter of millimetres and not of objects far outside the room. We
keep SceneEval's rule in the table. Because the points are sampled, SAGE's OOB varies between 3.0 and 3.1 over
five scoring runs; no other cell changed.

\paragraph{Generation time.}
Time is the mean wall-clock time per scene with a layout and includes waiting for the API; all runs share a limit
of 30 requests per minute. The times are not like for like. I-Design's time includes its failed attempts in the
135 scenes but not the 15 scenes without a layout. OptiScene's time is the generation on one GPU, without the size call. MANSION's
19.88\,s excludes its size call, which is reused from an earlier run; with it, the time lies between 23.12\,s and
43.00\,s. The baselines' asset and simulation stages are not run. WorldLayout is the slowest method (479.5\,s):
it makes 7.41 calls per scene at the highest reasoning setting, and nearly all of its time is waiting for these
calls (Sec.~\ref{sec:layout_method}).

\paragraph{Statistics for the qualitative comparison.}
Figure~\ref{fig:compare} shows three scenes. Table~\ref{tab:layout_clean} gives the corresponding counts over all
scenes, under the rule of Table~\ref{tab:panel_a_sceneeval}. WorldLayout has no collision, no unsupported object,
and no out-of-bounds object in 142 of the 150 scenes, against 96 for DirectLayout, 80 for Holodeck, and 47 for
MANSION, the three baselines of the figure. The three scenes were chosen to show typical failures of the
baselines; they are not a random sample.

\begin{table}[h]
\centering
\small
\setlength{\tabcolsep}{4pt}
\resizebox{\ifdim\width>\columnwidth \columnwidth\else\width\fi}{!}{%
\begin{tabular}{@{}lrrr@{}}
\toprule
Method & Scenes & No collision & No COL / SUP / OOB \\
\midrule
LayoutGPT~\cite{layoutgpt} & 150 & 48 & 39 \\
Holodeck~\cite{holodeck} & 150 & 109 & 80 \\
I-Design~\cite{idesign} & 135 & 31 & 31 \\
LayoutVLM~\cite{layoutvlm} & 150 & 104 & 1 \\
DirectLayout~\cite{directlayout} & 150 & 122 & 96 \\
OptiScene~\cite{optiscene} & 150 & 3 & 1 \\
SAGE~\cite{sage} & 150 & 66 & 39 \\
RoboLayout~\cite{robolayout} & 150 & 73 & 0 \\
MANSION~\cite{Che_2026_CVPR} & 150 & 66 & 47 \\
\midrule
WorldLayout & 150 & \textbf{147} & \textbf{142} \\
\bottomrule
\end{tabular}%
}
\caption{\textbf{Scenes passing the SceneEval bounding-box checks.}}
\label{tab:layout_clean}
\par\smallskip{\footnotesize\raggedright The last column counts scenes with no collisions, unsupported objects or out-of-bounds objects. As in Table~\ref{tab:panel_a_sceneeval}, the scorer puts unplaced objects at the room centre.\par}
\end{table}

\subsubsection{Limitations}
\label{sec:layout_limits}

\paragraph{Objects as boxes.}
Every object is a box, in the method and in the evaluation.
A corner sofa, an open shelf, or the space under a table is treated as solid.
The checks and the scores therefore hold for the boxes; we do not test them on the meshes that later fill the boxes.

\paragraph{Access checks.}
The access checks measure free floor space and distances in the floor plan. Testing arm reach and grasping requires task execution.



\subsection{Mesh Generation}
\label{supp:world:mesh}

\paragraph{Generator choice.}
RoomGen receives from WorldLayout one fixed box per object with its
category, a short description and its support (the floor, a wall or
another object), but no image.
We chose its generator with a layout-to-mesh comparison that gave every
method this input.
The comparison used five single-room scenes of SAGE-10k~\cite{sage}
with 313 objects and 184 unique assets, each a pair of category and
description.
Methods that work through an image generated it from the description,
and no method saw a dataset image or mesh.
A shared assembler fitted every asset into its box with RoomGen's
fitting rule, given below.
We rendered four views of each fitted asset for visual evaluation.
A support check tested the mesh beneath each resting object
(Table~\ref{tab:supp:world:meshcompare}).

For both VQA columns, we used the same evaluator, Qwen2.5-VL-7B-Instruct~\cite{Qwen2.5-VL}, for all seven generators. For each view it answers two yes/no questions: does the object match its description (desc.), and does it belong to the named category (cat.)? Each score averages the probability of yes, normalized over yes and no, across the four views.
SigLIP-2~\cite{tschannen2025siglip} measures the cosine similarity of description and view, times 100.
R@5 counts a hit when the description is among the 5 nearest of the 154
distinct ones.
Support (Sup.) counts a resting object whose base lies within 2\,cm of its
supporter's mesh, as the median over 25 downward rays, at least half of
which are required to hit that mesh.
It covers the pairs in which both objects were produced, 146 for the
methods with full coverage, 91 for Infinigen and 71 for Infinigen V2.
Coverage (Cov.) is the share of assets produced.
For VQA, SigLIP-2 and R@5, we average over all 184 unique assets and use boxes for missing assets. Mean rank covers the five metrics marked $\uparrow$.

\begin{table}[!tbp]
\centering
\footnotesize
\setlength{\tabcolsep}{2.2pt}
\begin{tabular}{@{}lrrrrrrr@{}}
\toprule
& Cov. & \multicolumn{2}{c}{VQA $\uparrow$} & SigLIP-2 & R@5 & Sup. & Mean \\
\cmidrule(lr){3-4}
Method & (\%) & desc. & cat. & $\uparrow$ & (\%)\,$\uparrow$ & (\%)\,$\uparrow$ & rank\,$\downarrow$ \\
\midrule
\multicolumn{8}{@{}l}{\emph{Procedural}} \\
Infinigen~\cite{infinigenindoors} & 72.3 & 0.225 & 0.328 & 8.8 & 52.2 & 74.7 & 6.2 \\
Infinigen V2 & 48.9 & 0.176 & 0.242 & 7.8 & 36.4 & \textbf{98.6}$^{*}$ & 5.8 \\
\multicolumn{8}{@{}l}{\emph{Retrieval}} \\
Holodeck~\cite{holodeck} & \textbf{100.0} & 0.321 & 0.435 & 10.8 & 69.0 & 81.5 & 5.0 \\
\multicolumn{8}{@{}l}{\emph{Text $\rightarrow$ 3D}} \\
Shap-E & \textbf{100.0} & 0.346 & 0.406 & 12.0 & 71.2 & 82.9 & 4.4 \\
TRELLIS-text & \textbf{100.0} & 0.497 & 0.504 & 12.8 & 78.3 & 83.6 & 3.2 \\
\multicolumn{8}{@{}l}{\emph{Text $\rightarrow$ image $\rightarrow$ 3D}} \\
Hunyuan3D-2.1 & \textbf{100.0} & \textbf{0.607} & 0.510 & 15.3 & 91.3 & 84.2 & 2.0 \\
EmbodiedGen V2 & \textbf{100.0} & 0.583 & \textbf{0.528} & \textbf{15.5} & \textbf{91.8} & 86.3 & \textbf{1.4} \\
\midrule
\emph{SAGE meshes}~\cite{sage} & 100.0 & 0.505 & 0.552 & 13.2 & 79.9 & 100.0 & -- \\
\bottomrule
\end{tabular}

\vspace{2pt}
{\scriptsize Bold: best method per column. $^{*}$Over 71 pairs only.\par}
\caption{\textbf{Mesh generation for five SAGE-10k rooms.} Seven methods share boxes, descriptions and one assembler.}
\label{tab:supp:world:meshcompare}
\end{table}

\begin{figure*}[!tbp]
\centering
\includegraphics[width=0.76\textwidth]{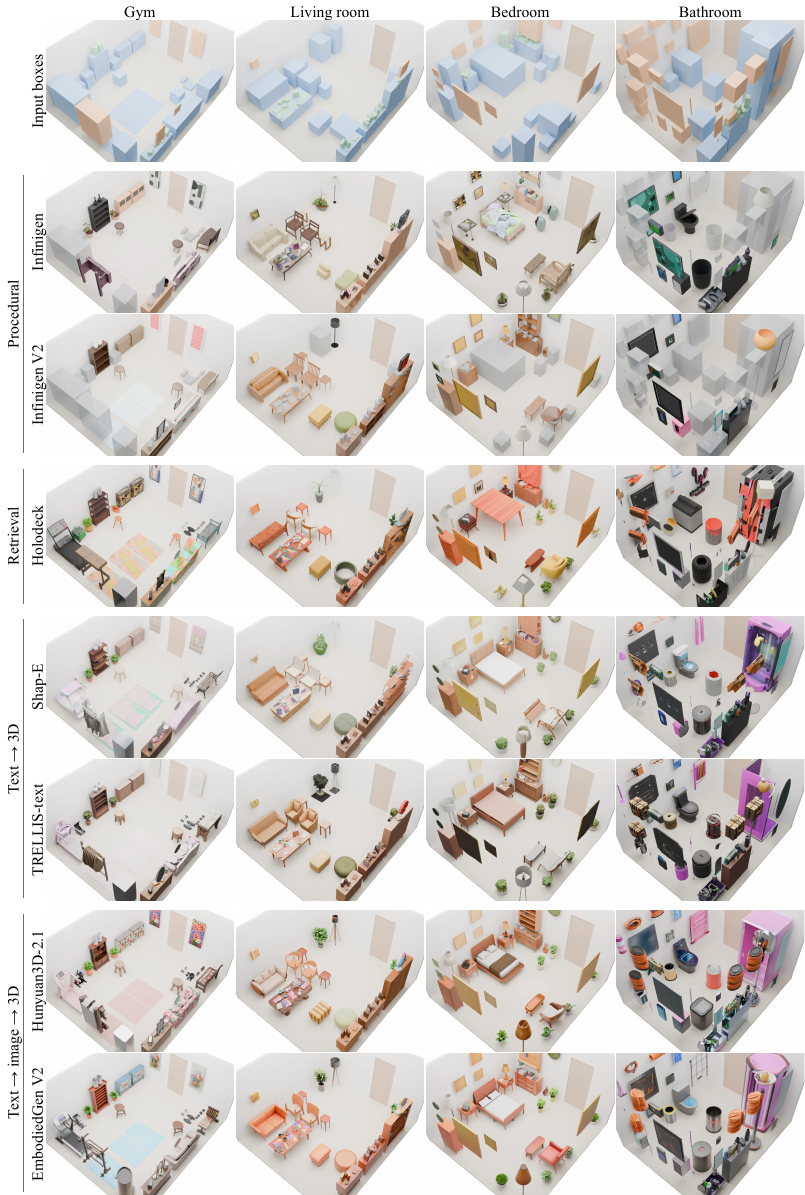}
\caption{\textbf{Layout-to-mesh comparison.} The top row shows the input boxes. Each method fits its meshes to the same boxes (Table~\ref{tab:supp:world:meshcompare}).}
\label{fig:supp:world:meshcompare}
\end{figure*}

EmbodiedGen V2~\cite{wang2026embodiedgen} has the best mean rank, 1.4
over the five scores, and leads on VQA category, SigLIP-2 and R@5.
Hunyuan3D-2.1~\cite{hunyuan3d2025hunyuan3d}, the other text-to-image-to-3D method, follows with mean
rank 2.0 and is better on VQA description (0.607 against 0.583).
Only Hunyuan3D-2.1 had a rule added at export that turns seating so
that its tallest side is at the back.
The 95\% bootstrap intervals of the two methods overlap on every score.
The largest half-widths are 0.06 for VQA, 0.8 for SigLIP-2, 7.3 points for R@5 and 8.8 points for support.

The two methods lead the native text-to-3D generators, TRELLIS-text~\cite{xiang2025structured} and
Shap-E~\cite{jun2023shap}, and Holodeck's retrieval from Objaverse~\cite{deitke2023objaverse}.
Holodeck's retrieval sometimes returns another category, such as a
kitchen table for the bed.
The procedural generators produced only 72.3\% (Infinigen) and 48.9\%
(Infinigen V2, using ProcFunc~\cite{raistrick2026procfunc}) of the assets and left the rest as boxes, translucent grey
in Fig.~\ref{fig:supp:world:meshcompare}.
Infinigen V2 has the best support score, but over only the 71 resting
pairs it produced.
Fig.~\ref{fig:supp:world:meshcompare} shows four of the five rooms; the
fifth, a restaurant with 122 objects, is left out for space.
Its top row is the input, with boxes that stand on the floor (blue), hang
on a wall (orange) or rest on another object (green).
Each column uses one camera for all rows, and the walls facing it are
hidden.

We therefore follow the text-to-image-to-3D route of EmbodiedGen V2.
SAGE's own meshes, which no method saw, are a reference row and are not
ranked.
EmbodiedGen V2 scores higher than SAGE's meshes on VQA description
(0.583 against 0.505) and lower on the support check (86.3\% against 100\%).
The EmbodiedGen V2 row of the comparison ran the pipeline, settings and
fitting that RoomGen used for Table~\ref{tab:task_readiness_expansion}.
Its scores thus measure the generator of the reported rooms.
The comparison does not test articulation, collisions or physical
stability.

\paragraph{Assets and sharing.}
RoomGen generates a mesh for every object of a verified layout.
An asset is identified by its category and description only, without
the instance ID, pose or box size.
Objects of a room with the same category and description share one
generated asset, and each instance is fitted to its own box.
In the 610 runs behind Table~\ref{tab:task_readiness_expansion}, retrieval supplied no
object, and RoomGen generated 6{,}025 assets for 6{,}753 placed objects.
WorldArt later adds joints to some of these meshes
(Sec.~\ref{supp:world:art}).

\begin{table}[!tbp]
\centering
\footnotesize
\setlength{\tabcolsep}{4pt}
\begin{tabularx}{\columnwidth}{@{}lX@{}}
\toprule
Step & Setting or test \\
\midrule
Prompt & ``Angled 3D view of one \{description\}, centered, no cropping, no occlusion, isolated product photo, placed horizontally, no surroundings, high-quality appearance, vivid colors, on a plain clean surface, 3D style revealing multiple surfaces'' \\
Text to image & $1024{\times}1024$ pixels, 25 steps, guidance 7, 3 images per try, at most 2 tries, seed 0 on the first \\
Image check & U$^2$-Net mask at least 5 pixels from the image border, else the last candidate \\
Image to 3D & seed 0, one retry after a failed inference, $2048{\times}2048$ texture, at most 200{,}000 faces \\
\midrule
Thin-axis turn & box and mesh thin (smallest extent below 0.35 of the next) along different axes \\
Yaw snap & tightest footprint rectangle below 0.93 of the axis-aligned area and turned by more than $3^\circ$ \\
Quarter turn & footprint aspect above 1.3 with box aspect below $1/1.3$, or the reverse \\
Tip & proportion error above 1.2 and more than halved by a $90^\circ$ tip \\
\bottomrule
\end{tabularx}
\caption{\textbf{RoomGen settings and rotation checks.}}
\label{tab:supp:world:mesh}
\end{table}

\paragraph{Generation chain.}
The generator is the released text-to-3D pipeline of EmbodiedGen
V2~\cite{wang2026embodiedgen}, version 2.1.0, run with its VLM quality
checkers switched off.
The object description fills a fixed prompt that asks for an angled view
of one centred object without surroundings
(Table~\ref{tab:supp:world:mesh}).
Stable Diffusion 3.5 Medium~\cite{sd35} generates candidate images, a
U$^2$-Net~\cite{qin2020u2} background remover masks the object, and SAM~3D~\cite{sam3d}
lifts one image to 3D.
These two generators are the frozen ``SD~3.5'' and ``SAM~3D'' of
Fig.~\ref{fig:world_generator}.
The pipeline remeshes the result and bakes its texture
from renders of the generated 3D Gaussians~\cite{kerbl20233d}.

Without the checkers, the selected image is the first whose mask stays
at least 5 pixels from the image border, or the last candidate if none
does.
The last-candidate fallback chose 35 of the 6{,}025 images.
Each asset is generated once, and both models start from seed 0.
A material that the exporter marks fully metallic by default, without a
metallic map, is made non-metallic.

\paragraph{Orientation and fitting.}
A fixed $90^\circ$ rotation about the vertical axis aligns the mesh
front with the front axis of the object frame.
Four corrective rotations follow when their tests in
Table~\ref{tab:supp:world:mesh} hold.
The proportion error of the tip test sums the absolute differences
between the mean-centred log extents of mesh and box.
For example, the thin-axis turn aligns a picture generated lying flat
with its wall box. It applied to 352 of 6{,}025 assets.

Let $\mathbf b_o$ be the width, depth and height of the box of object
$o$, and $\mathbf e_o$ the extents of its oriented mesh. RoomGen scales
the mesh along each axis by
\begin{equation}
s_{o,k}=b_{o,k}/e_{o,k},\qquad k\in\{1,2,3\},
\label{eq:supp:world:mesh:scale}
\end{equation}
centres it horizontally in the box, and rests its lowest point on the
box base.
The fitted mesh fills the box exactly, so its proportions follow the
box, not the generated asset, wherever the two differ.
Fitting changes neither the box nor its pose.
The shared assembler of the comparison applied the yaw snap, quarter
turn, tip and this scaling to every method.
The fixed $90^\circ$ rotation and the thin-axis turn belong to the
generation step of EmbodiedGen V2, which the comparison's EmbodiedGen V2
row shares with RoomGen.

\paragraph{Pose and support.}
A rigid pose then rotates the fitted mesh about the vertical axis to the
layout yaw and moves its bottom centre to that of the box.
The top of a box is not always the top of its mesh; the board of an open
shelf or a basin can lie lower.
An object resting on another object is therefore seated on the mesh of
its support, and supports are seated before the objects they carry.
Downward rays on a 1\,cm grid, coarser when it would exceed 10{,}000
rays, measure the support mesh from above.
As in the support regions of HSM~\cite{pun2026hsm}, a usable patch is
near-horizontal (upward normal component at least 0.95), covers at least
0.02\,m$^2$, and has sides of at least 5\,cm.

The object moves only vertically, onto the highest such patch under its
footprint that lies at most 1\,cm above its base.
The seat is refused when no such patch exists or when the patch does
not reach under the footprint centre.
It is also refused when other geometry of the support rises more than
1\,cm above the patch under the footprint.
A refused object keeps its pose, and the reason is recorded.
Meshes and support relations stay unchanged.
The comparison has no seat; there, 86.3\% of the resting EmbodiedGen V2
objects lie within 2\,cm of their supporter's mesh.

\paragraph{Room structure.}
The room structure is built from the room record of the layout.
It consists of a 5\,cm floor slab over the room polygon and straight
wall prisms whose inner faces lie on the wall lines of the layout.
A doorway is an open passage through its wall up to the door height,
with the wall closed above it and no door panel.
RoomGen refuses a doorway narrower than 0.6\,m or lower than 1.8\,m.
It accepts door openings only and builds no windows.

\paragraph{Checks, failures and IDs.}
This stage checks geometry only.
A fitted mesh passes when it is finite and non-degenerate, has the
extents of its box, and stands bottom-centred in it.
A delivered pose passes when it equals the layout pose up to the
recorded vertical seat.
These checks omit appearance, clearance and mesh collisions. Sec.~\ref{sec:world_integration} describes the collision recheck.

A failed asset or check stops the room, with no placeholder and no
retry of the asset.
In the 610 runs, every asset passed on its first generation.
Each mesh and pose is stored under the ID of its box.
A delivery passes only when its IDs equal those of the layout, so
instances of one asset keep distinct IDs.
Each delivered mesh has a SHA-256 hash. Resumed runs and the articulation stage verify these hashes.
\subsection{Task-Required Object Articulation}
\label{supp:world:art}

\paragraph{Object and part selection.}
WorldArt starts from the room delivered by RoomGen
(Sec.~\ref{supp:world:mesh}), where every object is a rigid mesh.
One call to GPT-6.1-sol~\cite{openai2026gpt61sol} at reasoning effort
\emph{xhigh} decides which objects receive joints (``Select parts'' in
Fig.~\ref{fig:world_generator}).
It receives the task, the room, and each object's ID, category,
description, role, size, pose, support, and the image its asset was
generated from.
The room design has already flagged the objects whose moving part the
task operates and named that part.
The call selects every object the task operates (basis \emph{task}).
Whatever the task, it also selects every piece of furniture or
wall-mounted item with doors, drawers, or hinged lids (basis
\emph{consensus}).
In the last 40 Panel~A rooms and in Panel~B, a third basis,
\emph{structure}, also admits other objects built with a moving part,
such as toolboxes, vises, and windows.
The part labels of a selected object are the doors, drawers, and hinged
lids or seats that its image shows.
In the reported rooms, 44 of the 384 objects selected in Panel~A and
none of the 1{,}135 selected in Panel~B have the task basis.

\begin{figure*}[t]
\centering
\includegraphics[width=\textwidth]{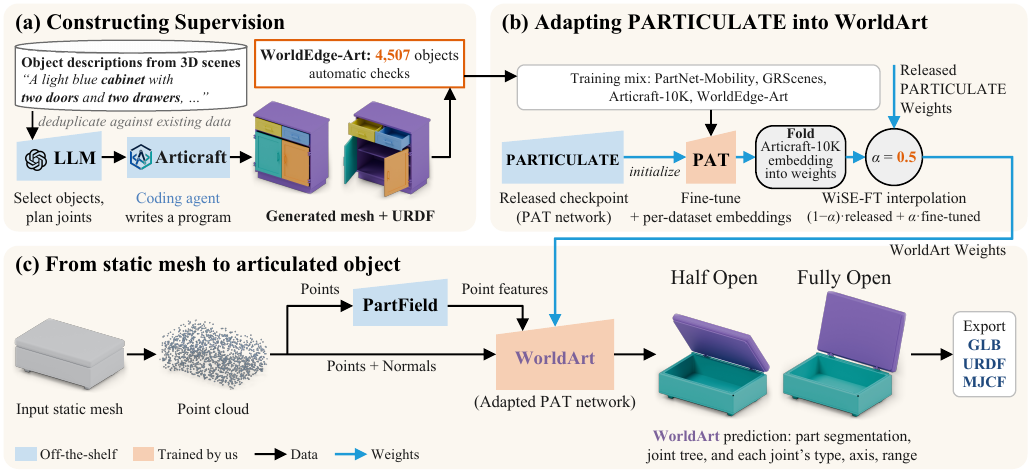}
\caption{\textbf{WorldArt.} (a)~WorldEdge-Art training data. (b)~Adaptation of PARTICULATE. (c)~Articulation of a static mesh.}
\label{fig:supp:world:art:pipeline}
\end{figure*}

\paragraph{Reuse, retrieval, and prediction.}
\emph{Reuse} keeps an existing object's own joints when every selected
part binds to a verified joint of that object's MuJoCo~\cite{todorov2012mujoco} model.
During room expansion, earlier objects keep their geometry. If an
earlier object lacks a joint the task needs, the expansion stops.
\emph{Retrieval} (``Retrieve \& Verify'') takes a complete articulated
asset from a library.
\emph{Prediction} (``Fallback'') runs the adapted
PARTICULATE~\cite{li2026particulate} on the object's own mesh
(Fig.~\ref{fig:supp:world:art:pipeline}c).
Neither reuse nor retrieval was active in the rooms reported in Sec.~\ref{sec:TaskGenRoom}. Every object examined there was newly generated and had no joints
to reuse, and no library was configured. Every joint in these rooms
therefore comes from prediction.
Fig.~\ref{fig:supp:world:art:pipeline} draws the training and the
prediction route; object selection, reuse, retrieval, the structural
screen, and the critic are not drawn.

\paragraph{Retrieval.}
Retrieval ports the articulated retrieval of
SceneSmith~\cite{pfaff2026scenesmith} with its preprocessed ArtVIP~\cite{jin2026artvip} library. The user supplies a catalog with hashes of the index and asset files.
PartNet-Mobility~\cite{xiang2020sapien} is an optional source, off by
default.
The OpenCLIP ViT-H/14 text encoder~\cite{ilharco2021openclip} (DFN5B weights~\cite{fang2024data}, 1{,}024 dimensions)
embeds the object's full description. Cosine similarity compares this
embedding with the library's CLIP image embeddings of assets of the same
placement type: floor-standing, hand-held, wall-mounted, or
ceiling-mounted.
By default, the pool contains the five closest matches across enabled sources, with no score threshold.

For a new room, retrieval runs before WorldLayout on objects marked for
articulation or belonging to a recognized closed category. Candidates
are tried in similarity order. An optional size prior reorders the pool
by the sum of absolute width, depth and height differences in metres.
The prior is the median size of the reference category closest to the
description in CLIP text space. The first asset that converts to MJCF
with verified native geometry and joints supplies its dimensions to
WorldLayout. If none converts, RoomGen generates the mesh. If the
selected asset later fails its visual or joint checks, RoomGen generates
a replacement in the reserved box and WorldArt can predict its joints.

When preparing an existing object for execution outside room expansion,
WorldArt can search after layout if the object's own joints cannot be
reused and its front axis is known. Here the L1 comparison uses the
object's current width, depth and height. Candidates are tried in that
order until one passes. A candidate needs a joint that passes the
structural screen below. GPT-6.1-sol reviews the object and candidate
images and the candidate's closed and open joint poses. It checks the
description (including colour, material and style), moving parts,
placement and dimensions. The joint critic then reviews the motion.
An accepted asset retains the object's ID and placement and supplies its own appearance, collision geometry, masses and joints at native scale. Otherwise, the predictor uses the original mesh.

\paragraph{Prediction.}
The predictor receives only the object's fitted mesh from RoomGen and
its up axis. No text, image, or part label reaches it.
The mesh is centred and scaled to unit longest side.
The network takes points sampled on it, with normals and frozen
PartField features~\cite{liu2025partfield}. It predicts up to 16 parts
as face labels, the joint tree, and the type, axis, pivot, and limits of
every joint.
The parts are the faces of the generated mesh grouped by label, and the
inverse transform restores their metric size and pose. Size and
appearance thus stay those of the fitted RoomGen mesh.
The predictor writes GLB, URDF, and MJCF files. Its own gate refuses a
result that is not one rooted tree of valid, moving joints or that fails
to load and step in MuJoCo~\cite{todorov2012mujoco}.
The example in Fig.~\ref{fig:supp:world:art:pipeline}c is the seed-zero
prediction for the storage ottoman of Fig.~\ref{fig:art:comparison}, a
test-only category, with every predicted joint at half and full range.

\paragraph{Candidate selection and the critic.}
A candidate is one prediction or one retrieved asset for an object.
The structural screen keeps a revolute or prismatic joint whose limits
contain the rest pose 0 (to 0.05\,rad or 1\,cm) and that travels at
least $\pi/4$\,rad or 10\,cm from rest.
A kept revolute joint also spans at most $\pi+0.1$\,rad and moves its part
by at least 5\% of the object's bounding-box diagonal, which excludes
knobs.
For a task-required object, the screen also requires the joint type to
match the part label.
A drawer takes a prismatic joint, a door either type, and a lid, seat,
flap, or hatch a revolute joint.
The critic (``VLM critic'' in Fig.~\ref{fig:world_generator}) is GPT-6.1-sol at reasoning effort
\emph{xhigh}, called through its API without fine-tuning; this is what
``frozen'' means here.
For every screened joint it sees a closed image at the rest pose and an
open image at the far limit. Each image has four oblique MuJoCo views of
256 pixels, with the moved body orange and the rest grey. The text gives
the category, description, part labels, and joint types.
For each joint the critic names the orange body and answers three
checks. It checks that the body is one of the selected parts and opens
like it, that the first image shows the body closed, and that the second
shows it open.
An object passes only if every screened joint passes every check.
The critic omits axis and limit errors. Validation against human labels remains open.

Prediction runs at most three times on the same mesh, and the runs can
differ.
A run fails when the gate refuses it, when no joint passes the screen,
or when the critic rejects it. Crashes and service outages are not
counted.
After three failed runs, a task-required object stops the room, which is
not delivered. Other objects are delivered rigid, at once if the
gate refuses a run or no joint passes the screen.
In 110 of the 150 Panel~A rooms, the critic asked only the identity
check, and a task-required object without an accepted joint was
delivered rigid. The endpoint checks and the stopping rule were added
before the other Panel~A rooms and all Panel~B rooms were generated.

\paragraph{Export for simulation.}
From the predictor's files, or from a retrieved asset's MJCF, the export
writes a room or a whole world as one MuJoCo scene. The scene holds the object
IDs, the joints and their closed and open positions, the initial state,
and the task.
The export takes the rest pose 0 as closed: if 0 lies inside a predicted
range, the nearer limit moves to 0 and the far limit is open.
CoACD~\cite{wei2022approximate} splits each nonconvex component of a rigid mesh into at most 64
convex pieces and each articulated part into at most 32. The parts of
one object do not collide with each other.
Densities are 500\,kg/m$^3$, or 100\,kg/m$^3$ for articulated objects,
with masses clipped to 0.1--500\,kg, and friction is 0.8.
Articulated bases are fixed, and a hinge that gravity opens at rest gets
a preloaded spring that holds it closed. A retrieved asset keeps its own
physical data.
The export requires the scene to reload and stay finite over 100 steps,
and every articulated object to have an accepted critic review. The rechecks of the
room are those of Sec.~\ref{sec:world_integration}.

\paragraph{WorldEdge-Art.}
WorldEdge-Art supplies program-derived supervision
(Fig.~\ref{fig:supp:world:art:pipeline}a).
Articraft~\cite{zhou2026articraft} turns an object description into a
program that specifies parts, hierarchy, joint axes, and ranges, and
compiles it to a mesh and a URDF file.
The object descriptions come from existing 3D scenes.
After normalization and merging of exact duplicates, descriptions close
to existing datasets are removed (Jaccard similarity of word-boundary
character trigrams of at least 0.80 within a headword group).
A language model screens the rest for articulated objects and lists each
object's moving parts and joint types.
Per-batch quotas on category, part count, joint type, and hierarchy then
select the targets. Final templates exclude
within-category matches at 0.60 or more.
Admission requires a rooted tree, supported joints, finite geometry and
motion, valid labels, and at most 16 parts.
The filters control coverage, and admission enforces structural
consistency; neither assesses mechanical realism or physical quality.
Objects that share a design group or source description go to one split,
with the test split taking precedence. Ottoman, locker, trunk, and workbench are test-only, with
known training neighbours excluded, though unrecorded overlap may remain.
The 4{,}507 objects split 3{,}606/760/141 into training, validation, and
test. The deployed model used 3{,}354 eligible training objects of an
earlier version and none of the 240/27/41 objects added later.

\begin{table}[!tbp]
 \centering
 \footnotesize
 \setlength{\tabcolsep}{4pt}
 \begin{tabularx}{\columnwidth}{@{}lX@{}}
  \toprule
  Setting & Value \\
  \midrule
  Initialization & PARTICULATE~\cite{li2026particulate} \\
  Trainable network & Full PAT and source embeddings \\
  Feature encoder & Frozen PartField~\cite{liu2025partfield} \\
  Hardware & Four GPUs \\
  Updates / seed & 20{,}000 / 42 \\
  Effective batch & $4\times16\times2=128$ \\
  Optimizer & AdamW~\cite{loshchilov2019decoupled}; $\beta=(0.9,0.999)$ \\
  Weight decay / clipping & 0.01 / gradient norm 1 \\
  Peak learning rate & $1.28\times10^{-5}$ \\
  Schedule & 1{,}000 warm-up updates; cosine decay to $0.1\times$ peak, without restarts \\
  Decoder points / parts & 2{,}048 / at most 16 \\
  Per-part reserve & $\min(16,\text{available points})$ \\
  Source probabilities & PartNet-Mobility .45 / GRScenes objects .20 / Articraft-10K .25 / WorldEdge-Art (earlier version) .10 \\
  Parameter interpolation & $\alpha=0.5$; one fixed source folded in \\
  \bottomrule
 \end{tabularx}
 \caption{\textbf{WorldArt adapts PARTICULATE on mixed sources.}}
 \label{tab:supp:world:art:train}
\end{table}

\paragraph{Training.}
WorldArt keeps PARTICULATE's Part Articulation Transformer (PAT) and
adds a zero-initialized source embedding $e_k$ for dataset $k$ to the
point tokens and part queries (Fig.~\ref{fig:supp:world:art:pipeline}b).
Training samples PartNet-Mobility~\cite{xiang2020sapien}, GRScenes
objects~\cite{wang2024grutopiadreamgeneralrobots},
Articraft-10K~\cite{zhou2026articraft}, and WorldEdge-Art with
probabilities 0.45/0.20/0.25/0.10. This is the ``mixed-source
supervision'' in Sec.~\ref{sec:art:main}.
All PAT parameters and source embeddings are trained for 20{,}000
updates with an effective batch of 128, gradient accumulation included
(Table~\ref{tab:supp:world:art:train}). The range loss covers
every ground-truth moving joint, including zero ranges, whereas the
released objective selects nonzero ranges.
We then fix the source to Articraft-10K ($k=2$), chosen before testing
because its categories cover every Lightwheel~\cite{lightwheel2025simready} category, and fold its
embedding into the weights.
We interpolate the resulting and released weights with WiSE-FT~\cite{wortsman2022wise}:
\begin{equation}
\begin{aligned}
\theta_{\mathrm{WorldArt}}&=(1-\alpha)\,\theta_{\mathrm{released}}\\
&\quad+\alpha\operatorname{Fold}(\theta_{\mathrm{adapted}},2),
\quad\alpha=0.5.
\end{aligned}
\label{eq:supp:world:art:wise}
\end{equation}
We used these same weights for room generation and the evaluation in
Table~\ref{tab:art:main_results}. Because the training mixture, source
conditioning, sampling, and weight averaging all changed together,
the comparison does not isolate the contribution of WorldEdge-Art or
any other individual change.
PartNet-Mobility is both a training source and a test set; none of the
74 test objects is in the training data of WorldArt or of the released
PARTICULATE weights.
Lightwheel is not a training source.

\paragraph{Evaluation protocol.}
Table~\ref{tab:art:main_results} scores the predictor alone, without screen,
critic, or retrieval.
It uses 241 of the 243 Lightwheel
objects~\cite{lightwheel2025simready,li2026particulate}, 74 of 77
PartNet-Mobility objects, and the 141 WorldEdge-Art test objects.
Three PartNet-Mobility objects lack meshes, and two Lightwheel objects
are inadmissible. Twenty-six Lightwheel objects keep the upstream
conversion, in which 25 sinks lack their generic-joint handles.
``Matched processed inputs'' means both networks receive identical
sampled geometry and PartField features. Each makes one seed-zero
prediction per object, with 40{,}000 context and 102{,}400 decoding
points, connectivity refinement, and export of the full tree.
Scoring uses 100{,}000 points instead of the source paper's one million,
so the scores are not comparable with its published values.
The test sets had been inspected during development and are not a blind
test.

\paragraph{Metrics.}
Following PARTICULATE, meshes are normalized to unit longest side, and
parts are matched at minimum cost on centroid distances.
We average $d$ symmetrically over $N_p$ predicted parts $p_i$ and $N_g$ ground-truth parts $g_j$:
\begin{equation}
\mathcal D_d=\frac12\Big[\frac1{N_p}\sum_i\widetilde d(p_i)
+\frac1{N_g}\sum_j\widetilde d(g_j)\Big].
\label{eq:supp:world:art:sym}
\end{equation}
Here $\widetilde d$ is the matched value or an unmatched penalty: $-1$
for gIoU, 0 for mIoU, and half the predicted bounding-box diagonal for
PC.
gIoU is the generalized intersection over union~\cite{Rezatofighi_2019_CVPR} of 3D part boxes, and
mIoU the overlap of axis-aligned part boxes.
PC is the symmetric squared Chamfer distance between matched parts, and
OC the same distance between whole objects without matching.
For PartNet-Mobility, both articulated OC means round to 0.001 in
Table~\ref{tab:art:main_results}. The archived object-level means are 0.000979 for PARTICULATE
and 0.001265 for WorldArt, so PARTICULATE has the lower OC under the unrounded comparison rule in Table~\ref{tab:art:main_results}.
The rest-state columns score the given pose; the articulated columns
move every joint to its maximum extent.
In Fig.~\ref{fig:art:comparison} and below, $s$ is the motion state as a
fraction of each joint's range.

\begin{table}[!tbp]
\centering
\footnotesize
\setlength{\tabcolsep}{1.2pt}
\renewcommand{\arraystretch}{1.05}
\begin{tabular*}{\columnwidth}{@{\extracolsep{\fill}}lrrrrrrr@{}}
\toprule
 & \multicolumn{3}{c}{(a) Rest state} & \multicolumn{3}{c}{(b) Articulated} & (c) \\
\cmidrule(lr){2-4}\cmidrule(lr){5-7}\cmidrule(l){8-8}
Method & gIoU$\uparrow$ & PC$\downarrow$ & mIoU$\uparrow$ & gIoU$\uparrow$ & PC$\downarrow$ & OC$\downarrow$ & Motion$\uparrow$ \\
\midrule
\multicolumn{8}{@{}l}{Lightwheel (241)} \\
PARTICULATE & 0.337 & 0.163 & 0.573 & 0.310 & 0.201 & 0.008 & \textbf{0.014} \\
\textbf{WorldArt} & \textbf{0.380} & \textbf{0.151} & \textbf{0.606} & \textbf{0.349} & \textbf{0.185} & 0.008 & 0.008 \\
\midrule
\multicolumn{8}{@{}l}{PartNet-Mobility (74)} \\
PARTICULATE & \textbf{0.884} & \textbf{0.001} & 0.886 & \textbf{0.845} & \textbf{0.005} & 0.001 & \textbf{0.610} \\
\textbf{WorldArt} & 0.882 & 0.003 & \textbf{0.888} & 0.840 & 0.007 & 0.001 & 0.513 \\
\midrule
\multicolumn{8}{@{}l}{WorldEdge-Art (141)} \\
PARTICULATE & 0.317 & 0.113 & 0.505 & 0.308 & 0.242 & 0.023 & $-0.079$ \\
\textbf{WorldArt} & \textbf{0.437} & \textbf{0.086} & \textbf{0.580} & \textbf{0.423} & \textbf{0.178} & \textbf{0.016} & \textbf{0.026} \\
\bottomrule
\end{tabular*}

\vspace{2pt}
{\scriptsize Bold: the better value where the rounded values differ.\par}
\caption{\textbf{Corrected motion gIoU (c)} for the predictions in Table~\ref{tab:art:main_results}, reproduced in (a, b).}
\label{tab:supp:world:art:motion}
\end{table}

\paragraph{Corrected motion.}
The inherited box-motion approximation behind the articulated gIoU of
Table~\ref{tab:art:main_results} ignores lower joint limits.
The corrected motion gIoU transforms each part box with its surface and
accounts for lower limits and both motion types. Moving-box overlap uses
2{,}048 samples. Matching, symmetric averaging, and the unmatched score
$-1$ are unchanged.
From PARTICULATE to WorldArt it falls on Lightwheel (0.014 to 0.008) and
PartNet-Mobility (0.610 to 0.513) and rises on WorldEdge-Art ($-0.079$
to 0.026; Table~\ref{tab:supp:world:art:motion}).
Object-level bootstrap intervals (10{,}000 resamples) resolve the last
two changes but not the Lightwheel one. They are conditional on the
objects and checkpoints, since each model has one training seed.
On WorldEdge-Art both models remain below the 0.230 of a one-part
baseline. Its single body is matched to a fixed ground-truth part, so
its score does not change with motion. The gain therefore does not show
that motion is recovered.
A post hoc audit measures axis, limit, and face-centroid errors at five
range fractions. It finds lower errors for WorldArt than for PARTICULATE
on Lightwheel and WorldEdge-Art, and higher errors on PartNet-Mobility.

\begin{figure*}[t]
\centering
\includegraphics[width=0.87\textwidth]{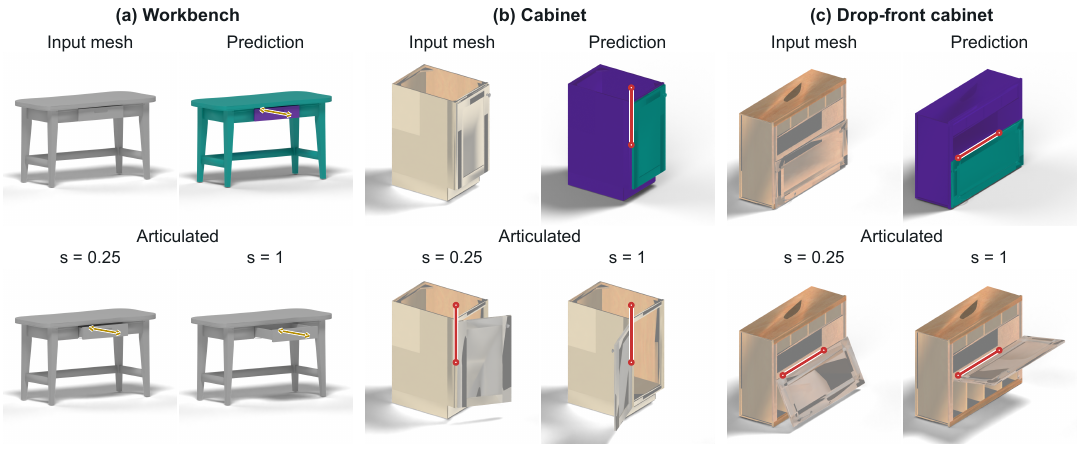}
\caption{\textbf{WorldArt predictions for three objects.}
Input meshes, predicted parts and motion constraints, and two states of motion.}
\label{fig:supp:world:art:gallery}
\end{figure*}

\begin{figure*}[t]
\centering
\includegraphics[width=\textwidth]{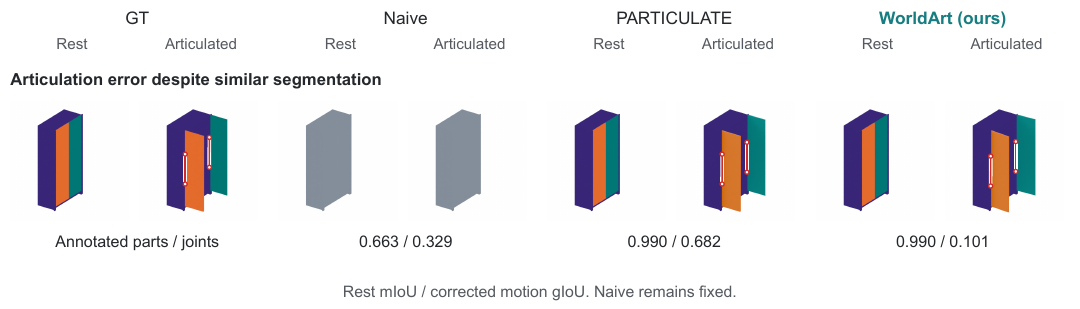}
\caption{\textbf{Two-door cabinet, PartNet-Mobility 46655.} Ground truth, the one-part baseline (Naive), PARTICULATE, and WorldArt at rest and at $s=1$.}
\label{fig:supp:world:art:regression}
\end{figure*}

\paragraph{Qualitative examples.}
The ottoman of Fig.~\ref{fig:art:comparison} is a seed-zero prediction for a
test-only WorldEdge-Art category. PARTICULATE moves the outer frame and
leaves most of the lid fixed, whereas WorldArt approximately recovers the
lid. Some small surfaces stay mislabelled, and hinge and limit errors
remain.
Each method moves its parts with its own tree, axes, and limits under
shared rendering rules; only the ground truth uses ground-truth motion.
Figs.~\ref{fig:supp:world:art:gallery}
and~\ref{fig:supp:world:art:regression} add three selected predictions
and one case of the PartNet-Mobility regression. Selected cases do not
estimate a success rate.

Fig.~\ref{fig:supp:world:art:gallery} drives each prediction to
$s=0.25$ and $s=1$ with WorldArt's own joint tree and limits.
Gold arrows mark prismatic directions, and red markers revolute axes.
On the workbench (a, seed zero), geometry goes to the active joint, but
the drawer faces stay fixed and the limit is too large.
PartNet-Mobility object 35059 (b) is shown at seed nine, chosen after
inspection for both models; at seed zero, used in the tables, WorldArt's
motion score is worse.
On PartNet-Mobility object 48167 (c, seed zero), the surface partition is
nearly right, but the flap faces stay fixed, with axis and range errors.
On the two-door cabinet of Fig.~\ref{fig:supp:world:art:regression},
object 46655, the two models reach the same rest mIoU (0.990) and similar
articulated silhouettes.
The corrected motion gIoU is nevertheless 0.101 for WorldArt against
0.682 for PARTICULATE.
Overlap of the thin doors is sensitive to axis and range errors.

\subsection{World Integration}
\label{supp:world:integration}

\paragraph{Grid and initial room.}
Rooms lie on a grid of shared column widths and row depths, each fixed
at first use. Besides its entrance, a new room gets one to three door
openings of 0.9\,m$\times$2.1\,m on walls facing an empty cell, fewer
when fewer such walls exist. The initial
room has at least two. In all 30 worlds in Table~\ref{tab:task_readiness_expansion}, the initial room has the designed type, measures 5.0\,$\times$\,5.0\,$\times$\,3.0\,m and has two free doorways.

\paragraph{Attachment.}
The next room fills the cell behind the first free doorway, in
identifier order, whose cell connects to the open exterior. Along the
shared wall the new room inherits the parent's width or depth. Along the
other axis it takes the designed room's extent, rounded to 0.5\,m and
clamped to 2.5--8\,m, unless another room has fixed that column or row.
New columns and rows start flush against built ones, so rooms never
overlap. Each room's footprint, 3\,m wall height and doorways come from
the world.

\paragraph{Permitted opening edits.}
Attachment changes existing rooms in three ways only. The used doorway
becomes a door, and a neighbour's free doorway facing the new room
becomes a second door into it. A neighbour's window on a newly shared
wall is removed. The symbolic rounds of Sec.~\ref{sec:next_task_proposal} may
also seal a doorway after three consecutive refusals of rooms behind it,
or after two consecutive identical ones. Sealing closes the opening. Existing
object records and the agent state stay unchanged, which a preservation
check enforces in the symbolic rounds.

\paragraph{Object identity.}
The Room Designer gives each object an ID unique in its room. WorldLayout
returns each kept ID once, and the mesh stage returns exactly the
layout's IDs. WorldArt's articulated delivery is verified against the
same ID, and the exported scene names parts and joints under it. Qualifying each ID with its room ID keeps new and existing IDs disjoint in the joined world (Eq.~\ref{eq:method:identity}).

\paragraph{Checks before commit.}
A structural audit requires aligned door halves, no overlapping rooms, a
connected world and a free doorway that reaches the exterior. Commit
also requires the room to pass the delivery checks of WorldLayout
(Sec.~\ref{supp:world:layout}), mesh generation
(Sec.~\ref{supp:world:mesh}) and WorldArt (Sec.~\ref{supp:world:art}).
In a grown world the layout check includes a path on the boxes for a
robot of 0.2\,m radius from the entrance to every other doorway. A
failed check stops the room, but violations left in a delivered layout
do not. In the grown worlds of Table~\ref{tab:task_readiness_expansion}, a room was committed once its
generation completed these delivery checks.

\paragraph{Rechecks at export.}
The integration checks of Sec.~\ref{sec:world_integration} use the compiled
MuJoCo~\cite{todorov2012mujoco} scene. The combined export checks nine
rays through a doorway and detects objects that extend more than
1\,cm beyond their room's walls. The export can fix an unstable body
in place when the task does not carry it. In the historical room
generation evaluation of Table~\ref{tab:task_readiness_expansion}, a task with failed preconditions
could be dropped from the later simulation export while its room
remained in the geometry-only world. For physical execution, the system checks the task's preconditions
before dispatch
(Sec.~\ref{supp:execution}).

\paragraph{Commit.}
Each added room is built on a copy of the stored world, which commit
swaps in only if the stored world's version and content hash are
unchanged. A journal completes or drops an interrupted commit. A failed
first room yields no room for execution. In the grown worlds of
Table~\ref{tab:task_readiness_expansion} the stored world holds geometry only, and each room's objects
join it at export.

\paragraph{Logged violations.}
The run record of a committed room keeps its layout status and the task
objects without a usable joint. It also counts the layout violations,
failed initial relations and unsupported objects left in its delivery.
An export adds the tasks it dropped with reasons, the bodies it fixed and
the deepest reach of an object past a wall. In the 30 grown worlds every
scheduled room was committed, and Table~\ref{tab:task_readiness_expansion} scores all 460 rooms with
their violations (Panel~B of Table~\ref{tab:supp:world:table1}).
\subsection{Task-Conditioned Room Generation}
\label{supp:world:table1}

\begin{figure*}[!tbp]
\centering
\includegraphics[width=0.88\textwidth]{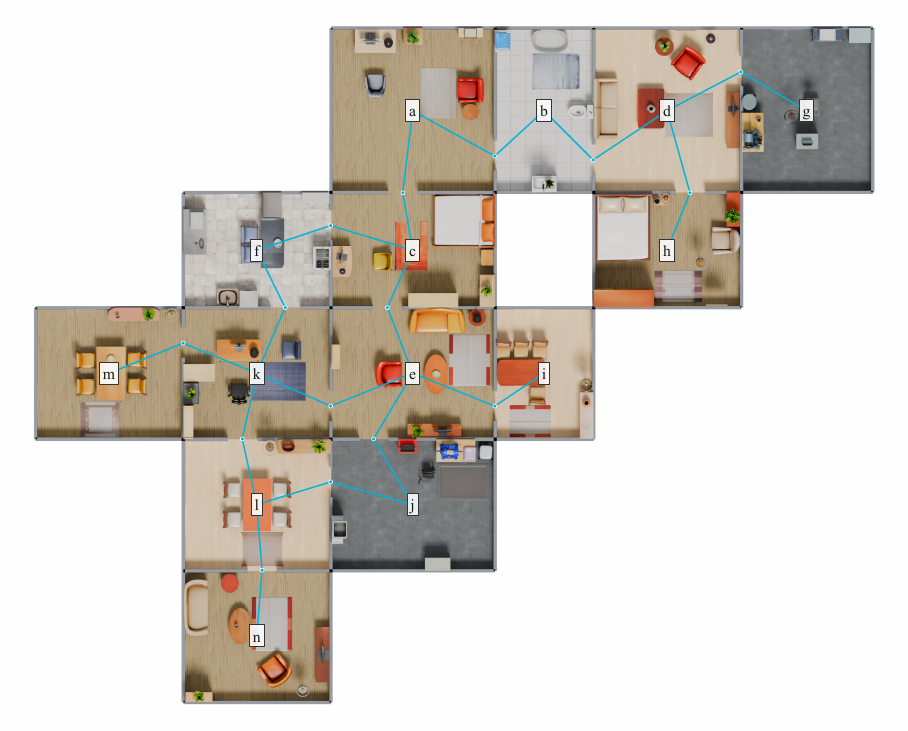}
\caption{\textbf{A world grown for one Panel~B sequence.}
Top-down view of the world \method{} grew from 14 tasks.}
\label{fig:supp:task:grownworld}
\end{figure*}

\begin{figure*}[!tbp]
\centering
\includegraphics[width=\textwidth]{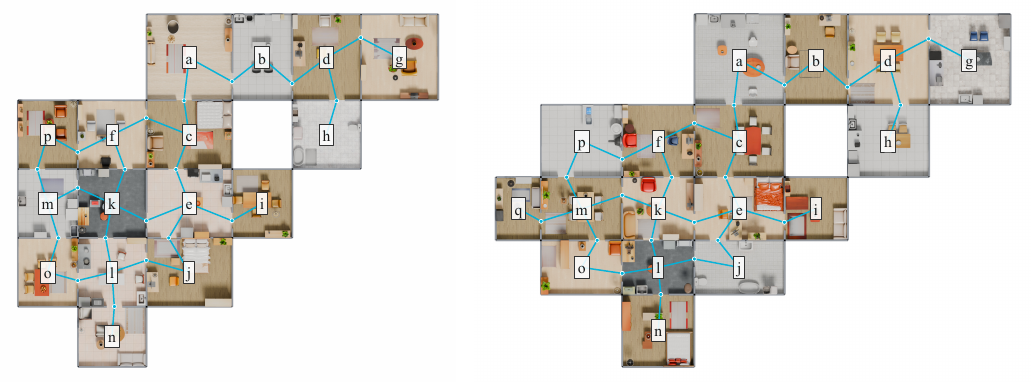}
\caption{\textbf{Two more grown worlds.}
Top-down views of a 16-room world (left) and a 17-room world (right),
marked as in Fig.~\ref{fig:supp:task:grownworld}.}
\label{fig:supp:world:grownworlds}
\end{figure*}

\paragraph{Evaluation protocol.}
Panel~A of Table~\ref{tab:task_readiness_expansion} uses 150 single-room Room-to-Task
tasks (Sec.~\ref{supp:task:roomtotask}) from 14 of the 16 source
datasets.
Their task graphs name 646 objects and 296 initial relations.
Each of the 30 Panel~B sequences has 13--17 tasks.
They are the easy tasks of the \method{} sessions of Table~\ref{tab:task_generation_expansion}
(Sec.~\ref{supp:task:online}), one sequence per session, kept in
round order.
Of the 509 easy tasks, the 49 whose instruction moves trinkets are left
out.
Figs.~\ref{fig:supp:task:grownworld} and~\ref{fig:supp:world:grownworlds}
show three of the grown worlds.
Each Panel~B task adds one new room and asks to put an object on a
support (443 tasks) or to switch on a lamp (17).
Its scripted outcome is the completion that the reference solver
committed in the session.
Room generation applies no outcome, and no Panel~B task depends on an
earlier room or object state.

\begin{figure*}[!tbp]
\centering
\includegraphics[width=\textwidth]{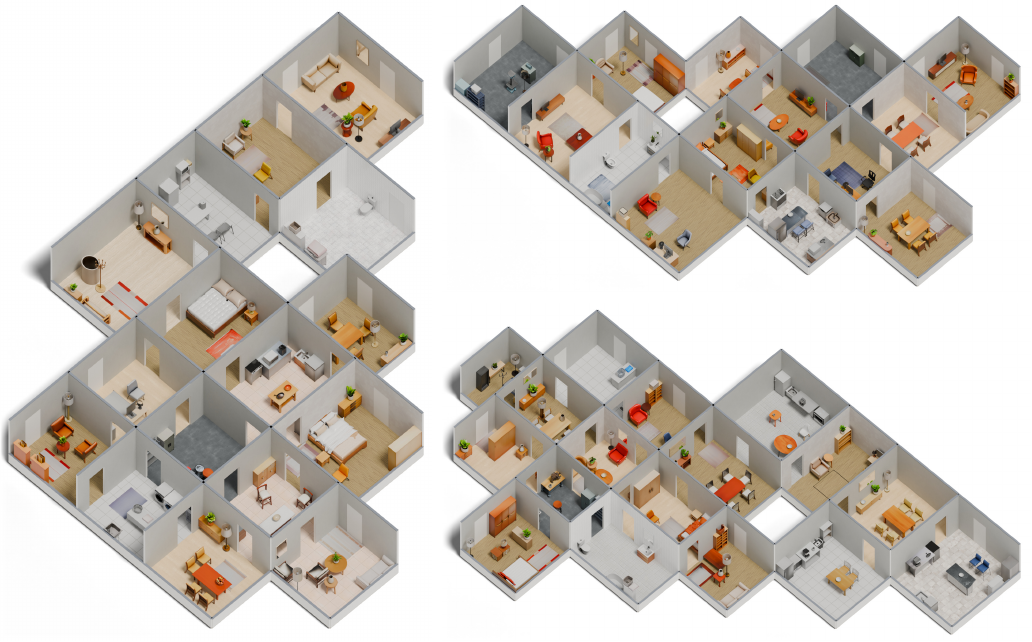}
\caption{\textbf{The three grown worlds seen at an angle.}
The world of Fig.~\ref{fig:supp:task:grownworld} (left) and the two
worlds of Fig.~\ref{fig:supp:world:grownworlds} (right, top and bottom).}
\label{fig:supp:world:grownworlds:oblique}
\end{figure*}

In the grown-world figures, letters mark the rooms in the order they were
added.
Room~a is the initial room, generated for the first task, and each later
room was generated for the next task and joined to the world at a free
doorway.
Cyan lines join the centres of rooms that share a doorway, through the
doorway's midpoint.
Objects are shown as generated, with no task executed.
Floor and wall materials were added for display by an optional
decoration step that ran after the Panel~B runs; it moves and adds no
object.
Top views cut all walls at 0.6\,m. In oblique views
(Fig.~\ref{fig:supp:world:grownworlds:oblique}), only the near walls are cut to that height.

Table~\ref{tab:supp:task:grownworldtasks} lists the task of each room of
Fig.~\ref{fig:supp:task:grownworld}.
Its Round column gives the round of the world's session (Table~\ref{tab:task_generation_expansion}) whose
easy task the room was generated for, counted as in
Table~\ref{tab:supp:task:session}.
Its instructions are verbatim from each room's task text, which has the
form of Table~\ref{tab:supp:world:design}.
For example, the task for room~a asks the agent to place the study's
desk lamp on its desk.
Panel~B leaves out rounds 6, 33 and 39 of this session, whose tasks move
trinkets.
Room~n was generated from scene script (a) of
Table~\ref{tab:supp:world:design}.

In Fig.~\ref{fig:supp:world:grownworlds}, the number after each room is
the same round.
The left world comes from another evaluation house with seed~1:
a~entrance hall (2), b~pantry break room (5), c~bedroom (7), d~study (10),
e~kitchen (13), f~office (18), g~living room (19), h~bathroom (22),
i~dining room (26), j~bedroom (30), k~workshop (33), l~kitchen (36),
m~laundry room (37), n~sunroom (42), o~dining room (45), p~library (50).
Panel~B leaves out easy round~46 of its session, whose task moves
trinkets.
The right world uses the easy tasks from the first house with seed~0,
the session sampled in Table~\ref{tab:supp:task:session}. Its rooms are
a~kitchenette (1), b~study (6), c~dining room (7), d~dining room (11),
e~bedroom (15), f~office (17), g~kitchen (20), h~kitchen (23), i~living
room (27), j~bathroom (29), k~living room (33), l~workshop (36), m~dining
room (39), n~bedroom (40), o~bedroom (44), p~laundry room (46), q~hallway
(49).
Each world was grown anew from its sequence and does not contain the
start house of its session.

\begin{table*}[!tbp]
\centering
\scriptsize
\setlength{\tabcolsep}{4pt}
\begin{tabularx}{\textwidth}{@{}llr>{\raggedright\arraybackslash}X@{}}
\toprule
Room & Type & Round & Instruction \\
\midrule
a & study & 1 & Put the desk lamp in the study on the desk in the study. \\
b & bathroom & 7 & Put the towel in the bathroom on the shelf in the bathroom. \\
c & bedroom & 11 & Put the desk lamp in the bedroom on the nightstand in the bedroom. \\
d & living room & 15 & Put the vase in the living room on the coffee table in the living room. \\
e & living room & 17 & Put the desk lamp in the living room on the side table in the living room. \\
f & kitchen & 20 & Put the pan in the kitchen on the kitchen island in the kitchen. \\
g & workshop & 23 & Put the jar in the workshop on the workbench in the workshop. \\
h & bedroom & 27 & Put the vase in the bedroom on the nightstand in the bedroom. \\
i & dining room & 29 & Put the plate in the dining room on the dining table in the dining room. \\
j & workshop & 36 & Put the towel in the workshop on the workbench in the workshop. \\
k & office & 40 & Put the desk lamp in the office on the side desk in the office. \\
l & dining room & 44 & Put the fruit container in the dining room on the dining table in the dining room. \\
m & dining room & 46 & Put the wineglass in the dining room on the dining table in the dining room. \\
n & living room & 49 & Put the wineglass in the living room on the side table in the living room. \\
\bottomrule
\end{tabularx}
\caption{\textbf{Rooms, generation rounds and task instructions for Fig.~\ref{fig:supp:task:grownworld}.} Letters match the figure.}
\label{tab:supp:task:grownworldtasks}
\end{table*}

\paragraph{Inputs and models.}
Each method receives a textual task description, as illustrated in Table~\ref{tab:supp:world:design}.
In Panel~B, HSM~\cite{pun2026hsm} additionally receives the room's floor size.
EmbodiedGen V2~\cite{wang2026embodiedgen} maps the task description to a room type
and complexity level, without explicitly conditioning on task-specific objects.
All baselines use their released implementations with GPT-6.1-sol~\cite{openai2026gpt61sol}
replacing their original LLMs. \method{} also uses GPT-6.1-sol.
Each task is evaluated once without retrying model outputs, and all runs finish within the 3\,h time limit.

\paragraph{Sequential addition.}
\method{} grows one world per Panel~B sequence, and the metrics use the
rooms it commits under the rule of Sec.~\ref{supp:world:integration}.
A baseline generates each Panel~B room on its own (``per room +
attach'').
Our wrapper then attaches room~$k$ through one door, reused or cut, to
the latest placed room it fits beside with at most 1\,cm of floor
overlap.
A room that fits beside no placed room counts as zero.
Every Panel~B room is scored alone, with only the openings that join it
to other rooms of its sequence.

\paragraph{Counting rule.}
Each scheduled task has equal weight: 150 tasks per method in Panel~A
and 460 tasks across 30 sequences per method in Panel~B. A failed
generation or unscored room contributes zero to all four percentage
columns. For each column we sum the per-task scores, divide by the
number of scheduled tasks, and multiply by 100.
\method{} uses post-processing to correct object placement in generated rooms.
For the 150 \method{} rooms in Panel~A, paired scoring before and after
the final repair stage gives Task Ground of 86.7\% and 85.0\% and Physics
Pass of 28.0\% and 30.7\%, respectively. Table~\ref{tab:task_readiness_expansion}
reports the former Task Ground and the latter Physics Pass. Task Nav
passes in 149/150 rooms and Task-Ready in 26/150 at each stage.
The final column of Table~\ref{tab:supp:world:table1} counts rooms passing all four geometric and access checks.

\paragraph{Failed and missing baselines.}
Holodeck~\cite{holodeck} delivered no room in any of the 460 Panel~B
runs and in 140 of the 150 Panel~A runs.
All 140 Panel~A stops and 365 of the Panel~B stops came from a
floor-plan line that did not split into the four fields its parser
expects.
The replacement LLM wrote these lines.
Holodeck's Panel~B row prints N/A where the counting rule would give
zeros.
SAGE~\cite{sage} enters Table~\ref{tab:panel_a_sceneeval} with box layouts (53.89\,s per scene).
Table~\ref{tab:task_readiness_expansion} needs SAGE's full pipeline, with text-to-3D meshes and physics
simulation; a partial Panel~A run of it was stopped and is not reported.
MANSION~\cite{Che_2026_CVPR} supplies only layouts in Table~\ref{tab:panel_a_sceneeval}. Its full pipeline and SceneSmith's are too slow for the 610 rooms in Table~\ref{tab:task_readiness_expansion}.

\begin{table}[!tbp]
\centering
\footnotesize
\setlength{\tabcolsep}{2.8pt}
\begin{tabular}{@{}lrrrrrr@{}}
\toprule
Method & Scored & Col. & Sup. & Acc. & OOB & All \\
\midrule
\multicolumn{7}{@{}l}{\textit{A. Single rooms, 150 tasks}}\\
Holodeck~\cite{holodeck} & 10  & 6   & 0   & 1               & 10  & 0   \\
HSM~\cite{pun2026hsm} & 150 & 38  & 0   & 53              & 67  & 0   \\
EmbodiedGen V2~\cite{wang2026embodiedgen} & 150 & 127 & 5   & n/a$^{\dagger}$ & 136 & 0   \\
\method{}      & 150 & 148 & 51  & 118             & 149 & 46  \\
\midrule
\multicolumn{7}{@{}l}{\textit{B. Sequential room addition, 460 tasks}}\\
Holodeck       & 0   & --  & --  & --              & --  & --  \\
HSM            & 457 & 156 & 2   & 191             & 211 & 0   \\
EmbodiedGen V2 & 412 & 304 & 47  & n/a$^{\dagger}$ & 351 & 0   \\
\method{}      & 460 & 457 & 264 & 350             & 460 & 199 \\
\bottomrule
\end{tabular}

\vspace{2pt}
{\scriptsize $^{\dagger}$Stored count 0, not reported (see text).\par}
\caption{\textbf{Scored rooms in Table~\ref{tab:task_readiness_expansion}.} Counts show how many pass each physics check and all four.}
\label{tab:supp:world:table1}
\end{table}

\paragraph{Metrics.}
The following definitions use SceneEval~\cite{sceneeval} for geometry
and accessibility.

\emph{Task Ground} scores a room by the share of satisfied items: the
task-graph objects other than floor and wall, and the initial relations.
A VLM assigns each room object, from its front render, to at most one
required category.
The one-to-one assignment of task objects to room objects that satisfies
the most items is kept, found exhaustively up to 100,000 candidate
assignments and greedily above.
A relation with the floor needs SceneEval's support check to find the
object on the ground.
A relation with a wall or $\mathrm{in}(x,y)$ needs SceneEval's wall or
inside score above 0.5.
$\mathrm{on}(x,y)$ needs the larger of the on-top and inside scores
above 0.5 and $x$ resting on $y$.
Resting means that a downward ray from 5\,mm above one of the lowest
vertices of $x$ first hits $y$ within 1\,cm.
An $x$ that rests on nothing instead needs the on-top score alone above
0.5 and the two meshes closer than 1\,cm.
Panel~B tasks have no initial relation, so there the column measures
object presence.

\emph{Task Nav} passes a room when the largest connected free region
covers at least 0.9 of the free floor and reaches every door.
The free floor is SceneEval's 256-pixel raster eroded for a 0.2\,m
agent.
Objects whose box centre is at most 2\,m high are obstacles, except
objects lower than 0.1\,m that rest on the floor, such as rugs.
A door counts as reached when a line across it, 0.15\,m into the room,
meets that region.

\emph{Physics Pass} requires every object of the room to pass four
SceneEval checks.
Collision (Col.) counts a contact only if it persists after a 5\,mm
shift.
Support (Sup.) casts rays from the object's contact side, offset by 5\,mm,
along the support direction that a VLM names, and needs a surface within
1\,cm.
Accessibility (Acc.) needs the best functional side, also named by the VLM, to
keep at least half of a 0.2\,m strip in front of it free.
An object for which the VLM names no functional side is not tested.
Out of bounds (OOB) needs at least 99\% of the object's samples over the
floor.
The per-object checks and their parameters are SceneEval's; the
whole-room rule and the 0.5 accessibility threshold are ours.

\emph{Task-Ready} (a Task Ground of~1 and both passes) does not test
that the goal is unmet at the start or that the task can be executed.
The downstream evaluation of Sec.~\ref{sec:downstream} addresses both.

\emph{Obj./room} counts the object instances of a generated room, without
walls, floor and ceiling.
\emph{Time} is wall-clock seconds from the input text to the finished
room, without format conversion and attachment for the baselines.
For \method{} it sums the jobs of a resumed task and includes the
Panel~B planning at a free doorway.
Holodeck, HSM and EmbodiedGen V2 ran on CPU only and \method{} on GPUs,
so Time is not hardware-matched.

\paragraph{Weakest check.}
Apart from the EmbodiedGen V2 accessibility counts, support passes in
the fewest rooms for every method and panel with scored rooms
(Table~\ref{tab:supp:world:table1}).
The 5 and 47 EmbodiedGen V2 rooms that pass support also pass collision
and out of bounds, and the stored records fail them on accessibility.
Table~\ref{tab:supp:world:table1} does not report this stored count of 0.
The exported EmbodiedGen V2 rooms do not record object fronts, and our
conversion gave all objects one front direction, on which this check
depends.
The loaders that read Holodeck and HSM rooms for scoring take the floor
height from the lowest object, even when the floor vertices carry a
height. In one HSM room this lowers the floor by about 109\,mm, so a
counter standing on the floor fails the support check.
The shared front direction also misdirects wall-support rays for EmbodiedGen V2.
\subsection{Preservation Across Expansion}
\label{supp:world:preserve}

Preservation means that adding a room leaves the existing world as it
was.
Existing rooms keep their geometry, apart from the opening edits of
Sec.~\ref{supp:world:integration}.
Existing objects keep their IDs, poses, box sizes, and state fields
(Sec.~\ref{sec:task_specification}).
The interaction history $H_t$ and later tasks therefore name the same
objects that earlier tasks changed.
In the 90 sessions with adapted task proposers (Voyager~\cite{wang2023voyager}, OMNI-EPIC~\cite{faldor2025omni} and
SEAgent~\cite{sun2025seagent}), replay of the shared staging and commit steps reproduced the
saved worlds in all 4,013 committed rounds (Sec.~\ref{supp:task:generator}).

\paragraph{Two conditions.}
The box-level ablation uses a separate set of 30 \method{} sessions from
Table~\ref{tab:task_generation_expansion}, ten houses with three seeds each.
The sessions hold 1,483 delivered tasks, each with its recorded new room,
instruction, and solver plan.
No LLM is called, and no task is proposed again.
\emph{Ours} rebuilds each session from its start house with the code of
the reported runs.
Each round grows its recorded room at the recorded doorway, runs its
recorded plan, and commits.
The per-round worlds of these sessions were not kept.
The replay reproduces the stored final world of all 30 sessions, compared
by content hash, and every grown room matches its stored record.

\emph{No-preserve} starts each round from the same world $W_t$ but first
generates the objects of every existing room again.
The box-level furnishing step refurnishes each room from its own request
with a new seed, and the start house receives new small objects.
Every regenerated object gets a new ID.
A draw is kept once it passes the growth step's room checks.
Each room gets up to 20 draws; 7 of the 59,652 room regenerations kept a
draw that passed none.
Rooms, doorways, door states, and the robot's room stay as in Ours.
The round's new room is then grown as in Ours, and it came out identical
in all 1,483 rounds.
No-preserve thus stands for a generator that rebuilds the world from the
accumulated room requests and keeps no object state.

\paragraph{Metrics.}
Values are percentages, aggregated as in Table~\ref{tab:task_generation_expansion}.
Each is the mean over the ten houses of each house's mean over its three
seeds, with the 95\% CI over houses as a subscript where it is not zero.
Identity retention is the share of objects present before the round whose
ID still exists after it.
Geometry retention is the share of objects that the round's task does not
move and that keep their room, pose, and box size.
Reference validity is the share of tasks for which every object named in
the recorded plan exists in the staged world $\widehat W_t$.
Of the 955 tasks with a \emph{Builds on} entry in the round records,
five cite only a room.
History validity covers the other 950, each of which uses an object
moved by an earlier task.
Such a task counts when each of these objects still stands on or in the
support or container where that earlier task left it.
Feasibility is the share of tasks for which the symbolic solver reaches,
from $\widehat W_t$, the end state of the recorded plan.
State carry-over follows every object changed by an earlier task.
It is the share of these objects that start the task with the placement
and the open, wiped, and power states left by their last change.

The last four metrics are computed under two bindings.
ID uses the recorded IDs.
Class replaces an absent ID with an unused object of the same class in
the same room, one on the same support class first.
Geometry retention is reported by class, so an object counts when an
object of its class still stands at its pose with its box.
Re-binding reads each object's room from Ours, which favors No-preserve.
IDs are renewed because the furnishing step composes them from the parent
room, wall, class, and index.
Without renewal, 1,881 of the 2,032 references to existing objects
(92.6\%) would match a regenerated ID, although the object behind it is a
new sample.

\begin{table*}[t]
\centering
\caption{\textbf{Preservation ablation.} Paired symbolic replays cover 1,483 delivered tasks across 30 \method{} sessions.}
\label{tab:supp:world:preserve}
\footnotesize
\setlength{\tabcolsep}{0.5pt}
\renewcommand{\arraystretch}{1.15}
\newcommand{\fx}[1]{\textcolor{neutralM}{#1}}
\begin{tabular*}{\textwidth}{@{\extracolsep{\fill}}lcccccccccc@{}}
\toprule
& Identity & Geometry & \multicolumn{2}{c}{Reference validity} & \multicolumn{2}{c}{History validity} & \multicolumn{2}{c}{Feasibility} & \multicolumn{2}{c}{State carry-over} \\
\cmidrule(lr){2-2}\cmidrule(lr){3-3}\cmidrule(lr){4-5}\cmidrule(lr){6-7}\cmidrule(lr){8-9}\cmidrule(l){10-11}
Condition & ID & Class & ID & Class & ID & Class & ID & Class & ID & Class \\
\midrule
\method{} (Ours) & \fx{100.0} & \fx{100.0} & \fx{100.0} & \fx{100.0} & \fx{100.0} & \fx{100.0} & \fx{100.0} & \fx{100.0} & \fx{100.0} & \fx{100.0} \\
No-preserve & \fx{0.0} & 1.8$_{\pm 0.3}$ & \fx{34.3$_{\pm 0.2}$} & 68.7$_{\pm 2.7}$ & \fx{0.0} & 0.8$_{\pm 0.6}$ & \fx{34.3$_{\pm 0.2}$} & 68.6$_{\pm 2.6}$ & \fx{0.0} & 0.7$_{\pm 0.3}$ \\
\bottomrule
\end{tabular*}

\vspace{2pt}
{\scriptsize\raggedright Percentages. Gray cells are fixed by design. Before each round, Ours keeps existing objects and No-preserve regenerates them. History validity covers 950 tasks that reuse a previously moved object.\par}
\end{table*}

\paragraph{Results.}
Ours keeps every cell of Table~\ref{tab:supp:world:preserve} at 100 by
construction.
It keeps IDs, geometry, and states, and its recorded tasks were generated
and solved on that world.
No-preserve gives every regenerated object a new ID, so its ID cells are
0 for objects that existed before the round.
Only easy tasks retain valid references and remain feasible under ID binding (509 of 1,483 tasks overall). The table reports the mean across houses.
Under class matching, 1.8\% of objects have a regenerated counterpart of
the same class, pose and box size.
After re-binding, 68.7\% of the tasks find every object they name, and
68.6\% remain feasible.
Easy tasks act only in the new room and stay valid and feasible in both
conditions.
Medium tasks remain feasible at 69.1$\pm$5.5\% after re-binding, and hard
tasks at 33.9$\pm$6.3\%.
History validity falls to 0.8\% and state carry-over to 0.7\%.
A regenerated room holds the classes its request names, not the objects
that earlier tasks moved in or out.
Of the 2,032 references to existing objects, 567 cannot be re-bound, and
500 of these name objects that an earlier task had moved.
A second No-preserve sample, with every regeneration seed changed, gives
68.4$\pm$2.4\% for reference validity and feasibility by class.
It gives 0.5$\pm$0.4\% for history validity and 1.8$\pm$0.3\% for
geometry retention.

\begin{figure*}[t]
\centering
\includegraphics[width=\textwidth]{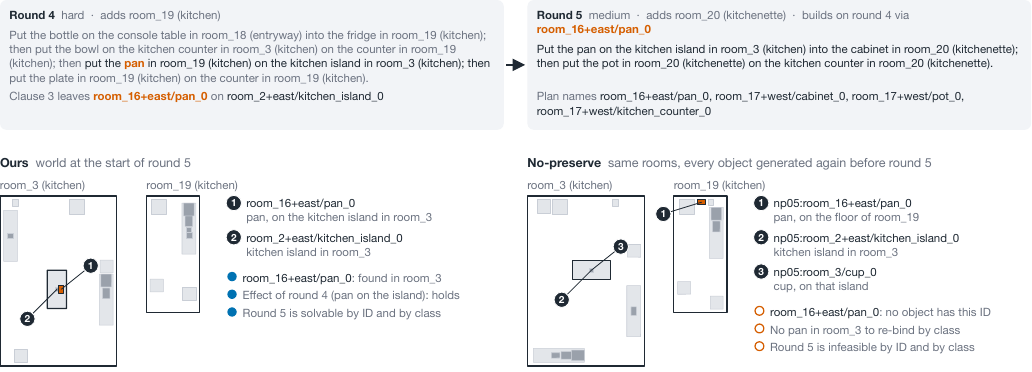}
\caption{\textbf{Where the pan starts round~5.} Top views of rooms~3 and~19 compare preservation with refurnishing after round~4 of Table~\ref{tab:supp:task:session}.}
\label{fig:supp:world:preserve}
\end{figure*}

\paragraph{A pan carried across two kitchen tasks.}
Fig.~\ref{fig:supp:world:preserve} follows rounds 4 and 5 of the session
of Table~\ref{tab:supp:task:session}.
The figure gives the original instructions and a top view from each
replay. Light boxes mark objects on the floor and dark boxes mark
objects on furniture. The view omits objects inside containers.
Filled discs mark passes and open rings mark failures.

In round~4 the agent moves a pan from the newly added kitchen to an
island in the starting house. Round~5 asks it to put that same pan in
the new kitchenette's cabinet. With preservation, the pan is still on
the island at the start of round~5. After refurnishing, the island
holds a cup. A newly generated pan lies on the floor of the kitchen
added in round~4. Even a search by object class finds
no pan in the specified room, so the solver cannot execute round~5
under either binding. Refurnishing also erases the other three round~4 placements,
removing the bottle from the new kitchen's fridge and the bowl and plate from its counter.

\paragraph{Limits.}
Ours is a replay, so its intermediate worlds are checked through each
grown room and the final world, not against stored copies.
Feasibility takes the end state of the recorded plan as the goal, since
the bound goals were not stored, and it runs no physics check.
No-preserve regenerates objects only; it keeps every room and doorway and
grows each new room as Ours does.
The start houses were built before the code of the reported runs.
From the original seeds, the furnishing step of these runs returns 303
of the 450 stored start-house layouts.
It returns all 1,483 layouts of the rooms grown in the sessions.

\section{Task Execution and State Persistence}
\label{supp:execution}

\paragraph{Room delivery and execution.}
For a task that calls for a new room, the integration step checks the
assembled scene for collisions, doorway clearance, and the task's
initial conditions before execution (Sec.~\ref{sec:world_integration}).
The system records the accepted room and its initial object states
before the policy starts. Tasks that use existing rooms start from the
objects' current states. If the new room cannot be built or attached,
the next proposal starts from $W_t$.

\paragraph{State and feedback after a task.}
The simulator returns a trajectory and a terminal state. An independent
checker evaluates the bound task goals against the measured state.
Under the default \emph{retain} protocol, the next task receives the
terminal object states after either success or failure, and the delivered
rooms stay in the world. Under \emph{rollback}, a failed execution
restores the pre-execution object states while keeping the newly
generated room (Eq.~\ref{eq:method:persistence}). Both protocols record the
trajectory and return the outcome to the task generator. If a measured
terminal state cannot be represented in the stored world, the retain
sequence stops and the trajectory is saved for inspection.

\subsection{Downstream Evaluation Protocols}

\begin{figure*}[t]
\centering
\includegraphics[width=.97\textwidth]{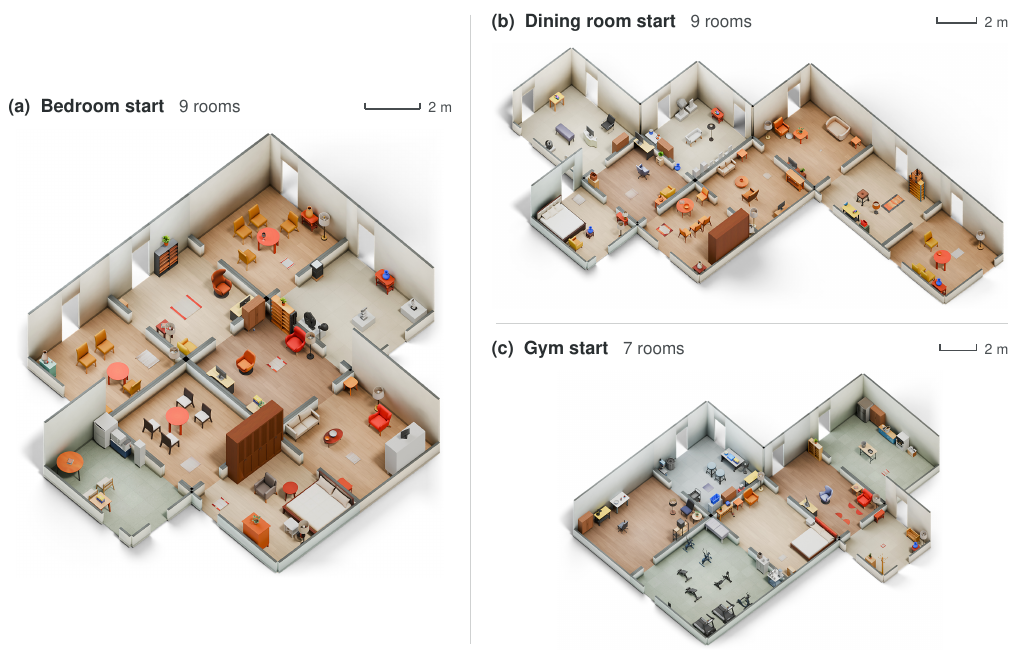}
\caption{\textbf{Three worlds grown from different starting rooms.}
The cutaways show the archived layouts and furniture poses for three
starting rooms (bedroom, dining room and gym). Surface finishes, tabletop plants
and books are illustrative.}
\label{fig:supp:online:worlds}
\end{figure*}

\paragraph{Robot policy evaluation.}
For the policy comparison in Table~\ref{tab:robot}, each of the 30 Panel~B sequences supplies its first ten scheduled tasks. The world of a sequence is called a house below. Of these 300 tasks,
141 already meet their goals at the start and ten have no goal the
simulator checker can verify, leaving 149 eligible tasks. Table~\ref{tab:robot}
counts an interaction when the target object moves at least 3\,cm
during the task. This criterion was added after the first strict
results and later selected for the policy comparison.
Under strict goal completion the five learned policies record zero successes on their evaluated tasks.
SR@5 is the
fraction of houses with a successful interaction on every eligible
task among the first five scheduled tasks. Tasks/10 is the mean
number of successful interactions per house over its ten tasks.

The $\pi_{0.5}$ policy~\cite{black2025pi05} runs on Lift2 with the EBench checkpoint~\cite{gao2026ebench}.
On G1, we evaluate GR00T N1.6 and N1.7~\cite{bjorck2025gr00t},
OpenHLM~\cite{hu2026openhlm}, and $\omega$-0~\cite{li2026omega0}. The OpenHLM and $\omega$-0 sequence scores
aggregate independent task episodes,
after excluding one OpenHLM episode and eleven $\omega$-0 episodes
that ended in evaluation errors.
Their interaction counts are 35/148 and 19/138.
For the Agentic row, the logged base position, heading, room and held
object agree at all 270 boundaries between successive tasks.
These summaries cover only those four state fields.

The single-room column also uses policy-specific subsets.
Agentic scores 24/29 and $\pi_{0.5}$ scores 0/29 over 30 scheduled
rooms. In the ten-room subsets, GR00T N1.7, OpenHLM and $\omega$-0
score 2/9, 3/9 and 2/9, respectively. Table~\ref{tab:robot} omits a single-room
result for N1.6.

\paragraph{Planner comparison (Table~\ref{tab:agentic_llms}).}
The five planners use the same prompts and navigation, pick and place
skills on TidyBot~\cite{wu2023tidybot}. They receive privileged scene state and use assisted
grasping by kinematic attachment. The skills were tuned on the evaluation
worlds before freezing. Each task allows an initial plan and at most
three replans. Strict success requires all task goals to hold for 2\,s
after the gripper releases the manipulated objects and they are stable.
The checker records the first time this condition holds. Task SR, SR@5
and Tasks/10 use strict goal completion on the 30 house sequences.
Interact and mean planner calls per task pool the 149 eligible
sequence tasks and 29 eligible single-room tasks. For GPT-6.1-sol~\cite{openai2026gpt61sol},
90/149 sequence tasks and 9/29 single-room tasks satisfy the
strict goal. The pooled interaction rate is 143/178 (80.3\%), and
402 planner calls yield 2.3 calls per eligible task.

\paragraph{Video world models (Table~\ref{tab:wm_game_v2}).}
Each model receives the first frame and the commands from the same
60 single-room walkthroughs of 30\,s each. For Yume-1.5~\cite{mao2026yume1} the adapter
maps each command window to its most frequent action at a fixed
magnitude. Each of the
30 rooms supplies an exploration path and a revisit path. LPIPS~\cite{zhang2018unreasonable}
compares generated and rendered frames at 5 and 30\,s. Action
agreement uses 1,677 command segments whose motion is identifiable
in the reference video. VGGT~\cite{wang2025vggt} camera estimates are compared with the
rendered path after anchoring both trajectories at the start of each
evaluation window.
Rotation error in degrees and MEt3R~\cite{asim2025met3r} on frame pairs
0.5\,s apart cover all 60 clips. For translation error, we fit a
nonnegative scale and normalize by reference displacement. The metric is
defined for 59 clips per model. Object recall counts OWLv2~\cite{minderer2023scaling} detections over 6,371
visible category--frame observations, without instance or spatial matching. On the
rendered references the same detector recovers 5,345/6,371 (83.9\%).
Figure~\ref{fig:exec} supplies separate qualitative text-conditioned interaction
examples.

\subsection{Online Expansion and Execution Feedback}
\label{supp:online}

In two online runs, a scripted TidyBot attempted placement tasks in MuJoCo.
WorldEdge saved the measured object state after execution. For each follow-up,
the generator received the current inventory, the robot's report from the most
recent scored execution in that world, and the independent validator's verdict. Each run
scheduled ten independent worlds and used a seeded difficulty schedule.
The pilot ran for up to ten rounds per world. Figure~\ref{fig:supp:online:worlds}
shows three expansion worlds, and Table~\ref{tab:supp:online:worlds} summarizes
both runs.

\begin{figure*}[t]
\centering
\includegraphics[width=.87\textwidth]{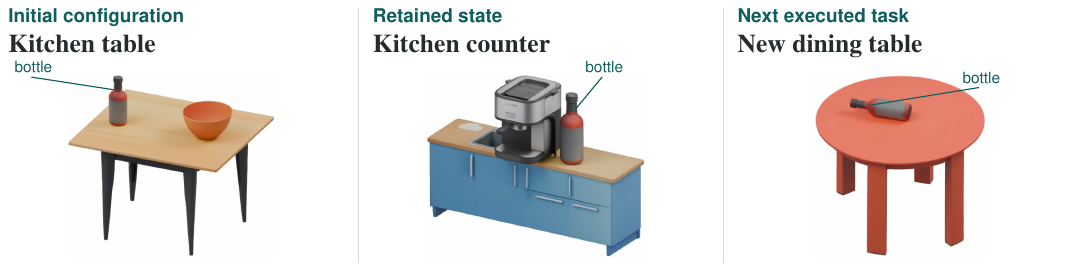}\par\smallskip
\begin{minipage}{.96\textwidth}
\small
\begin{tabularx}{\linewidth}{@{}p{.17\linewidth}p{.24\linewidth}X@{}}
\toprule
Record & Source in the archive & Recorded observation\\
\midrule
Generator input & Proposal prompt & The bottle is now on the kitchen counter. Its previous support was the kitchen table.\\

Selected task & Agent request & Move the bottle from the counter and the fruit bowl from the kitchen table to the new dining table.\\

Robot events & Episode log & The gripper attached to the bottle at 20.34\,s and released it at 66.28\,s.\\

Terminal state & Execution receipt & Bottle on dining table: true. Fruit bowl on dining table: false.\\
\bottomrule
\end{tabularx}
\end{minipage}
\caption{\textbf{Bottle position across two executed tasks.}
The bottle moved from the kitchen table to the counter. Its measured pose on
the counter was unchanged at the start of the next dispatched task. The bottle
reached the new dining table. The fruit bowl remained in the kitchen.
The table summarizes the follow-up task's archived prompt, selected instruction,
robot event log and terminal goal predicates. Times are relative to the start of that execution.}
\label{fig:supp:online:carryover}
\end{figure*}

\paragraph{Choosing the next task.}
The generator proposed tasks for a new room using the saved rooms, objects,
supports and execution feedback. WorldEdge checked each candidate for feasibility and gave priority to
candidates matching the scheduled difficulty (Table~\ref{tab:supp:online:feedback}).
If no model proposal was selected, the generator tried a fallback task. Selected tasks specified
one to five placement relations on supports or in containers.

\medskip
\noindent\begin{minipage}{\linewidth}
\centering
\captionsetup{type=table}
\caption{\textbf{Rooms, objects and doorways in the saved worlds.}
We count each retained object instance once and each pair of rooms joined by
a doorway once. Every retained room had a path to its world's starting room
in the doorway graph. The room-count range covers worlds with a saved first room.}
\label{tab:supp:online:worlds}
\small
\setlength{\tabcolsep}{4pt}
\begin{tabularx}{\columnwidth}{@{}Xrr@{}}
\toprule
Measure & Pilot & Expansion\\
\midrule
Worlds with a first room & 8 & 10\\
Retained rooms & 35 & 75\\
Rooms per world (range) & 2--6 & 3--9\\
Object instances & 410 & 851\\
Objects per room & 11.7 & 11.3\\
Doorway connections & 31 & 69\\
\bottomrule
\end{tabularx}
\end{minipage}\\[.6\baselineskip]
The bedroom-start world in Fig.~\ref{fig:supp:online:worlds}a grew to nine rooms
over fourteen task rounds. In five rounds, the proposed scene failed its initial
checks, so the robot did not start the task and the room was not retained.

\raggedbottom
\paragraph{State reuse.}
In three pilot worlds, a later executed task reused an object after the
robot had moved it to a different support. In the case shown in
Fig.~\ref{fig:supp:online:carryover}, the prompt identified the kitchen counter
as the bottle's support. The first four proposals failed the feasibility check.
From the next four, WorldEdge selected a medium-difficulty task requiring
a new dining room and the transfer of the bottle and fruit bowl to its table.
The change of support and the next pickup location refer to the same archived
object ID. The retained position and orientation matched the initial pose
for the next dispatched task in all seven stored components.

\medskip
\noindent\begin{minipage}{\linewidth}
\centering
\captionsetup{type=table}
\caption{\textbf{From execution feedback to the next task.}
Panel A counts all closed follow-up rounds and excludes the supplied first
task in each world. Panel B describes the selected tasks. An existing-object
reference counts when a placement relation names an instance in the saved
inventory.}
\label{tab:supp:online:feedback}
\small
\setlength{\tabcolsep}{3pt}
\begin{tabularx}{\columnwidth}{@{}Xrr@{}}
\toprule
Measure & Pilot & Expansion\\
\midrule
\multicolumn{3}{@{}l}{\itshape A. Closed follow-up rounds}\\
Prompt included both feedback reports & 72/72 & 95/95\\
Next task selected & 72/72 & 94/95\\
Execution scored by validator & 27/72 & 63/95\\
\midrule
\multicolumn{3}{@{}l}{\itshape B. Selected tasks}\\
Model proposal / feasible fallback & 69 / 3 & 93 / 1\\
Existing object or support named (\%) & 41.7 & 53.2\\
Matched the scheduled difficulty & 69/72 & 92/94\\
Easy / medium / hard & 24 / 25 / 23 & 34 / 31 / 29\\
\bottomrule
\end{tabularx}
\end{minipage}\\[.6\baselineskip]
We matched both feedback excerpts in each saved prompt, including retries,
to the latest scored execution in the same world. In some trials the robot moved a target to the requested support. The
validator judged every scored follow-up incomplete in both runs.

\paragraph{Paired replay.}
We replayed one physical checkpoint twice, varying the bottle pose and
support reported to the generator. With the measured state, the generator identified the counter as the pickup
location. With the earlier state, it identified the kitchen table. Both proposals specified the new dining table as the destination.
The robot located the bottle by instance ID in both replays. Neither
execution passed physical validation.

\stopcontents[appendix]

\end{document}